\documentclass[11pt]{article}

\usepackage[preprint]{acl}

\usepackage{textcomp}  % Required for encoding \textbigcircle
\usepackage{scalerel}  % Required for emoji \scalerel
\def\uhel{\textsuperscript{\scalerel*{\includegraphics{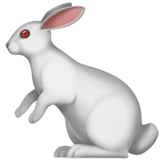}}{\textrm{\large\textbigcircle}}}}
\def\ptor{\textsuperscript{\scalerel*{\includegraphics{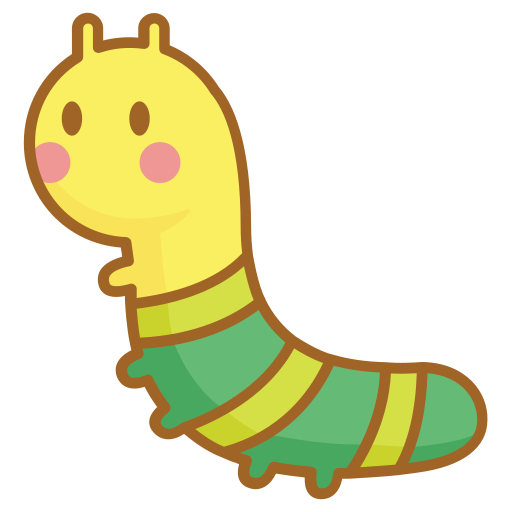}}{\textrm{\large\textbigcircle}}}}
\def\ubs{\textsuperscript{\scalerel*{\includegraphics{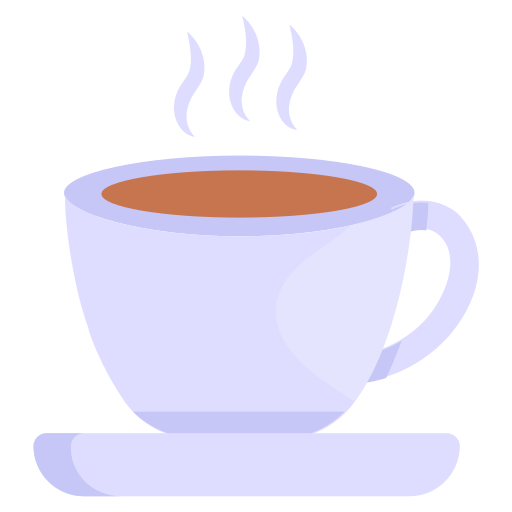}}{\textrm{\large\textbigcircle}}}}
\def\uutr{\textsuperscript{\scalerel*{\includegraphics{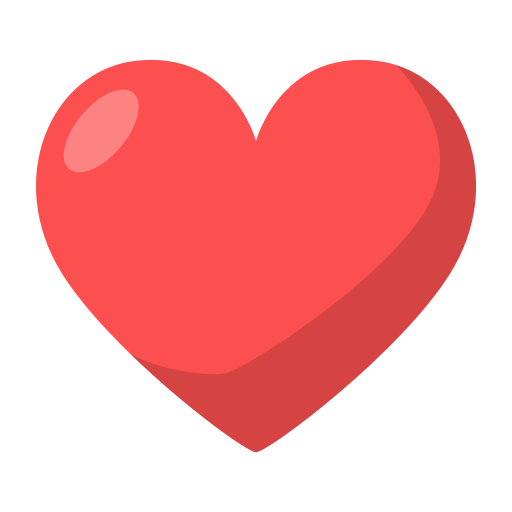}}{\textrm{\large\textbigcircle}}}}
\def\ucop{\textsuperscript{\scalerel*{\includegraphics{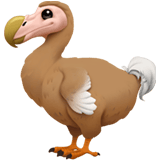}}{\textrm{\large\textbigcircle}}}}
\def\ugre{\textsuperscript{\scalerel*{\includegraphics{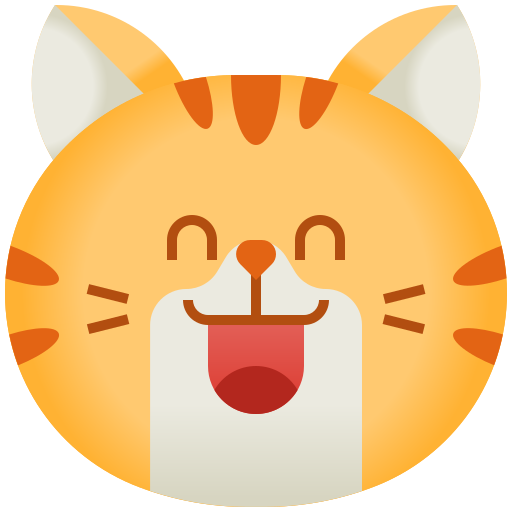}}{\textrm{\large\textbigcircle}}}}
\def\ulor{\textsuperscript{\scalerel*{\includegraphics{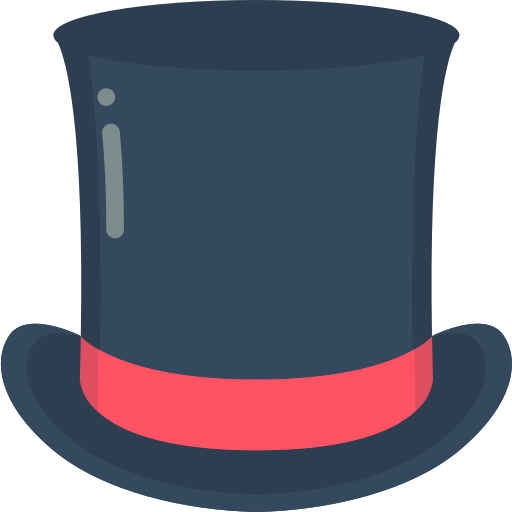}}{\textrm{\large\textbigcircle}}}}
\def\mzb{\textsuperscript{\scalerel*{\includegraphics{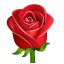}}{\textrm{\large\textbigcircle}}}}

\usepackage{times}
\usepackage{latexsym}
\usepackage{booktabs}
\usepackage[all]{nowidow}

\usepackage[T1]{fontenc}
\usepackage[utf8]{inputenc}

\usepackage{microtype}

\usepackage{inconsolata}

\usepackage{graphicx}
\usepackage{amsmath}
\usepackage{amssymb}
\usepackage{booktabs}
\usepackage{multicol, multirow}
\usepackage{colortbl}
\usepackage{comment}
\usepackage{enumitem}
\usepackage{cleveref}
\usepackage{subcaption}
\usepackage{tikz}
\usetikzlibrary{shapes.geometric, arrows.meta}
\usepackage{appendix}
\usepackage{tcolorbox}
\usepackage{bbding}
\usepackage{subcaption}
\usepackage{supertabular}

\usepackage{enumitem}
\usepackage{relsize}
\usepackage{siunitx}
\usepackage{multirow}
\usepackage{placeins}
\usepackage[all]{nowidow}

\setitemize{noitemsep,topsep=0pt,parsep=0pt,partopsep=0pt}

\usepackage{soul,xcolor}

\sethlcolor{red!25}
\newcommand{\hlA}[1]{\hl{#1}}

\newcommand{\hlB}[1]{%
  {\sethlcolor{red!45}\hl{#1}}%
}

\newcommand\orgname[1]{\noindent\textbf{#1}}

\title{Overview of SHROOM-Visions 2026: A Shared Task on Hallucination Detection in Large Vision-Language Models}

\author{Raúl Vázquez\uhel \hfill \textbf{Aman Sinha}$^\bigstar$\ulor \hfill
\textbf{Chuyuan Li}$^\bigstar$\ugre \hfill Artem Shelmanov\mzb \\ 
\textbf{Artem Vazhentsev}\mzb \hfill \textbf{Claudio Savelli}\ptor \hfill \textbf{Eduardo Calò}\uutr \hfill 
\textbf{Emilio Raimond}\ubs \\
\textbf{Stella Frank}\ucop \hfill \textbf{Hengyu Luo}\uhel \hfill \textbf{Flavio Giobergia}\ptor \hfill \textbf{Vincent Segonne}\ubs \\
\textbf{Lorenzo Vaiani}\ptor \qquad \textbf{Jörg Tiedemann}\uhel \qquad \textbf{Timothee Mickus}\uhel
\\%[0.2cm] 
$^\bigstar$\small{These authors have equal contributions}\\
\uhel University of Helsinki \hfill \ptor Politecnico di Torino \hfill \uutr Universiteit Utrecht \hfill
\ubs Université Bretagne Sud \\\ \ucop University of Copenhagen \hfill \ugre University Grenoble Alpes \hfill \ulor University of Lorraine \hfill \mzb MBZUAI\\
}

\begin{document}
\maketitle
\begin{abstract}
In 2026, we held the fourth iteration of the SHROOM Shared Task series
: SHROOM-Visions (\textbf{S}hared-task on \textbf{H}allucinations and \textbf{R}elated \textbf{O}bservable \textbf{O}vergeneration \textbf{M}istakes in \textbf{Vision} language model\textbf{s}), which is hosted at the UncertaiNLP Workshop co-located with EMNLP 2026. Following the success of the 2024 and 2025 tasks, this time we 
aim to tackle hallucinations through a model-agnostic detection task focused on large vision-language models. Building on the recently introduced SHEEP dataset, designed for long-term evaluation across model generations, the task invites participants to detect and classify fine-grained hallucination spans in image-conditioned text generation (VQA, image captioning, etc.). The evaluation uses a five-class taxonomy of hallucinations spanning four languages: Chinese, English, French, and Italian. The shared task generated strong interest in the NLP community worldwide, with 27 teams contributing 600+ system submissions.
The best systems achieve average scores of 0.58 in character-level correlation, 0.46 in label-conditioned correlation, and 0.51 in intersection-over-union (IoU) across four languages, outperforming the baselines by 30–40 points.

% UNCOMMENT IN CAM. READY
\begin{comment}
    \begin{center}
      \begin{minipage}{\linewidth}
        \centering
        \raisebox{-0.2\height}{\includegraphics[width=1em]{figs/logos/github-mark.png}}%
        \hspace{0.5em}%
        {\small\texttt{\href{https://github.com/Helsinki-NLP/shroom-vision}{\tt Helsinki-NLP/SHROOM-VISION}}\hphantom{12}}
      \end{minipage}
      \begin{minipage}{\linewidth}
        \centering
        \raisebox{-0.2\height}{\includegraphics[width=1.2em]{latex/figs/logos/website-vector.pdf}}%
        \hspace{0.5em}%
        {\small\texttt{\href{https://helsinki-nlp.github.io/shroom/2026}{\tt SHROOM-Series/SHROOM-VISIONS}}}
      \end{minipage}
    \end{center}
    \end{comment}

\end{abstract}

\section{``Begin at the beginning'': Introduction}
\label{sec:introduction}

Hallucinations---outputs that are fluent and plausible but factually incorrect or ungrounded in the input---in Large Vision-and-Language Models (LVLMs) pose distinct reliability challenges beyond those observed in text-only generation \citep{huang_survey_2025}. The critical difference lies in an additional grounding requirement: LVLM outputs must remain faithful not only to world knowledge and linguistic coherence, but to the specific visual content of an accompanying image. This gives rise to failure modes unique to the multimodal setting, including object fabrication, entity misdescription, misreading of on-image text, and miscounting of visible items \citep{liu2024survey}.

Complicating matters further, existing benchmarks on hallucination detection depend on outputs from a small, fixed set of models. Whether these models are used as the source of faulty generations to annotate \citep{ravichander-etal-2025-halogen} or as generators of synthetic hallucinated data \citep{muhlgay-etal-2024-generating}, such reliance makes it difficult to disentangle progress in hallucination understanding from overfitting to the idiosyncrasies of specific systems. Given the rapid replacement cycle of language models, benchmarks built this way risk losing their diagnostic value shortly after release \citep{laskar-etal-2023-systematic, Liu2025-df}. A further practical obstacle is coverage. Naively sampling model outputs yields a sparse and skewed representation of rarer hallucination types, whereas more targeted sampling strategies (e.g., LLM-judge to preselect candidates) shift the problem to the judge's reliability.

\begin{figure}
    \centering
    \includegraphics[width=0.55\linewidth]{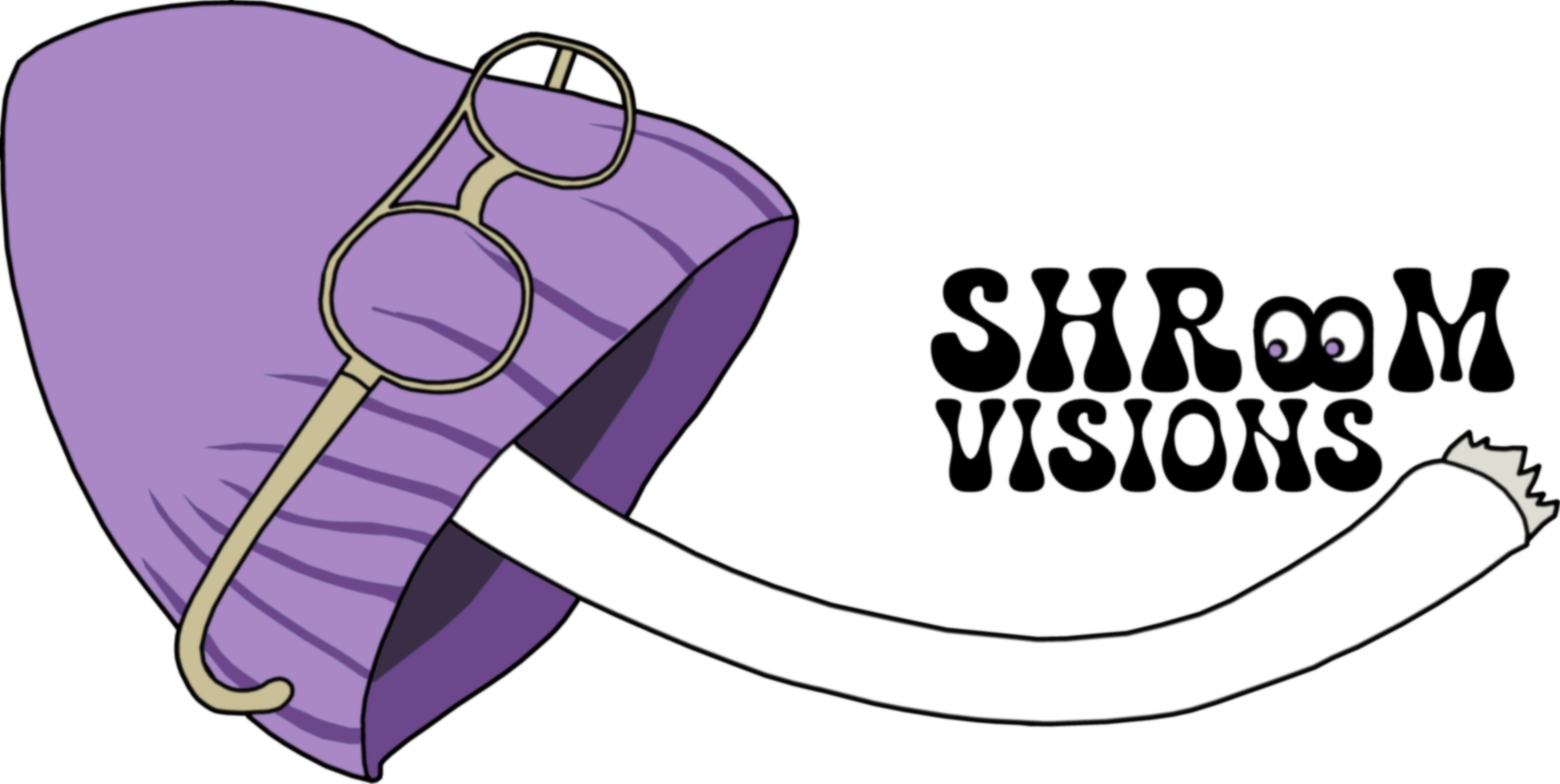}
    \caption{Shroom-Visions 2026 Shared Task logo.}
    \label{fig:logo}
\end{figure}

The SHROOM-Visions shared task is organized around the recently released SHEEP (a Set of Human-written and Electronic Erroneous Productions) dataset \citep{mickus2026humans}, a multilingual, large-scale benchmark that addresses these structural limitations using human-written hallucinations with deliberately inserted, labeled errors, providing a model-independent alternative to LVLM-generated data.
The dataset spans 20{,}000 samples across four languages (Chinese, English, French, and Italian), combining outputs from five LVLMs (sampled randomly and via LLM-judge-assisted preselection) with 1{,}600 human-written items. 
Instances are annotated at the span level using a five-way hallucination taxonomy (\textit{invention}, \textit{mischaracterization}, \textit{misreading}, \textit{miscounting}, and \textit{other}). Crucially, the authors show that human-written samples achieve higher inter-annotator agreement, allow precise control over the distribution of hallucination types, and produce detector rankings that correlate more strongly with LVLM output rankings than cross-model comparisons do. Hence positioning human-authored hallucinations as viable, durable substitutes for constantly refreshed, model-derived benchmarks.

Building on this resource, SHROOM-Visions delves into the multimodal domain by evaluating detection systems on both LVLM-generated and human-written data. This setup introduces model-independence as a key metric, testing whether performance on model-derived benchmarks generalizes to hallucinations untied to specific generators.

\section{Down the rabbit hole: Related works}
\label{sec:related_work}

Hallucination in vision-and-language models is generally understood as generated content that is unsupported by, or contradicts, the visual input \citep{liu2024survey, bai2025survey}. Early taxonomies organized LVLM hallucinations along a coarse object–attribute–relation triad \citep{liu2024survey}, later refined by work introducing event-level fabrications \citep{jiang2024haleval} and finer-grained distinctions such as miscounting, text misreading, and identity incongruity \citep{rani2024visual}. Existing benchmarks for LVLM hallucination detection broadly fall into three paradigms: caption-centric approaches that assess the factual accuracy of the descriptions \citep{rohrbach2018object, petryk2024aloha}; discriminative approaches that reframe evaluation as classification \citep{shekhar2017foil, li2023evaluating, lovenia2024negative}; and hybrid approaches combining generative and verification tasks, including work targeting entangled visual illusions \citep{guan2024hallusionbench}, input perturbation \citep{ding2024hallupi, saito2026haldec}, automated benchmark construction \citep{wu2024autohallusion}, and fine-grained failure classification \citep{yebin2024beaf, wang2024amber}. Closest to the present setting are M-HalDetect \citep{gunjal2024detecting} and HalLoc \citep{park2025halloc}, which provide fine-grained and token-level hallucination localization, respectively.

A recurring limitation is that resources are built around English outputs from a single or a small handful of LVLMs. This creates two compounding problems: benchmarks inherit the idiosyncrasies of their source models and lose diagnostic value as those models are superseded \citep{laskar-etal-2023-systematic, Liu2025-df}, a concern also raised in LLM-centric factuality work \citep{ravichander-etal-2025-halogen, vazquez-etal-2025-semeval}; and even synthetic-generation approaches intended to sidestep this issue \citep{muhlgay-etal-2024-generating} still depend on an LLM as the underlying generator, reintroducing model-specific bias. \citet{mickus2026humans}  introduce SHEEP, a multilingual, span-level dataset pairing human-written text with LVLM outputs. They demonstrate that human-authored data yields higher annotation agreement and correlates strongly with model performance, validating it as a robust, model-independent benchmark.

SHROOM-Visions adopts the SHEEP dataset of \citet{mickus2026humans}, extending the SHROOM hallucination detection shared tasks to the multimodal domain. Prior iterations addressed monolingual English NLG \citep[SHROOM;][]{mickus-etal-2024-semeval}, multilingual Q\&A \citep[Mu-SHROOM;][]{vazquez-etal-2025-semeval}, and scientific text \citep[SHROOM-CAP;][]{sinha-etal-2025-shroom}; SHROOM-Visions extends this by bridging prior challenges whose interests were solely in multilingual \citep{li-etal-2025-overview-scihal25,mubarak-etal-2025-islamiceval,nguyen2026dsc2025vihalluchallenge,10.1007/978-3-031-71908-0_3} or multimodal settings (\citealp{10.1145/3746027.3762048}; a.o.). A key focus of this edition is to develop benchmarking practices that are robust to generative model obsolescence. By treating human-written hallucinations as a first-class evaluation source, SHROOM-Visions distinguishes its design from prior resources tied to specific model generations.

\section{``Oh dear! Oh dear! I shall be late!'': Task timeline and organization}
\label{sec:dataset}

\begin{figure}[!t]
    \centering
\begin{subfigure}{\linewidth}
\centering
\includegraphics[width=0.65\linewidth]{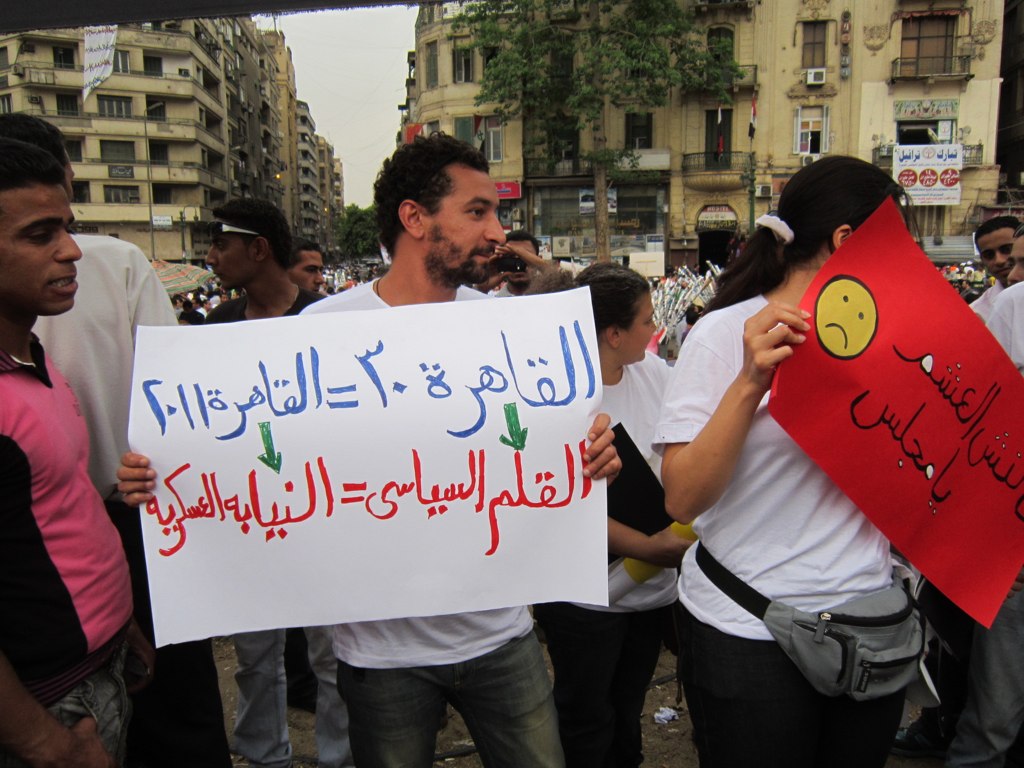}
\caption{Associated image from HaloQuest \citep{10.1007/978-3-031-72980-5_17}.}
\end{subfigure}
\begin{subfigure}{\linewidth}
%\begin{figure}
%    \centering
\begin{tikzpicture}
\node[anchor=north west] (box) at (0, 0) {
    \begin{tcolorbox}[colframe=black!75, colback=yellow!10, width=0.92\columnwidth]
    \scriptsize
    % { 
    \{\\
\null~\textbf{"id"}: "train-en-9823",\\
\null~\textbf{"split"}: "train",\\
\null~\textbf{"language"}: "en",\\

\null~\textbf{"prompt"}: "What does the blue banner about?",\\
\null~\textbf{"image"}: "5104\_5769583198\_4926a17525\_o.jpg",\\

\null~\textbf{"response"}:
%"The banner with blue writing appears to be a demonstration or protest sign, expressing a political or social message. However, without specific knowledge of the context in which the photo was taken, it is difficult to provide a precise translation or interpretation of the text on the banner. The writing is in Arabic, and it includes mathematical symbols such as a fraction and an equation with a negative sign, which might suggest a critical stance towards the system or policies being protested. The yellow smiley face on the red banner adds a contrasting element to the sign, potentially symbolizing positive values or demands. If you have specific questions about the banner, its message, or the context of the protest, please provide more information.",
%"The banner with blue writing appears to be a demonstration or protest sign, expressing a political or social message. However, without specific knowledge of the context in which the photo was taken, it is difficult to provide a precise translation or interpretation of the text on the banner. [The writing is in Arabic, and it [includes mathem]atical symbols [such as a fraction ]and an equation with a negative sign,] which might suggest a critical stance towards the system or policies being protested. The [yellow] smiley face on the [red banner] adds a contrasting element to the sign, [potentially symbolizing ][positive values or ]demands. If you have specific questions about the banner, its message, or the context of the protest, please provide more information.",
"The banner with blue writing appears to be a demonstration or protest sign, expressing a political or social message. However, without specific knowledge of the context in which the photo was taken, it is difficult to provide a precise translation or interpretation of the text on the banner. \hlA{The writing is in Arabic, and it }\hlB{includes mathem}\hlA{atical symbols }\hlB{such as a fraction }\hlA{and an equation with a negative sign,} which might suggest a critical stance towards the system or policies being protested. The \hlA{yellow} smiley face on the \hlA{red banner} adds a contrasting element to the sign, \hlA{potentially symbolizing }\hlA{positive values or }demands. If you have specific questions about the banner, its message, or the context of the protest, please provide more information.",

%\null~"comments": \{\\
%\null~~"yMF": "The annotator doesn't know enough Arabic to be sure, and the smiley is rather nonsmiling (and not contrasting or symbolizing positive values)",\\
%\null~~"amT": "Unable to tell",\\
%\null~~"gTA": "Does not mention that there is no blue banner"\\
%\null~\},

%\null~"raw\_annots": \{\\
%\null~~"yMF": "335:412;510:516;547:558;603:630|E;B;B;B",\\ 
%\null~~"amT": "326:411;510:517;604:613|E;B;B",\\
%\null~~"gTA": "363:378;391:411;510:516;604:630|C;C;B;B"\\
%\null~\},

\null~\textbf{"labels"}: [\{"start": 326, "prob": 0.3333, "label": "E", "end": 335\}, \{"start": 335, "prob": 0.6667, "label": "E", "end": 411\}, \{"start": 364, "prob": 0.3333, "label": "C", "end": 378\}, \{"start": 391, "prob": 0.3333, "label": "C", "end": 411\}, \{"start": 510, "prob": 1, "label": "B", "end": 516\}, \{"start": 547, "prob": 0.3333, "label": "B", "end": 558\}, \{"start": 604, "prob": 1, "label": "B", "end": 612\}, \{"start": 612, "prob": 0.6667, "label": "B", "end": 630\}],

%\null~"metadata": \{"MAP\_prelabel": "C", "orig\_dataset": "haloquest", "model": %"llava", "strategy": "MAP"\}\\
\}
    \end{tcolorbox}
};
\end{tikzpicture}
%\caption{Example datapoint from  English split of the CAP dataset \citep{gamba2025confabulationsaclpublicationscap}.}
 %   \label{fig:cap-instance}
%\end{figure}
\vspace{-0.75em}
\subcaption{Annotated datapoint.
}
\end{subfigure}
    \caption{Sample data point. The original data contain additional meta data including raw annotation and sampling strategy. Red highlight in the ``response'' corresponds to the marked hallucination span, darker shade denotes overlapped spans.}
    \label{fig:datapoint}
\end{figure}

\subsection{Task definition}
\label{sec:task}
Given an image, a prompt and a response, participants must identify the specific character spans that correspond to hallucinations and assign each span to a category (example datapoint in \Cref{fig:datapoint}).

The taxonomy of hallucinations consists of 
\textbf{(1)~\texttt{Invention}}, where entities, objects, properties, or events are mentioned despite not being present in the image; 
\textbf{(2)~\texttt{Mischaracterization}}, where visible content is described incorrectly; 
\textbf{(3)~\texttt{OCR Problem}}, for errors caused by misreading text; 
\textbf{(4)~\texttt{Miscounting}}, where item quantities are reported inaccurately; 
and \textbf{(5)~\texttt{Other}}, hallucinations that do not fit the preceding categories. 
For every character in the response, participants have to estimate the probability that it belongs to a hallucinated span and determine the corresponding hallucination class for each detected span. We encouraged flexible and innovative submissions, allowing participants to use any approach, including LLM-based methods and external resources, and to submit systems for any subset of the four supported languages: Chinese, English, French, and Italian.

\subsection{Data}
\label{sec:data}
We use the SHEEP dataset \citep{mickus2026humans},\footnote{Data available at \url{https://helsinki-nlp.github.io/shroom/2026}, released under a CC-BY-NC license.} which was constructed specifically to support model-independent benchmarking of hallucination detection in LVLMs. It contains 20{,}000 samples evenly distributed across four  languages: Chinese, English, French, and Italian. 

\paragraph{Source material.} 
In the SHEEP dataset, images and questions originate from two existing English-language resources: \textsc{HaloQuest} \citep{10.1007/978-3-031-72980-5_17}, a visual question answering dataset designed to elicit hallucinations through visually ambiguous images and false-premise questions, and \textsc{VISaGE} \citep{frank-allaway-2025-visage}, which pairs images of objects with atypical properties (e.g., a three-legged cat) with questions targeting those properties. Only non-synthetic images from both sources were retained. Prompts were machine-translated into French and Italian using NLLB-200 (3.3B parameters; \citealp{nllbteam2022languageleftbehindscaling}), and into Chinese using Qwen3-8B \citep{qwen3technicalreport}.

\paragraph{Response generation.} LVLMs responses were sampled from four 8B-scale open models: InternVL3, MiniCPM-V~4.5, Llava-NeXT, Qwen3-VL, and one larger model Gemma3-27B, all using $\tau=0.7$, a 512-token generation cap, and five random seeds per input.

\paragraph{Sampling strategies.} 
The annotation set combines three complementary sampling strategies: random sampling of LVLM outputs, model-assisted pre-selection (MAP) using an LLM-judge to obtain a label-balanced subset, and human-authored examples that were translated and post-edited to ensure multilingual coverage.

\paragraph{Data splits.}
We partition the dataset into training and test splits. The MAP and Random strategies contribute to both splits, while human-written items appear exclusively in the test set. 
% providing a model-independent evaluation set. 
Table~\ref{tab:splits} summarizes the composition of each split by language and sampling strategy.

\begin{table}[t]
\centering
\small
\begin{tabular}{@{}lrrrr@{}}
\toprule
 & \textbf{EN} & \textbf{FR} & \textbf{IT} & \textbf{ZH} \\
\midrule
Train -- Random & 1{,}918 & 1{,}925 & 1{,}936 & 1{,}948 \\
Train -- Silver (MAP)    & 1{,}881 & 1{,}842 & 1{,}810 & 1{,}842 \\
Test -- Random  & 382     & 375     & 364     & 352     \\
Test -- Silver (MAP)     & 419     & 458     & 490     & 458     \\
Test -- Human   & 400     & 400     & 400     & 400     \\
\bottomrule
\end{tabular}
\caption{Statistics per language, split, and sampling strategy from SHEEP dataset \cite{mickus2026humans}.}
\label{tab:splits}
\end{table}

\subsection{Timeline}
The shared task began with a \textbf{Training Phase} (from May 10, 2026) providing a multilingual training set of $\sim$15.2K annotated samples; a participant kit with a scoring program, format checker, and two baseline systems\footnote{One constant baseline that marks nothing as hallucination; and one neural baseline is based on HalluShift++ \cite{nath2025hallushift++} implementation.}  and a submission platform (See Appendix, \cref{fig:submission-platform}).
The \textbf{Evaluation Phase} (ending July 31, 2026) involved predictions on a hidden test set of 4.8K samples (1.2K per language). Test labels were withheld to preserve integrity, although diagnostic analyses were available upon request. 
Finally, the \textbf{Post-Evaluation Phase} concluded with system description papers submissions by August 10, 2026.

\section{``What is the largest number that you know?'': Evaluation metrics}
\label{sec:evaluation}
Following the evaluation methodology established in a prior edition of the SHROOM series \citep{vazquez-etal-2025-semeval}, participant systems are evaluated at the \textit{character level} against the multi-annotator gold labels. For each item $d$ of character length $L_d$, systems are expected to output, for every character $c \in \{1, \dots, L_d\}$, a probability $Pr(c)$ that the character belongs to a hallucinated span; we denote the resulting system vector $\hat{r} \in \mathbb{R}^{L_d}$, with $\hat{r}_c = Pr(c)$. The corresponding gold vector $r \in \mathbb{R}^{L_d}$ is derived from the multi-annotator span labels, with $r_c$ aggregating annotator probabilities at each character. Participants are ranked according to three primary metrics computed from $r$ and $\hat{r}$, computed separately for each of the four target languages.

%Character-level
\paragraph{Unlabeled Correlation (Corr).}
We compute the Spearman correlation between the gold vector $r$ and the system vector $\hat{r}$:
\begin{equation}
\text{Corr}(d) = \rho(r, \hat{r})
\end{equation}
When either vector is constant (i.e., no gold or no predicted hallucination), $\rho$ is undefined and we instead score exact match on the presence versus absence of hallucination. Item scores are averaged over the test set.

\begin{table*}[t!]
\centering
\small
\setlength{\tabcolsep}{8pt}
\renewcommand{\arraystretch}{1.1}
\resizebox{\textwidth}{!}{
\begin{tabular}{l c p{14cm}}
\toprule
\textbf{Team} & \textbf{Lang} & \textbf{Description} \\
\midrule
Bit-by-bit \cite{bit-by-bit}          & All & QLoRA-finetuned Qwen2.5-VL-7B with teacher-forced token scoring and hallucination probability heads.\\
\rowcolor{gray!15}
Bubus \cite{bubus}              & EN             & Fusion of teacher-forced Qwen3-VL-32B and InternVL3-38B with hidden-state logistic regression and claim verification. \\
CPS Lab \cite{cps-lab}            & All & Linear probe on frozen Qwen3-VL-8B-Instruct with soft-label multi-class hallucination span detection.\\
\rowcolor{gray!15}
DSR                 & All & Character-level XLM-RoBERTa–SigLIP ensemble for hallucination span and category prediction.\\
HalluVision \cite{halluvision}        & All & LoRA-finetuned Qwen2-VL-2B trained on 15K image-text samples for hallucination prediction.\\
\rowcolor{gray!15}
HalluciNauts \cite{hallucinauts}       & EN             & Output-only soft token tagger trained on English annotations for hallucination span detection.\\
IrumS \cite{irums}              & EN             & Claude Haiku 4.5 vision verifier with structured hallucination span extraction and rule-based character alignment.\\
\rowcolor{gray!15}
L1cache  \cite{l1cache}           & All & Text-only multilingual XLM-RoBERTa span detector with clean gating and ensembling. \\
NSU TEAM            & EN             &  VLM-based hallucination span extraction with self-consistency calibration and optional claim verification.\\
\rowcolor{gray!15}
Risotto\_AI\_Funghi & All & Committee of VLM judges and activation probes with character-level confidence aggregation.\\
SKstars \cite{skstars}            & EN             & Ensemble of zero-shot Qwen2.5-VL-72B and LoRA-finetuned Qwen2.5-VL-7B. \\
\rowcolor{gray!15}
SIT \cite{sit}                & All & Ensemble of LoRA-finetuned Qwen3.5-4B and XLM-RoBERTa–SigLIP multimodal token tagger.\\
Scalar\_Nitk        & All & Multimodal mDeBERTa–SigLIP span classifier with language-specific LoRA adaptation and gated cross-attention.\\
\rowcolor{gray!15}
Sckwoky \cite{sckwoky}            & All & Language-specific character-level BIO taggers with probability scoring, clean gating, and span post-processing.\\
SmurfCat \cite{smurfcat}           & All & LoRA-fine-tuned Qwen3-VL-8B-Instruct model trained with varying combinations of real data, class-aware upsampling, heuristic injection, and LLM-assisted synthetic augmentation.\\
\rowcolor{gray!15}
%Testing             & All & \\
TÜRKSAT \cite{turksat}            & All & Family of multimodal hallucination detection systems based on LLM-judge ensembles and LoRA-fine-tuned Qwen VL token classifiers, leveraging character-level distillation, ensemble prediction, and language-specific thresholding.\\
USP \cite{usp}                 & All & Multilingual XLM-RoBERTa token-level span tagger with joint training, language-specific thresholds, and text-only inference.\\
\rowcolor{gray!15}
WINNAR749 \cite{winnar749}          & All & QLoRA-finetuned Qwen2.5-VL-3B-Instruct for per-character hallucination tagging.\\
%\rowcolor{orange!25}
%baseline            & All & \\
caml \cite{caml}                & All & Frozen Qwen3.5-4B layer-19 features with five-fold classifiers and language-specific span thresholds.\\
\rowcolor{gray!15}
champ \cite{champ}              & All & Gemma 4 31B VLM MLP probe with MuSHROOM transfer learning and LoRA.\\
dynamos \cite{dynamos}            & All & Fine-tuned XLM-RoBERTa token classifier with graded character-level span reconstruction.\\
\rowcolor{gray!15}
eye-be-am \cite{eye-be-am}          & All & Zero-shot prompting.\\
falcons             & ZH             & Qwen3-VL-8B token features with a frozen BiLSTM ensemble for hallucination span detection.\\
\rowcolor{gray!15}
hccl-jy \cite{hccl-jy}             & EN             & Constant baseline. \\
medusa              & EN             &  Fine-tuned Qwen3-VL-8B-Instruct  via LoRA to infer multiple samples to obtain per-character soft probability for hallucination. \\
\rowcolor{gray!15}
% smurfcat            & All & \\
vroom-vroom \cite{vroom-vroom}         & All     & LoRA-finetuned VLMs with inline hallucination tagging and confidence-based span extraction.\\
\midrule
\citet{non-task} & All & Non-system paper, discusses the effect of abstention rates on metrics.\\
\bottomrule
\end{tabular}}
\caption{Participating teams and the language tracks in which they submitted systems in alphabetical order.}
\label{tab:all_system_description}
\end{table*}

%Label-conditioned
\paragraph{Labeled Correlation ($\text{Corr}_\text{lbl}$).}
To capture whether systems correctly categorize hallucination labels, we compute the character-level correlation separately for each hallucination label present in either the gold or predicted annotations $\mathcal{L}(d)$.
We restrict the gold and predicted spans to those annotated with $\ell$, compute their character-level correlation $\rho(r_\ell, \hat{r}_\ell)$:
\begin{equation}
\text{Corr}_\text{lbl}(d) = \frac{1}{|\mathcal{L}(d)|} \sum_{\ell \in \mathcal{L}(d)} \rho(r_\ell, \hat{r}_\ell)
\end{equation}
Items with empty predictions or references get 1.0.

\paragraph{Intersection-over-Union (IoU).}
To assess span localization independently of confidence and category, we binarize $r$ and $\hat{r}$ into the character index sets they cover, $R(d) = \{c : r_c > 0\}$ and $\hat{R}(d) = \{c : \hat{r}_c > 0\}$, and compute:
\begin{equation}
\text{IoU}(d) = \frac{|R(d) \cap \hat{R}(d)|}{|R(d) \cup \hat{R}(d)|}
\end{equation}
Items with empty predictions or references get 1.0.

The three metrics are complementary: IoU rewards precise localization of hallucinated spans regardless of confidence or category; Corr rewards well-calibrated confidence profiles that mirror annotator agreement across the full response; and $\text{Corr}_\text{lbl}$ requires systems to correctly categorize the type of hallucination they detect.

\paragraph{Ranking.} Systems are ranked independently per language and metric. Participants could submit systems on a subset of the four languages, and rankings only include the corresponding language. We report performance separately on the three test subsets described in Section~\ref{sec:dataset} (Random, MAP, and Human-written), enabling analysis of whether performance obtained on LVLM-generated data generalizes to the model-independent, human-written subset -- the central empirical question motivating this shared task's design (Section~\ref{sec:introduction}).

\section{The caucus race: Participating systems}
\label{sec:participants}
% \lisa{add a section to spotlight people who participated in previous shared tasks / people who reused data from previous shared task specifically,}

\paragraph{Overall statistics.} 
Our shared task received participation from 27 teams with 623 submissions across the four target languages. 
On average, each language received 155.75 submissions, with English (EN) attracting the most (208), followed by Italian (IT) (141). French (FR) and Chinese (ZH) received 137 submissions each. 
% and present a more detail description for 5 most interesting submissions.

\subsection{Recurrent participants} 
It is encouraging to see several teams participate consistently across the SHROOM shared task series, reflecting sustained interest in hallucination detection research. 

In particular, \texttt{SmurfCat} has participated in all four editions of the shared task. \texttt{MEDUSA}, \texttt{NSU-AI}, and \texttt{Scalar\_Nitk} have each participated in three editions, while \texttt{AILS-NTUA}, \texttt{DeepPavlov}, \texttt{MALTO}, and \texttt{UMUTeam} have returned for two editions.
In total, \textbf{105 unique teams} have participated across the four editions of the SHROOM shared task.
Such continued engagement demonstrates the growing interest in hallucination detection and contributes to the development of a strong and active SHROOM community.

\subsection{Selected system descriptions}
% \textcolor{blue}{
27 teams submitted their systems during the evaluation phase, 20 teams wrote a system description paper; we also received 1 `shared non-task' paper discussing our choice of metrics.
Overall, we remark a wide variety of approaches, ranging from LoRA fine-tuning of LVLMs (e.g., \texttt{Bit-by-bit}, \citealp{bit-by-bit}; \texttt{HalluVision} \citealp{halluvision};  \texttt{WINNAR749}, \citealp{winnar749}), to token- and span-level classifiers built on pretrained multimodal or text-only encoders (e.g., \texttt{CPS Lab}, \citealp{cps-lab}; \texttt{L1cache}, \citealp{l1cache}; \texttt{USP}, \citealp{usp}), and ensemble or reasoning-based methods leveraging self-consistency, claim verification, and VLM judges (e.g., \texttt{Bubus}, \citealp{bubus}). %; \texttt{NSU TEAM}; \texttt{Risotto\_AI\_Funghi}).
An overview of all participating teams is provided in \Cref{tab:all_system_description}.
We also spotlight a few teams below, noteworthy for their performance and innovativeness.

\textbf{vroom-vroom} \cite{vroom-vroom}
%\textcolor{blue}{winning team, no submission yet}
fine-tunes vision-language models with LoRA for hallucination span detection. Given an image, question, and response, the models generate inline annotations marking hallucinated spans with category and confidence labels, which are then converted into character-level spans with associated probabilities.

\textbf{TÜRKSAT} \cite{turksat}
%\textcolor{blue}{winning team, no submission yet}
explores both inference-only and supervised approaches for character-level hallucination detection. Early systems use ensembles of multimodal LLM judges to assign soft character-level hallucination probabilities. Later variants distill these signals into LoRA-fine-tuned Qwen vision-language token classifiers. The submissions study different student backbones, curriculum learning with silver and teacher-generated labels, and ensemble strategies combining multiple distilled models with language-specific decision thresholds.

\textbf{Dynamos} \cite{dynamos} uses token-level and character-level hallucination detection models based on pretrained language and vision-language encoders. The systems evolve from fine-tuned XLM-RoBERTa token classifiers to multimodal Qwen2.5-VL models with LoRA/QLoRA adaptation, ensemble prediction, multi-class span classification heads, and Bayesian-tuned thresholds for graded span reconstruction.

\textbf{Champ} \cite{champ} explores probe-based hallucination detection using VLM representations. The systems range from official HalluShift++ baselines and MLP probes with transfer learning from Mu-SHROOM to Qwen-based detectors with LoRA adaptation, multitask span classification, and language-specific confidence calibration. Later variants incorporate deeper probes, vision fine-tuning, threshold tuning, abstention strategies, and span boundary refinement.

\textbf{Smurfcat} \cite{smurfcat}
%\textcolor{blue}{worth mentioning, participated in all 4 editions, no submission yet}
uses a LoRA-fine-tuned Qwen3-VL-8B-Instruct model to identify hallucinated spans from image-question-response inputs and output structured predictions with span text, category, and confidence scores. The team explored multiple training variants using real data alone and with additional augmentation strategies, including class-aware upsampling, heuristic injection, and LLM-assisted synthetic generation for rare categories.

% figure 3, min/max/mean all scores/lang
\begin{figure}[t!]
    \centering
    \includegraphics[width=\columnwidth, trim= 0 0cm 0 1.9cm, clip]{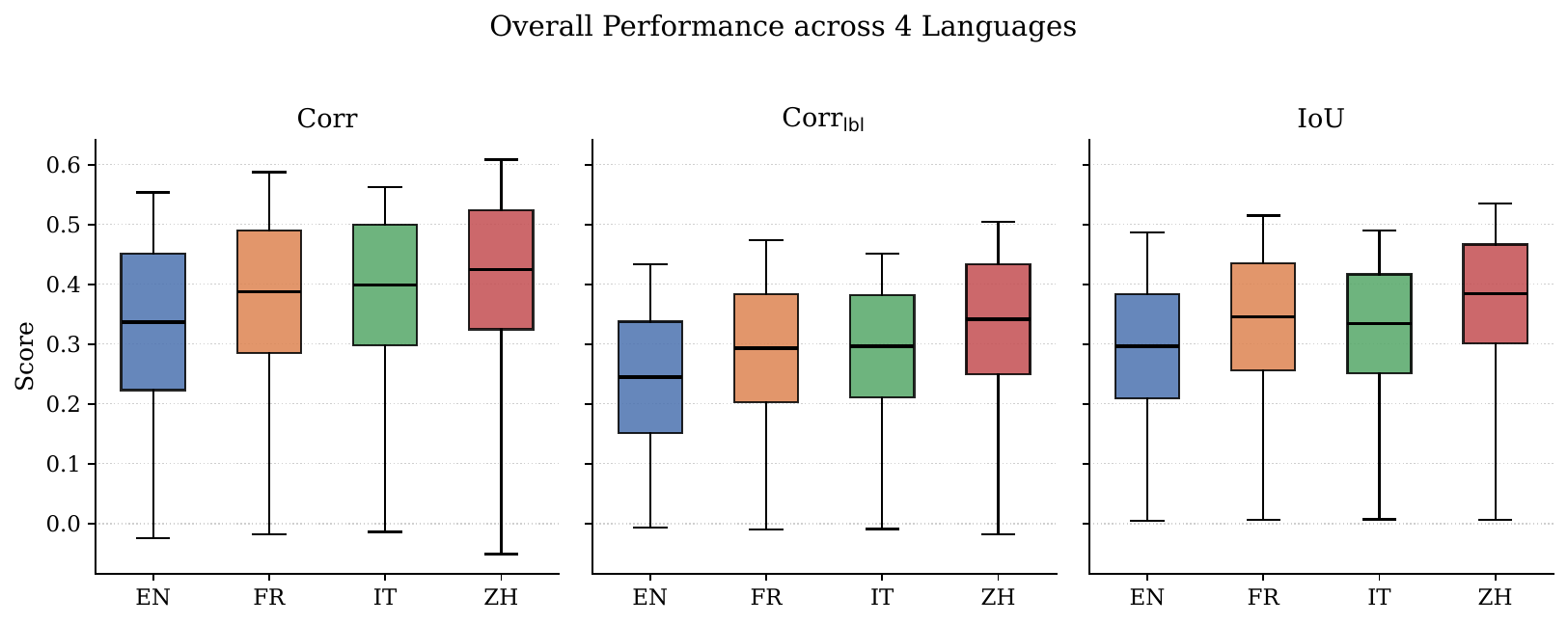}
    \caption{Overall Performance of all participating teams across 4 languages.}
    \label{fig:overall_stat}
\end{figure}
%=======================
\section{Through the looking glass: Results and analysis}

\Cref{fig:overall_stat} summarizes performance distributions, reporting the min, max, mean, and st. dev. per language. The full leaderboard is provided in \Cref{append:final-leaderboard}. 
All systems outperformed the baseline but absolute performance remains modest. Most mean scores fall below 0.4, with no system exceeding 0.6 in any language or metric. Chinese (ZH) is a partial exception, achieving the highest average performance slightly above 0.4 in Corr scores, whereas English (EN) proves most challenging. Despite these differences, the narrow score range and comparable variability across languages suggests a performance bottleneck that is largely independent of language.

% Figure 6, bootstrap rank
\begin{figure}[t!]
        % EN
    \centering 
    \begin{subfigure}{0.49\columnwidth}
        \centering
        \includegraphics[width=\linewidth,  trim={0.33cm 0.7cm 0.7cm 1cm}, clip]{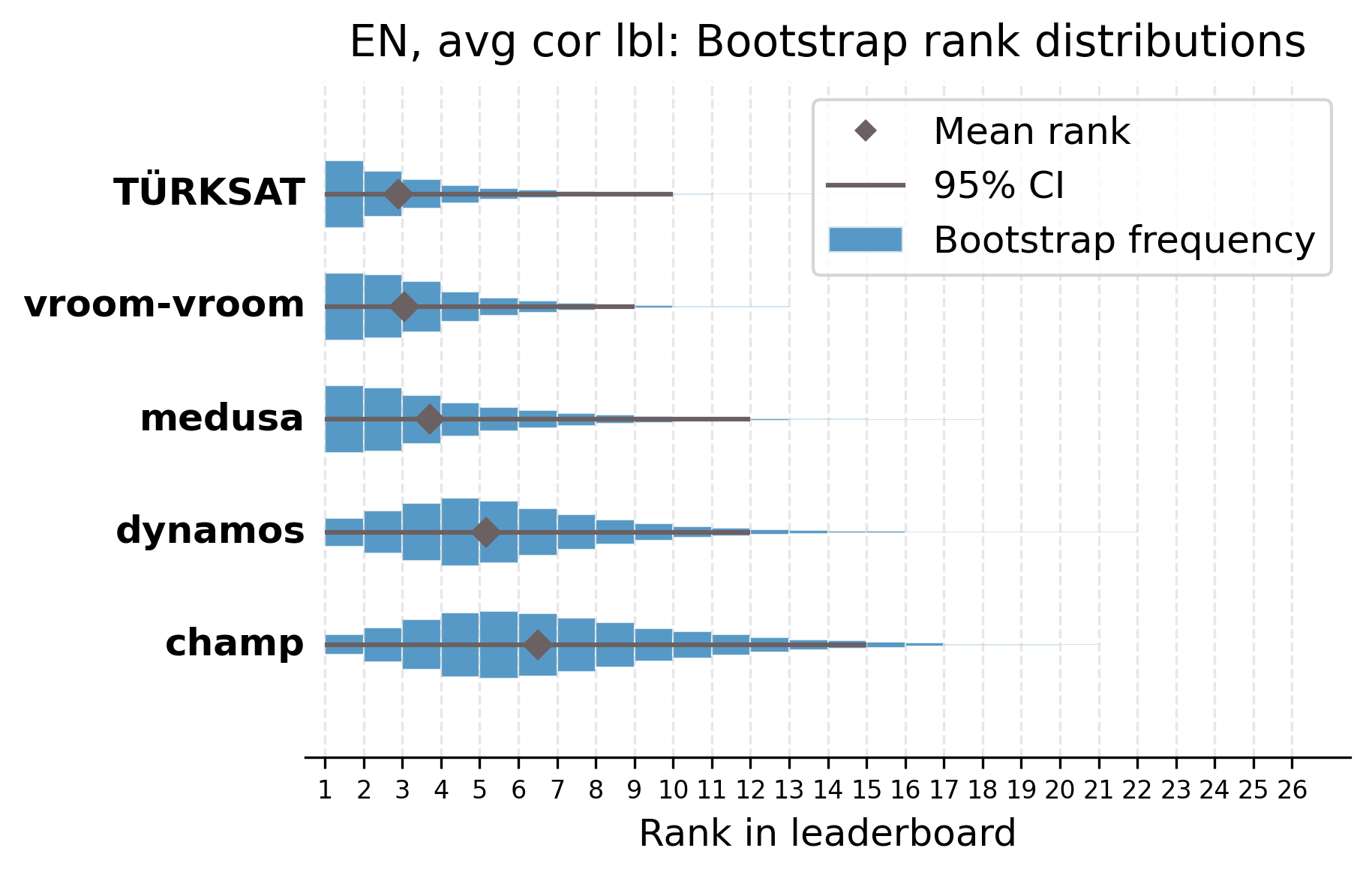}\caption{EN}
    \end{subfigure}
        %ZH
        \begin{subfigure}{0.49\columnwidth}
        \centering
        \includegraphics[width=\linewidth,  trim={0.5 0.7cm 0.7cm 1cm}, clip]{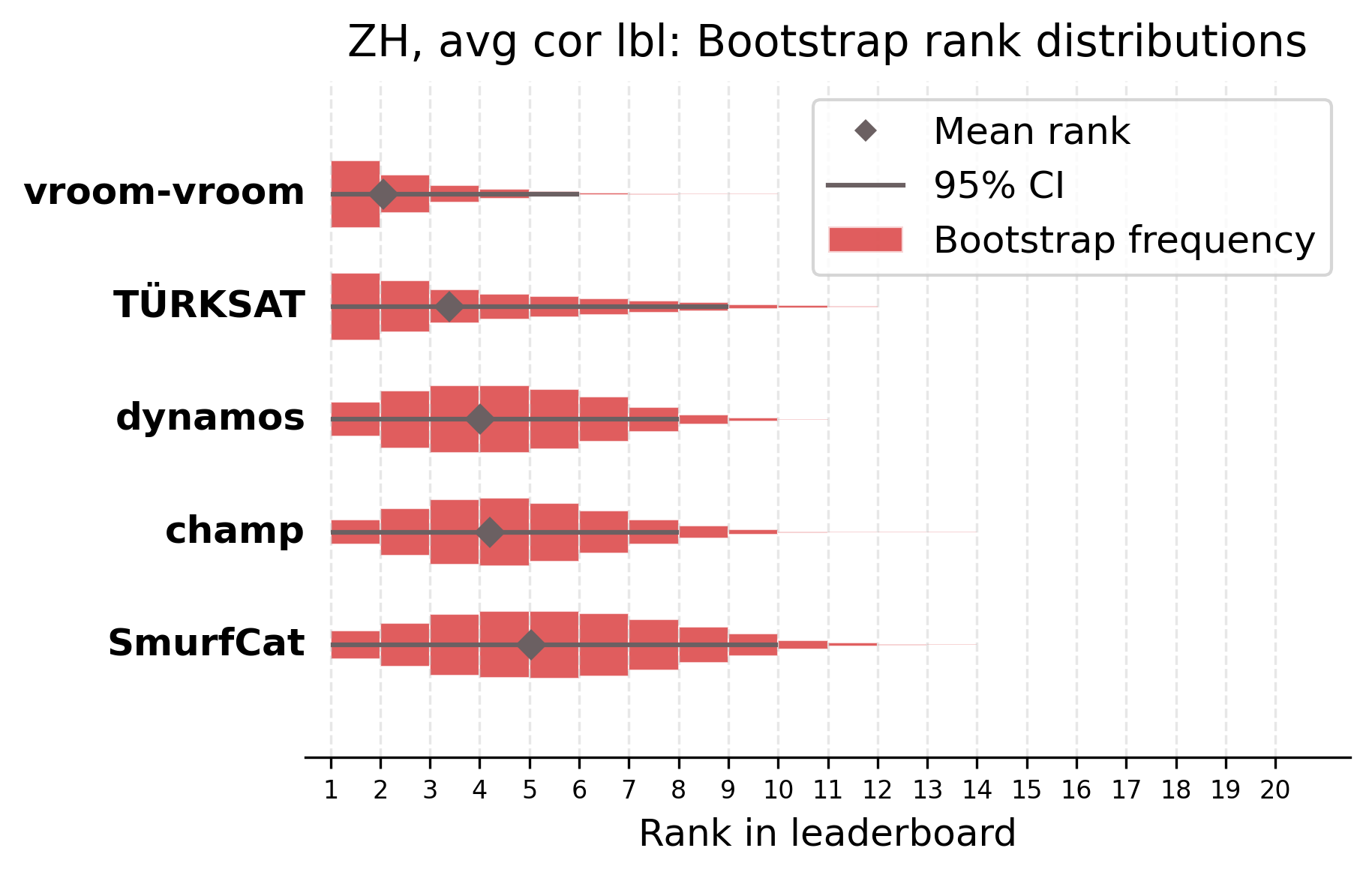}\caption{ZH}
    \end{subfigure}
        % IT
            \begin{subfigure}{0.49\columnwidth}
        \centering
        \includegraphics[width=\linewidth,  trim={0.4cm 0.7cm 0.7cm 1cm}, clip]{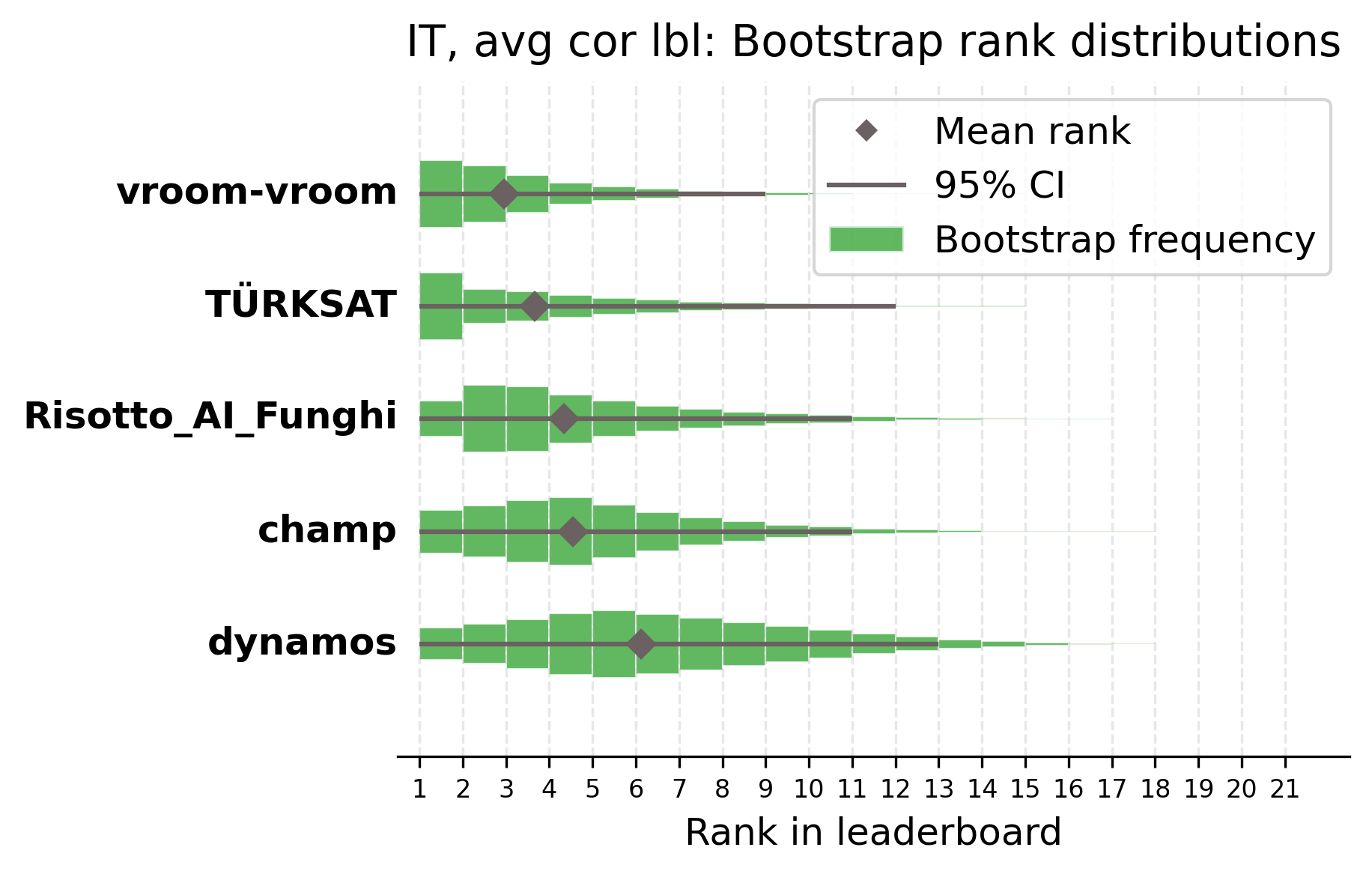}\caption{IT}
    \end{subfigure}
        % FR
            \begin{subfigure}{0.49\columnwidth}
        \centering
        \includegraphics[width=\linewidth,  trim={0.5 0.2cm 0.7cm 1cm}, clip]{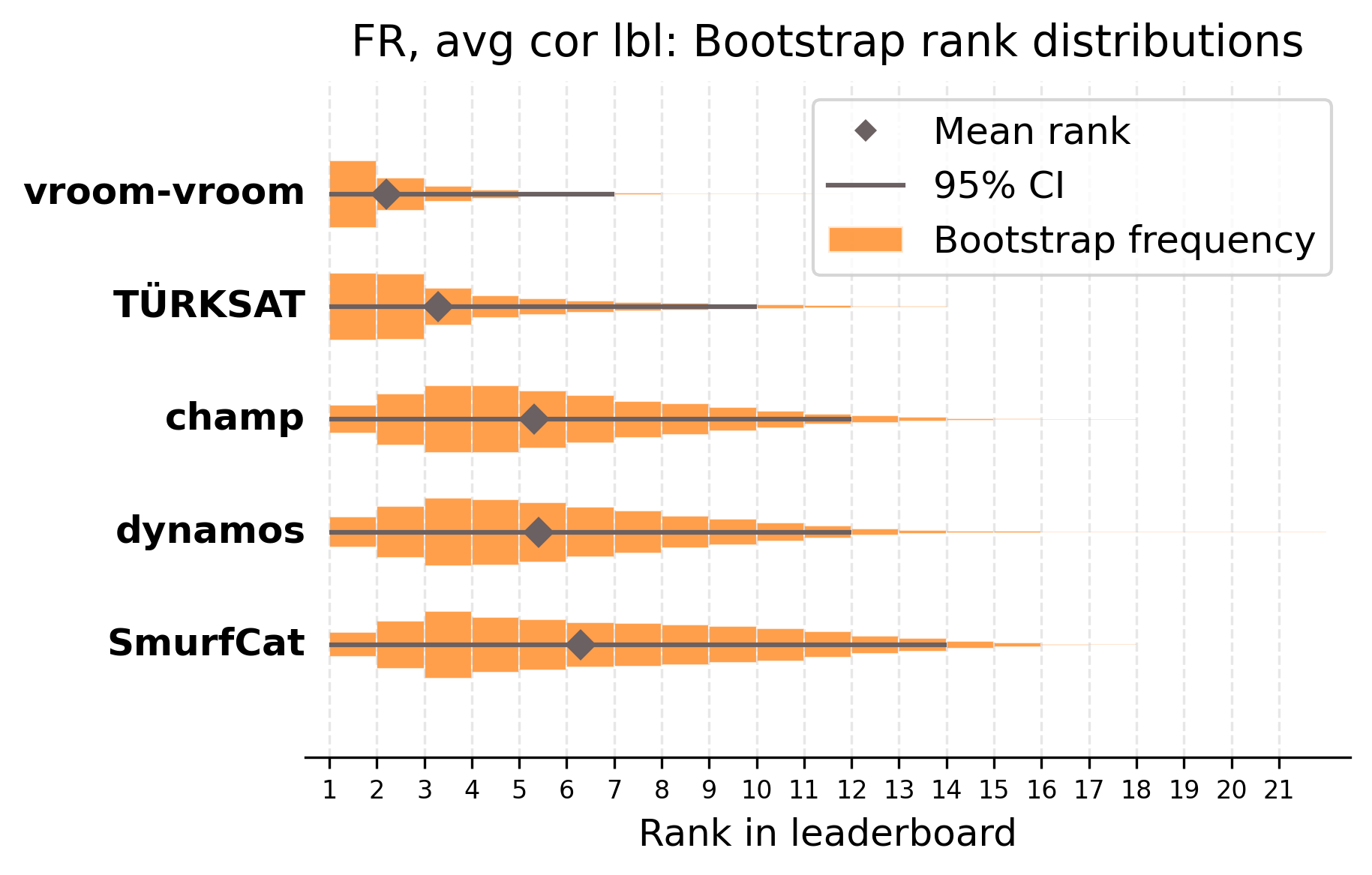}\caption{FR}
    \end{subfigure}
        \caption{Bootstrap rank distributions (25,000 samples) with means and confidence intervals at 95\% for the top-5 ranking submissions per language under $\text{Corr}_\text{lbl}$ metric.}
    \label{fig:top5_rank_dist}
\end{figure}

\subsection{Leaderboard stability \& Ranking uncertainty}
Contrasting with aggregate performance shown in \Cref{fig:overall_stat}, the bootstrap rank distributions in Figures \ref{fig:top5_rank_dist} and \ref{fig:all_rank-dist} show substantial ordering uncertainty. Although the point estimates suggest a clear leaderboard, resampling the test set frequently changes the relative positions of the leading systems. For instance, the top-team in EN (\texttt{TÜRKSAT}) has a mean rank of 2.9 but its 95\% rank interval spans positions 1 to 10. This instability is present in all systems, languages and metrics, with the top-five systems exhibiting rank intervals of up to 10 positions in ZH, 14 in IT and FR, and 15 in EN. 

In Tables \ref{tab:leaderboard_en}-\ref{tab:leaderboard_it} we show the full set of results and the rank distributions plots for all systems in \Cref{fig:all_rank-dist}. Notably, ranking stability varies by language. Chinese present the narrowest rank distributions among top teams (CIs spanning $\sim$5--6 positions), with \texttt{vroom-vroom}'s mean rank being 2.0 and a CI interval of [1, 7], and increasing uncertainty for systems down the leaderboard. In contrast, EN shows the widest spreads (with leading systems showing CIs up to 10+ positions) alongside the lowest scores. This inverse relationship suggests that when systems operate near a performance floor, stochastic effects dominate, rendering fine-grained rankings unreliable. Consequently, small leaderboard differences  should not be interpreted as definitive evidence of system superiority when performance differences are small.

\subsection{Strategy-level analysis}\label{sec:strategy-analysis}

We investigate whether systems trained on model-derived outputs generalize to generator-agnostic hallucinations by evaluating performance across the three SHEEP dataset partitions: random, human-written, and silver-label examples (see \Cref{sec:dataset}). We assess whether system performance and, in particular, system rankings depend on how hallucinations are constructed and annotated.

\begin{figure}[t!]
    \centering
    \includegraphics[width=0.99\linewidth, trim= {3cm 0 1cm 1cm}, clip]{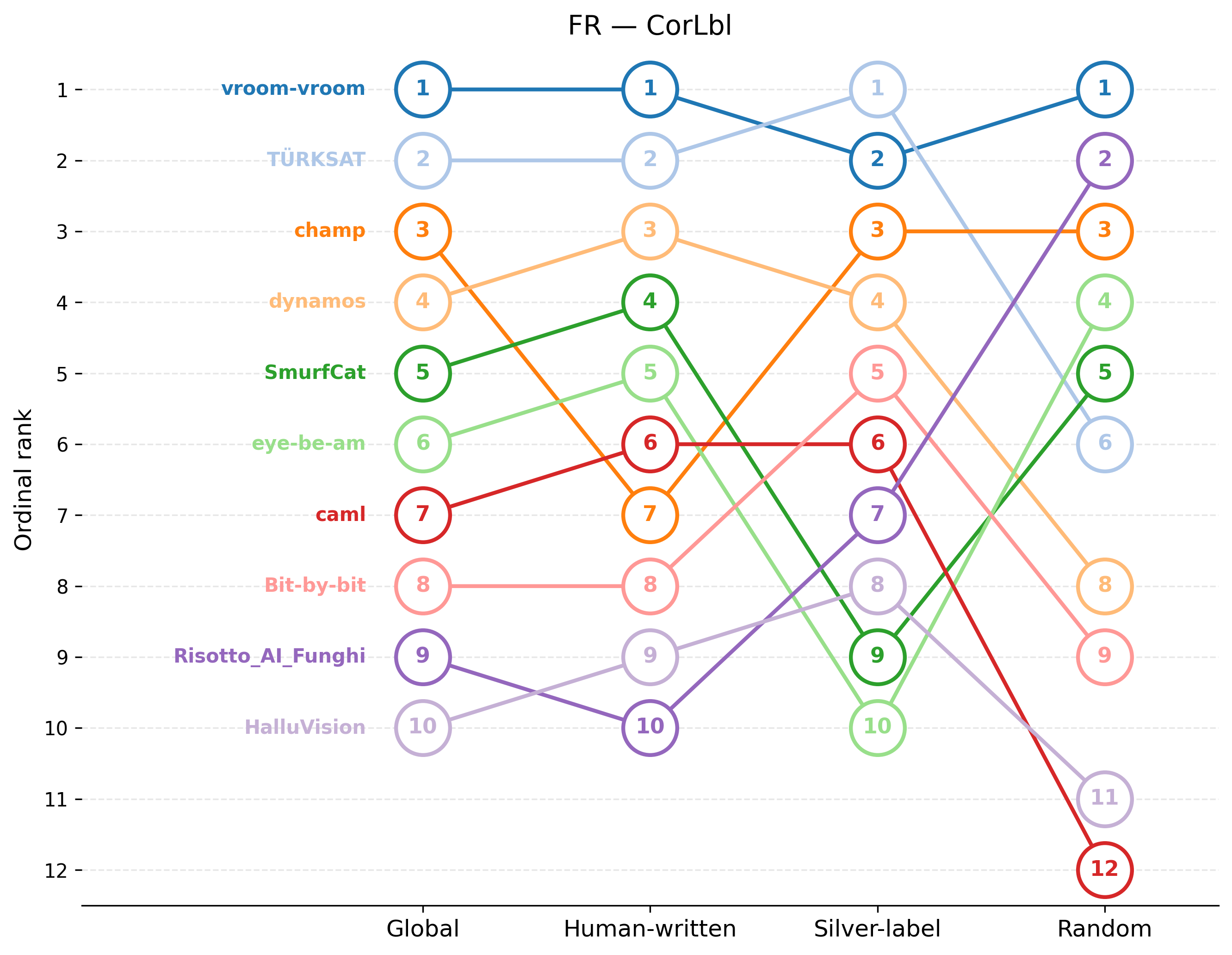}
    \caption{Rank flows across data partitions for the top-10 French systems on $\text{Corr}_\text{lbl}$. Each node represents a system's rank within a partition (General vs. Human/Silver/Random subsets). The variation in ranks illustrate that individual system positions can vary substantially with data sampling strategy.}
    \label{fig:rank-flow-fr-corlbl}
\end{figure}

The broad ordering of systems is relatively robust across strategies, but this robustness masks substantial variation of the positions of individual systems. In \Cref{fig:rank-stability-en-corlbl}, shows top-performing systems on EN $\text{Corr}_\text{lbl}$ that exhibit pronounced rank changes between the general leaderboard and individual strategy-specific subsets. Thus, agreement in the overall ordering does not imply ranking position stability across evaluation conditions. \Cref{tab:strategy-to-trategy} quantifies this consistency via Spearman correlations ($\rho_S$). In particular, human-written and silver-label examples produce highly similar rankings, with $\rho_S$ between $.861$ and $.973$. Correlations involving the random strategy are generally lower, indicating that randomly selected examples induce greater changes in system ordering. This effect is most pronounced for $\text{Corr}_\text{lbl}$ in Chinese, where the correlation is lowest. The full set of results are in \Cref{append:strategy-scores} (Tables \ref{tab:strategy_scores_en_cor}--\ref{tab:strategy_scores_fr_iou}), including all $\text{Corr}_\text{lbl}$ rank flow charts.
\begin{table}[t!]
    \centering
    \small
    \resizebox{0.8\columnwidth}{!}{
    \begin{tabular}{llrrrl}
        \toprule
        Lang. & Metric & R--H & R--S & H--S & Fried. $p$ \\
        \midrule
         & Corr                 & 0.802 & 0.933 & 0.905 & 0.0015 \\
         EN & Corr$_\text{lbl}$ & 0.684 & 0.864 & 0.861 & <.0001 \\
         & IoU                  & 0.828 & 0.941 & 0.920 & 0.0344 \\ 
         \midrule
         & Corr                 & 0.805 & 0.784 & 0.951 & <.0001 \\
         ZH & Corr$_\text{lbl}$ & 0.575 & 0.667 & 0.893 & <.0001 \\
         & IoU                  & 0.734 & 0.746 & 0.970 & <.0001 \\
        \midrule
        & Corr                  & 0.869 & 0.893 & 0.967 & 0.5220 \\
        IT & Corr$_\text{lbl}$  & 0.847 & 0.798 & 0.955 & 0.0037 \\
        & IoU                   & 0.910 & 0.911 & 0.973 & 0.5488 \\
        \midrule
        & Corr                  & 0.840 & 0.831 & 0.908 & 0.8187 \\
        FR & Corr$_\text{lbl}$  & 0.801 & 0.770 & 0.928 & 0.0006 \\
        & IoU                   & 0.840 & 0.860 & 0.895 & 0.5220 \\
        \bottomrule
    \end{tabular}}
 \caption{Rank correlation (Spearman $\rho$) between team rankings computed on Random (R), Human-written (H), and Silver-label (S) data partitions, per language and metric. Friedman $p$-values test for systematic raw scores differences across the three strategies computed via bootstrapping. Lower R--H and R--S values indicate that randomly sampling outputs without any prior signal can induce greater reordering of systems.}
    \label{tab:strategy-to-trategy}
\end{table}

Friedman tests computed on the raw scores provide a complementary view by testing for systematic differences across data selection strategies. They provide strong evidence of strategy-dependent absolute scores for all metrics in EN and ZH, as well as for $\text{Corr}_\text{lbl}$ in IT and FR. In contrast, IT and FR show no evidence of a systematic score difference for Corr and IoU. Hence, the effect of evaluation strategy can change the overall performance level depending on metric and language. Importantly, changes in absolute scores and changes in system ordering need not coincide. E.g., EN Corr exhibits a high $\rho_S(R-H)=.802$ despite strong evidence of a difference in score distributions (Fried. $p = .0015$). Conversely, ZH $\text{Corr}_\text{lbl}$ is sensitive to strategy at both  levels ($\rho_S(R-H)=.575$, $p < .0001$).

Ultimately, the data selection strategy influences absolute performance estimates more consistently than the coarse ordering of systems. Nevertheless, the ranking robustness is not uniform: randomly sampling substantially alters relative positions (especially for ZH $\text{Corr}_\text{lbl}$). These findings suggest that conclusions about the relative performance of hallucination detectors can depend on how evaluation examples are constructed, even when the broad ordering of systems remains similar.

\subsection{Cross-lingual performance}
\begin{figure}[t]
    \centering
    \includegraphics[width=0.8\linewidth]{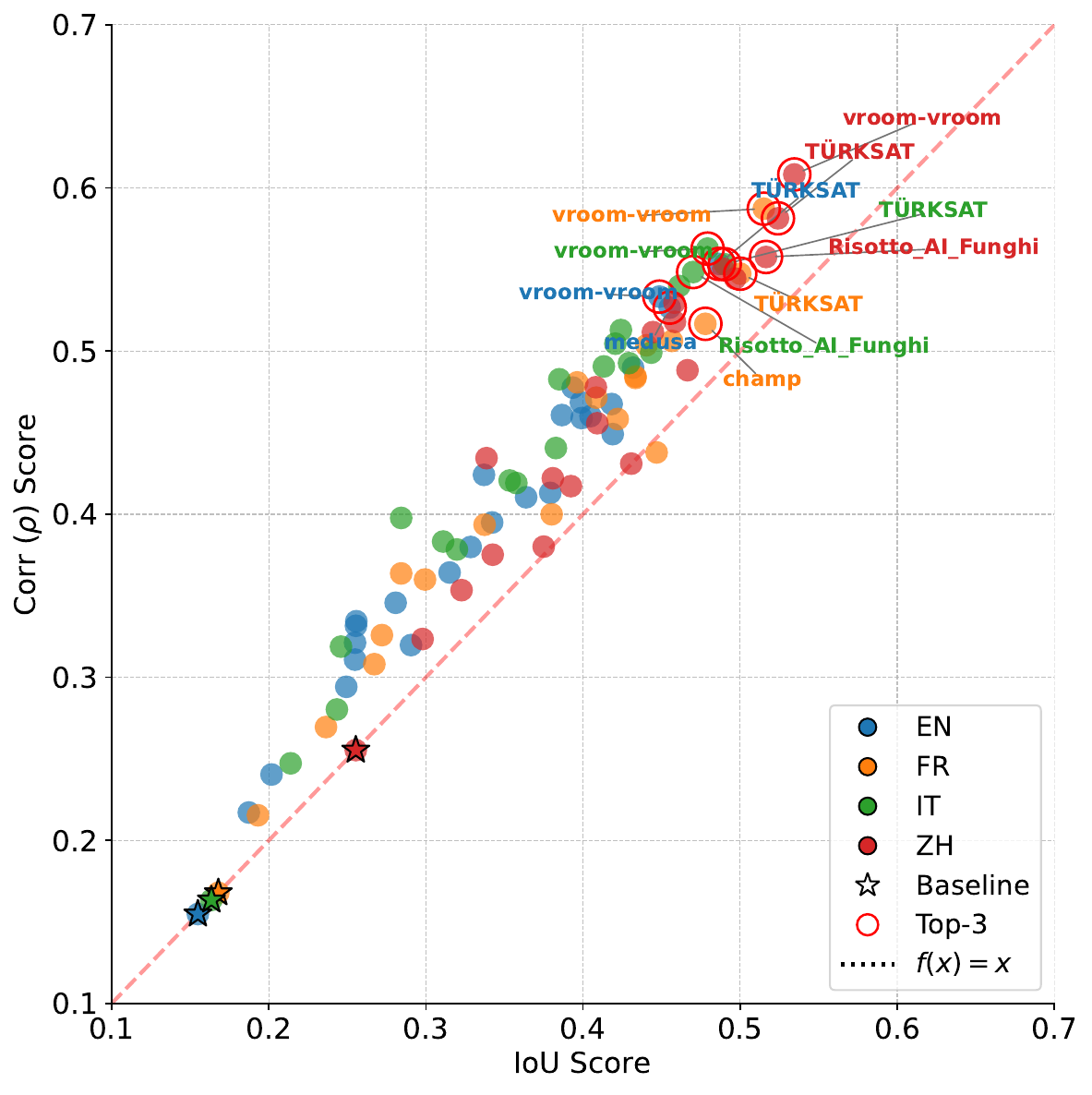}
    \caption{Scatter plot of IoU versus Corr scores for all participating teams, ranked by average IoU scores.}
    \label{fig:iou-corr}
\end{figure}

\begin{figure}[t!]
    \centering
    \includegraphics[width=0.8\linewidth]{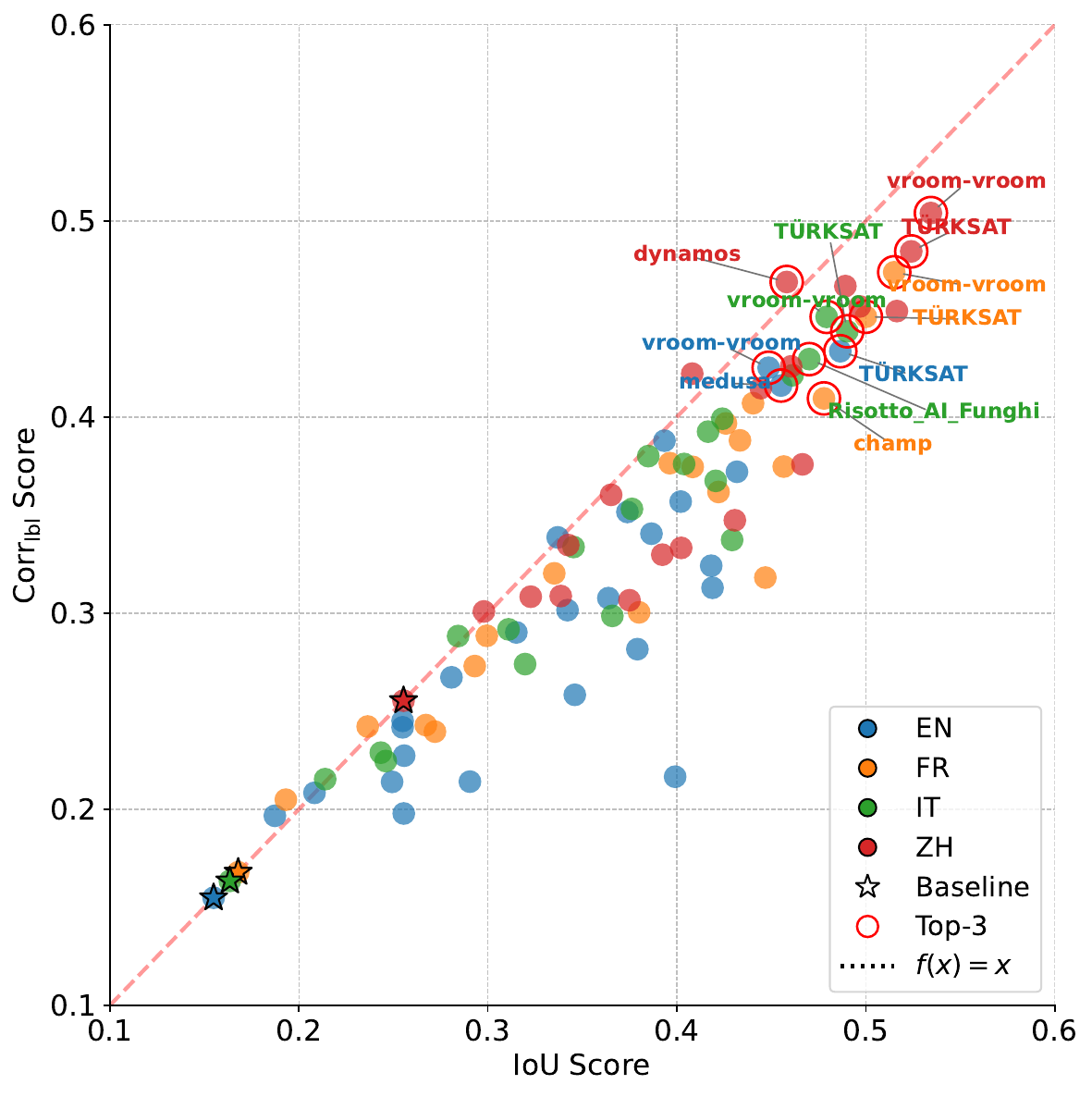}
    \caption{Scatter plot of IoU versus $\text{Corr}_\text{lbl}$ scores for all participating teams, ranked by average IoU scores.}
    \label{fig:iou-corr-label}
\end{figure}

Figures~\ref{fig:iou-corr} and \ref{fig:iou-corr-label} show the relation between IoU and the two correlation metrics (Corr and $\text{Corr}_\text{lbl}$) for all submitted systems. The top-3 systems in each language (with red circles) form a dense cluster in the upper-right corner and lower-performing systems are concentrated toward the lower-left. Hence, strong systems achieve accurate hallucination localization and reliable character-level ranking. The strong linear trends indicate a high degree of consistency between IoU and the two Corr metrics, while their systematic differences reflect the complementary aspects of system performance captured by each measure. Specifically, character-level correlation scores are generally higher than IoU. because Corr evaluates the full character-level probability profile, allowing a system to receive credit when its confidence broadly follows the annotated hallucination spans even if the predicted span boundaries are imperfect. IoU, is more sensitive to boundary errors since it binarizes the predictions. $\text{Corr}_\text{lbl}$ is generally lower than Corr because it imposes the additional requirement that hallucinated spans are assigned to the correct category.

% figure 6, scatter plot IoU vs corr

More precisely, we demonstrate the best systems from each team in language-split vision (\Cref{fig:1x4}). Across four languages, \texttt{TÜRKSAT} achieves the strongest performance in English ($\rho$=0.55, IoU=0.48). On the other hand, \texttt{vroom-vroom} leads the remaining three languages (FR: $\rho$=0.58, IoU=0.52, IT: $\rho$=0.56, IoU=0.48, and ZH: $\rho$=0.61, IoU=0.53) with particularly strong and well-balanced results on both metrics, while \texttt{TÜRKSAT} and \texttt{Risotto\_AI\_Funghi} remain close competitors in IT and ZH.

%    \item worth checking if possible lower perf on human data comes from OCR / miscounting spans (which are rare in the training data) / if items containing OCR / miscounting tend to statistically produce lower scores 

\section{Out of the rabbit hole: Conclusion}
SHROOM-Visions extends the %
%% UNCOMMENT IN CAM. READY
%\href{https://helsinki-nlp.github.io/shroom/}
{$^*$SHROOM shared task series} to span-level multilingual multi-class hallucination detection in LVLMs. With 27 teams contributing more than 620 submissions across four languages (EN, ZH, IT, FR), the task attracted diverse approaches, including multimodal prompting, probing, fine-tuning, synthetic data generation, and ensemble-based methods. The strongest systems substantially outperformed the baselines, reaching average scores of 0.58 in character-level correlation, 0.46 in label-conditioned correlation, and 0.51 in span-level IoU. Reliable fine-grained hallucination detection and classification remain difficult.

% figure 7, scatter plot IoU vs corr_lbl
Beyond the absolute performance, our results highlight two important evaluation challenges. First, leaderboard positions are far less certain than their point estimates suggest. We show that leading systems can occupy a wide range of ranks across resampled test sets. Therefore, reporting ranking uncertainty should be considered an essential complement to point-based rankings. Second, performance depends on how the benchmark is constructed. Although the broad system ranking is generally preserved across the three SHEEP dataset partitions, absolute scores can differ systematically, and individual systems can change substantially in relative position. Human-written and silver examples produce highly consistent rankings but random sampling induces larger volatility.

Ultimately, progress in multimodal hallucination detection should be assessed along two dimensions: detection accuracy and how reliably the evaluation distinguishes between systems. SHROOM-Visions  reveals considerable headroom in absolute performance and underscores the need for robust benchmarks that combine model-independent test data, fine-grained multilingual annotations, and explicit uncertainty reporting.

\section*{Building a house of cards: Limitations}

SHROOM-Visions shared task inherits limitations from the underlying dataset, including its restricted coverage of languages, domains, and LVLM outputs, which may not fully represent hallucination phenomena in broader real-world settings. The static benchmark also introduces potential risks of leakage and overfitting, although test annotations were kept private. 

%In addition, the main evaluation metrics used for ranking the participant systems focus on span overlap and probability agreement and do not explicitly assess abstention. Consequently, systems can receive non-zero correlation-based scores even when all annotations are empty, making the evaluation less sensitive to whether a system correctly recognizes hallucination-free responses. Future evaluations could incorporate explicit metrics for empty-instance detection and abstention alongside span-level performance.

%The shared task inherits limitations from the underlying dataset, including its restricted coverage of languages, domains, and LVLM outputs, which may not fully represent hallucination phenomena in broader real-world settings. The static benchmark also introduces potential risks of leakage and overfitting, although test annotations were kept private. 

A further limitation concerns datapoints without marked hallucinations, shown in \Cref{tab:stats_emptyannot}.  %empty annotations account for 15.5--25.5\% of test instances across languages, whereas systems produce empty predictions for 30.7--46.0\%. This gap highlights substantial variation in systems' abstention behavior, which is not explicitly captured by the span-overlap and probability-correlation metrics. In particular, correlation-based metrics can remain non-zero for instances with empty annotations, motivating explicit abstention or empty-instance metrics in future evaluations.
Empty annotations account for 15.5--25.5\% of test instances across languages, whereas systems produce empty predictions for 30.7--46.0\%. This gap highlights substantial variation in systems' abstention behavior, which is not explicitly captured by the span-overlap and probability-correlation metrics. For instances without annotated hallucinations, the evaluation defaults to an exact-match outcome when no span is available for comparison, which does not explicitly assess whether the system's abstention decision is appropriate. Future evaluations could therefore incorporate an explicit empty-instance or abstention metric.

%\section*{Acknowledgments}

%\clearpage
\bibliography{custom}
%\bibliography{custom}

%\newpage
%\clearpage
\appendix
%\section{}
%Talk about \Cref{tab:stats_emptyannot} and \Cref{fig:1x4}

%\begin{comment}
\section{The Mad hatter's tea party: Organizers' role}
\label{adx:chompers}
% Our long line of CHOMPS friendly people behind this edition of the SHROOM-CAP Shared task are as follows:
The team of SHROOM-Visions contributors behind this edition of the $^*$SHROOM series of shared tasks is as follows: 

\orgname{Raul Vazquez}: Shared Task Ideation/Creation: writing + analysis for the overview paper.

\orgname{Aman Sinha}: Writing and analysis for the overview paper; task advertisement; post-eval participant communication.

\orgname{Chuyuan Li}: Advertisement campaign lead; writing for the overview paper; post-eval participant communication.

\orgname{Artem Shelmanov}: Openreview website.

\orgname{Artem Vazhentsev}: Format checker.

\orgname{Claudio Savelli}: Hallushift++ baseline implementation. Data creation (Italian).

\orgname{Eduardo Calò}: Data creation (Italian).

\orgname{Emilio Raimond}: Data creation (French).

\orgname{Flavio Giobergia}: Inter-annotator agreement, data
quality checking.

\orgname{Hengyu Luo}: Data creation (Chinese).

\orgname{Jörg Tiedemann}: Organization support.

\orgname{Lorenzo Vaiani}: Code development.

\orgname{Stella Frank}: Data creation (English).

\orgname{Vincent Segonne}: Data creation (French).

\orgname{Timothee Mickus}: Data creation, submission interface, scoring program, in-loop-with participants; overall organization.
%\end{comment}

\begin{figure}[t!]
    \centering
    \begin{subfigure}{\linewidth}
        \centering
        \includegraphics[width=.9\linewidth]{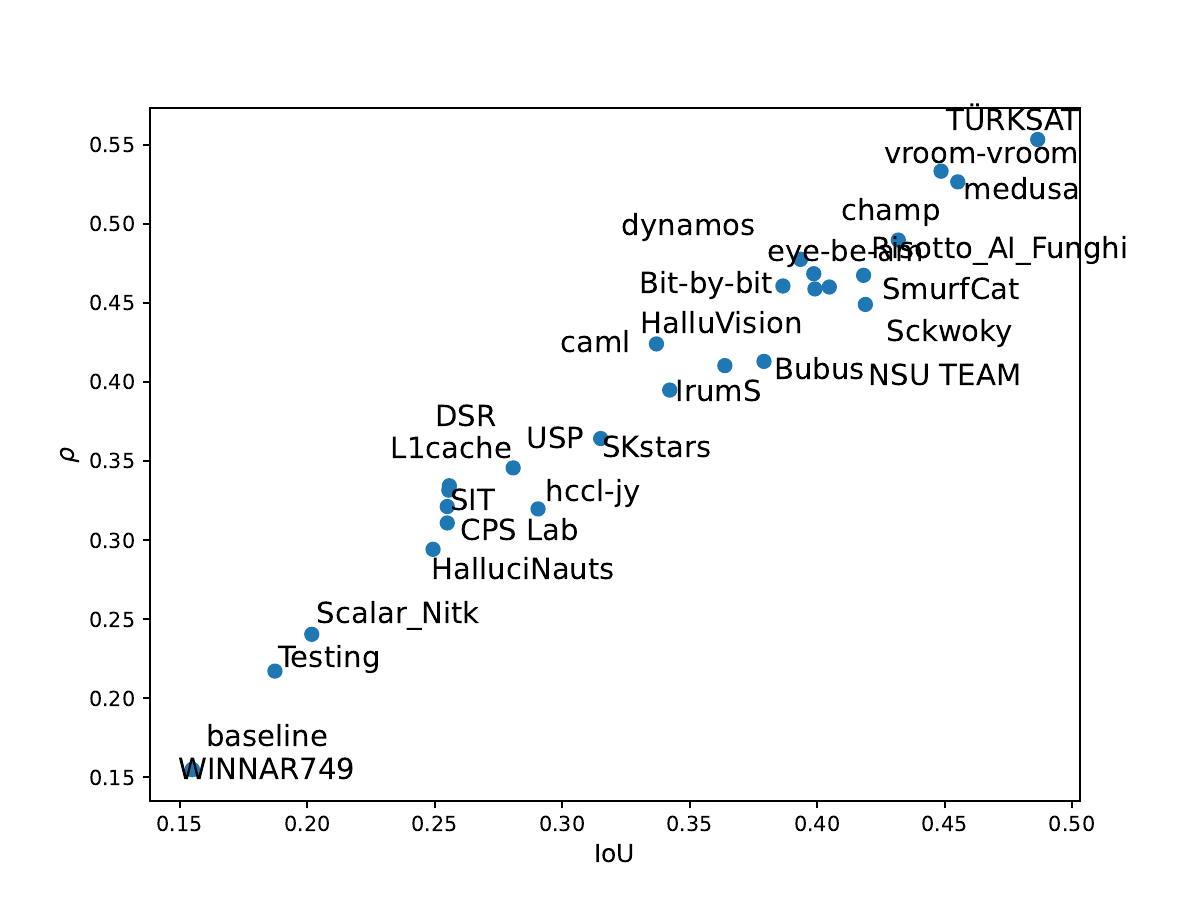}
        % \caption{EN}
    \end{subfigure}
    % \vspace{-5mm}
    \begin{subfigure}{\linewidth}
        \centering
        \includegraphics[width=.9\linewidth]{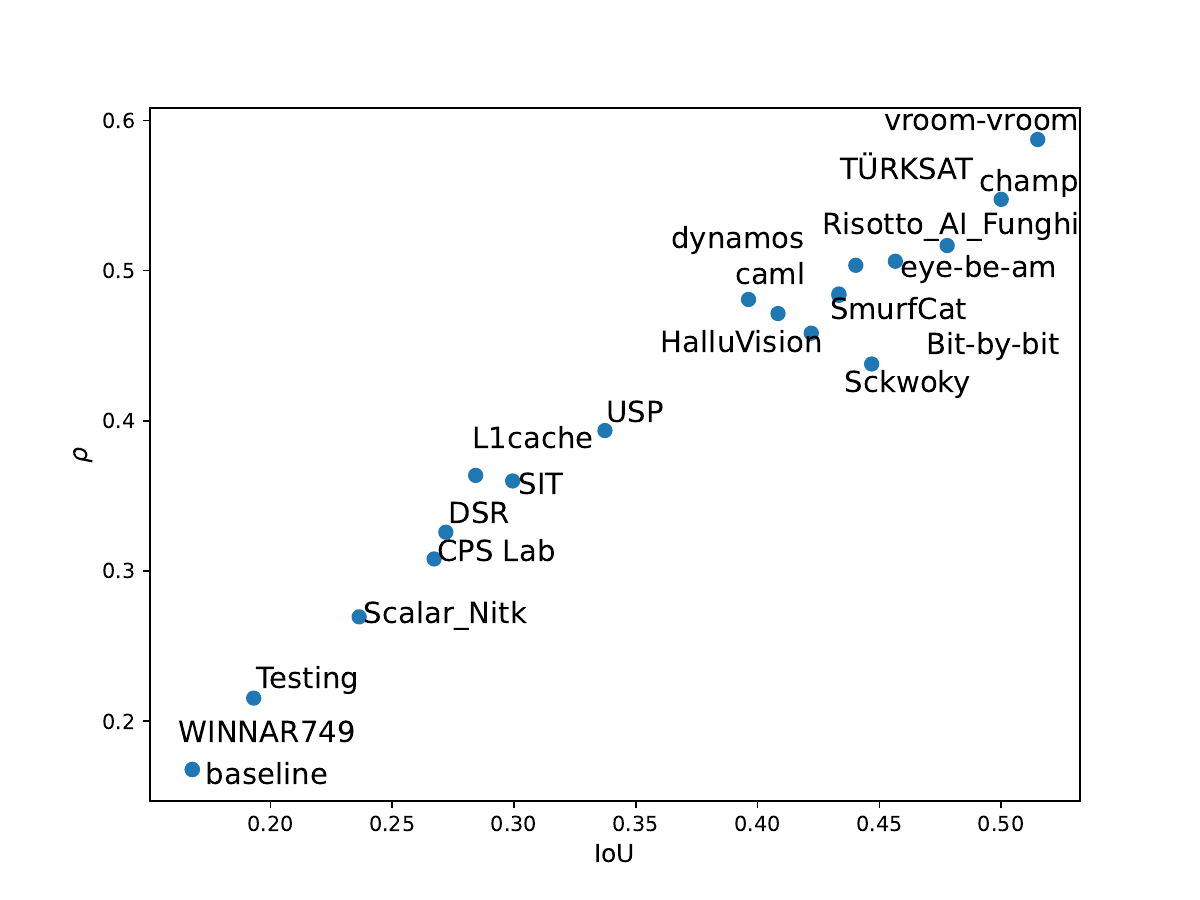}
        % \caption{FR}
    \end{subfigure}
    % \vspace{-2mm}
    \begin{subfigure}{\linewidth}
        \centering
        \includegraphics[width=.9\linewidth]{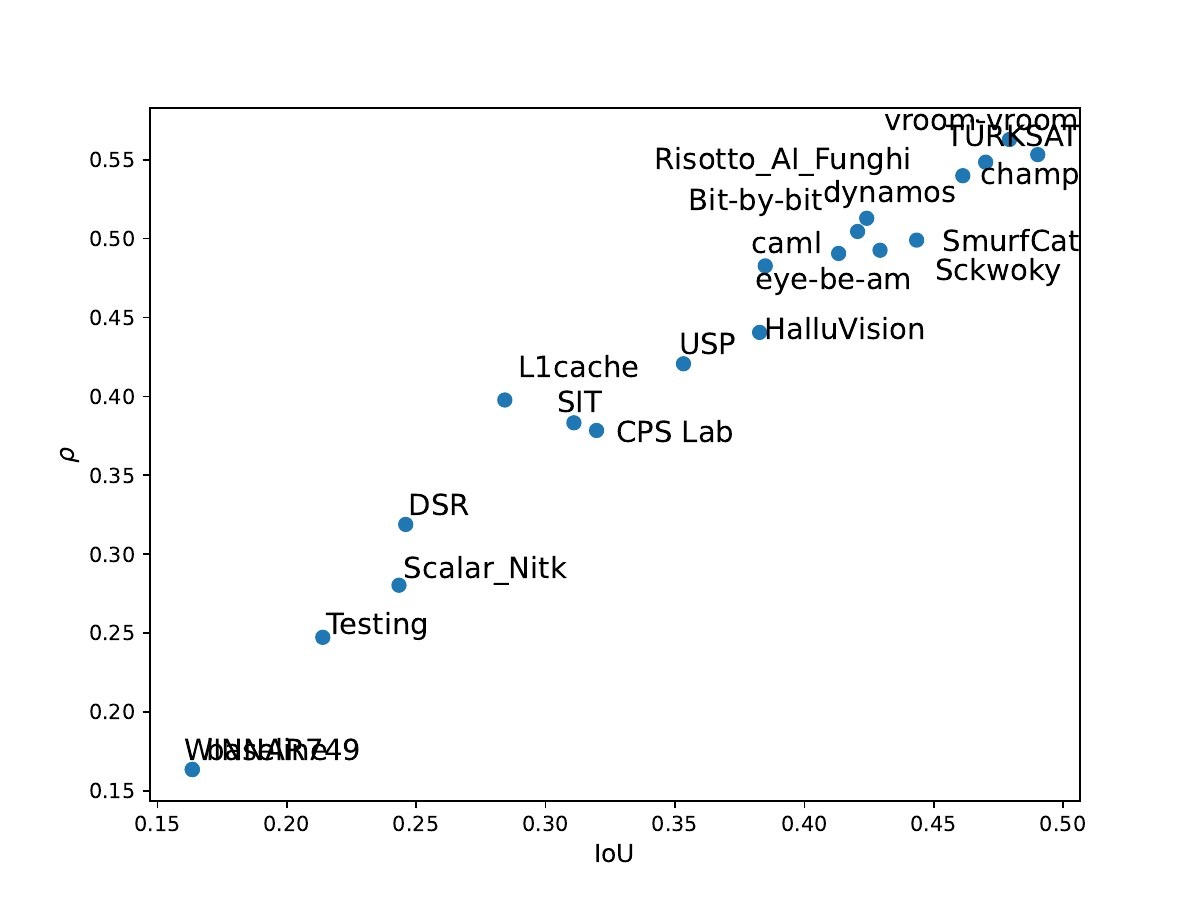}
        % \caption{IT}
    \end{subfigure}
    % \vspace{-2mm}
    \begin{subfigure}{\linewidth}
        \centering
        \includegraphics[width=.9\linewidth]{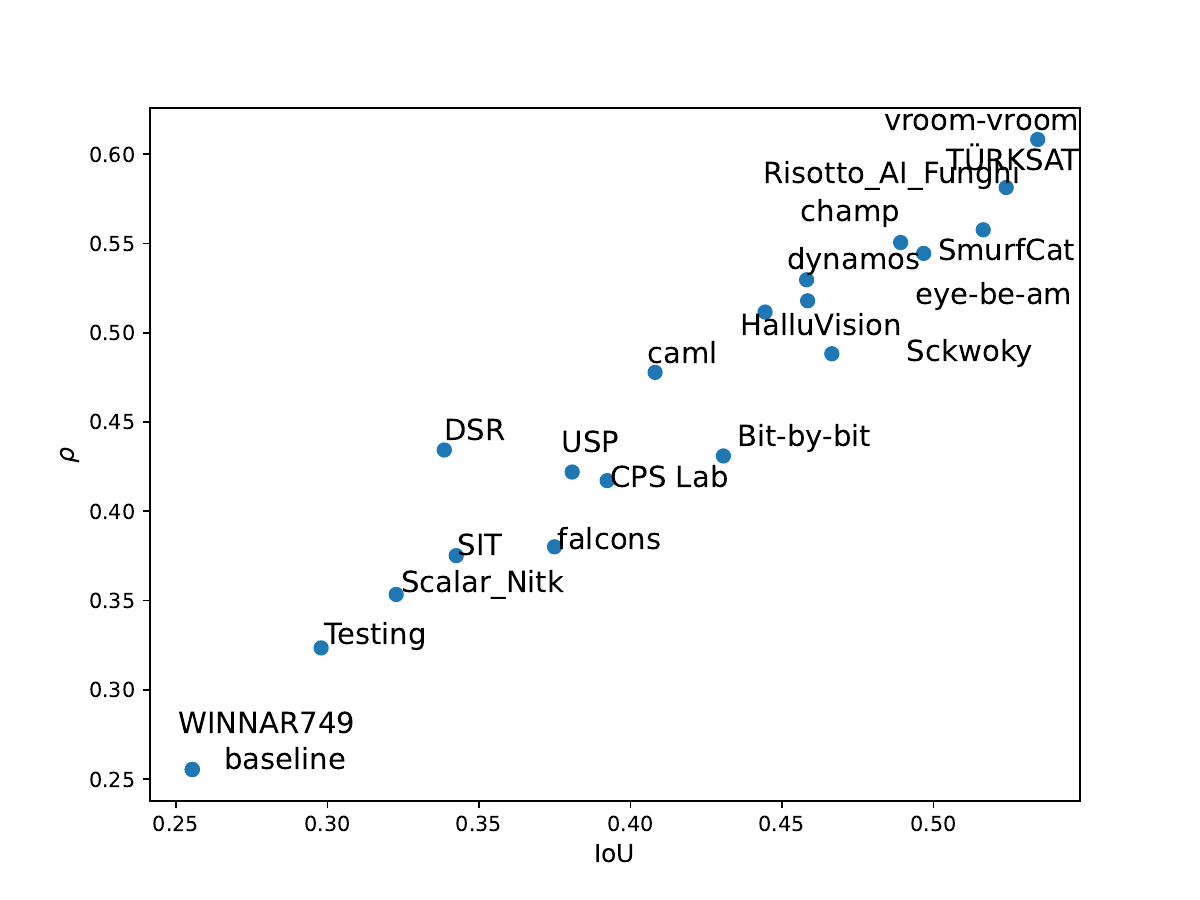}
        % \caption{ZH}
    \end{subfigure}
    % \vspace{-5mm}
    \caption{Overview of the performance by the best systems from each team in each language. From top to bottom: EN, FR, IT, ZH.}
    \label{fig:1x4}
\end{figure}

\section{Off with their heads! Test phase leaderboard}
\label{append:final-leaderboard}
In \Cref{tab:leaderboard_en}-\ref{tab:leaderboard_zh} we present the final test phase leaderboard. It is ranked based on the $\text{Corr}_\text{lbl}$ metric performance. We show the results obtained by each team's best-performing system after bootstrapping with 25,000 samples on each metric, accompanied by 95\% confidence intervals for each score, the mean rank and a rank interval (also at 95\%).  The baseline row is marked by light orange color. 

We also include the probability $P(>next)$ of any given submission outranking the submission one rank below, which we compute as part of the bootstrapping. The column quantifies instability in the ranking: values near 0.35--0.47 for adjacent top teams imply only a modest probability of consistently outperforming the next-ranked system. The bootstrap probabilities of pairwise inversion further indicate that the observed ordering of adjacent systems is often not robust. While some pairwise comparisons approach significance (e.g., \texttt{vroom-vroom} vs.\ \texttt{TÜRKSAT} in ZH at $P=0.35$), most fall well below conventional thresholds, suggesting that small mean rank differences should not be interpreted as definitive superiority. For the top systems in English, for example, the probability associated with the observed ordering of adjacent pairs ranges from 0.32 to 0.45, rather than approaching the high probabilities expected for a stable separation. Similar values are observed across the other languages. Thus, we would like to highlight that even is point estimates provide a useful summary of system performance, the leaderboard alone can give a misleading impression of fine-grained distinctions between systems.

\begin{table*}[t]
\centering
\small
\setlength{\tabcolsep}{3pt}
\begin{tabular}{r l r r r r r r r r r c}

\toprule
\multicolumn{2}{c}{} &
\multicolumn{3}{c}{Corr$_\text{lbl}$} &
\multicolumn{3}{c}{Corr} &
\multicolumn{3}{c}{IoU} &
\\
\cmidrule(lr){3-5} \cmidrule(lr){6-8} \cmidrule(lr){9-11}
\# & Team &
Score$_{\textcolor{darkgray}{[95\% CI]}}$ & MR & RI &
Score$_{\textcolor{darkgray}{[95\% CI]}}$ & MR & RI &
Score$_{\textcolor{darkgray}{[95\% CI]}}$ & MR & RI &
$P(>\!\text{next})$ \\
\midrule
1 & TÜRKSAT & 0.433$_{\textcolor{darkgray}{[.29,.57]}}$ & 2.9 & [1, 10] & 0.554$_{\textcolor{darkgray}{[.41,.69]}}$ & 2.8 & [1, 10] & 0.486$_{\textcolor{darkgray}{[.39,.58]}}$ & 2.0 & [1, 7] & 0.45 \\
2 & vroom-vroom & 0.425$_{\textcolor{darkgray}{[.29,.56]}}$ & 3.1 & [1, 9] & 0.533$_{\textcolor{darkgray}{[.39,.68]}}$ & 3.5 & [1, 10] & 0.448$_{\textcolor{darkgray}{[.35,.55]}}$ & 4.1 & [1, 11] & 0.45 \\
3 & medusa & 0.417$_{\textcolor{darkgray}{[.29,.55]}}$ & 3.7 & [1, 12] & 0.526$_{\textcolor{darkgray}{[.39,.66]}}$ & 4.1 & [1, 13] & 0.455$_{\textcolor{darkgray}{[.36,.55]}}$ & 3.7 & [1, 10] & 0.32 \\
4 & dynamos & 0.388$_{\textcolor{darkgray}{[.26,.52]}}$ & 5.2 & [1, 12] & 0.478$_{\textcolor{darkgray}{[.34,.61]}}$ & 7.3 & [2, 15] & 0.393$_{\textcolor{darkgray}{[.30,.48]}}$ & 9.4 & [3, 15] & 0.38 \\
5 & champ & 0.372$_{\textcolor{darkgray}{[.24,.51]}}$ & 6.5 & [1, 15] & 0.490$_{\textcolor{darkgray}{[.35,.63]}}$ & 6.4 & [1, 15] & 0.432$_{\textcolor{darkgray}{[.34,.53]}}$ & 5.5 & [1, 12] & 0.40 \\
6 & eye-be-am & 0.357$_{\textcolor{darkgray}{[.23,.49]}}$ & 7.8 & [2, 17] & 0.469$_{\textcolor{darkgray}{[.33,.60]}}$ & 8.2 & [2, 17] & 0.410$_{\textcolor{darkgray}{[.32,.50]}}$ & 7.8 & [2, 15] & 0.46 \\
7 & SmurfCat & 0.351$_{\textcolor{darkgray}{[.21,.50]}}$ & 8.4 & [2, 18] & 0.460$_{\textcolor{darkgray}{[.31,.61]}}$ & 9.0 & [2, 19] & 0.405$_{\textcolor{darkgray}{[.31,.51]}}$ & 8.3 & [2, 15] & 0.44 \\
8 & Bit-by-bit & 0.340$_{\textcolor{darkgray}{[.21,.48]}}$ & 9.2 & [2, 19] & 0.461$_{\textcolor{darkgray}{[.32,.60]}}$ & 8.8 & [2, 17] & 0.393$_{\textcolor{darkgray}{[.30,.49]}}$ & 9.5 & [3, 16] & 0.48 \\
9 & caml & 0.338$_{\textcolor{darkgray}{[.21,.47]}}$ & 9.4 & [3, 17] & 0.425$_{\textcolor{darkgray}{[.29,.56]}}$ & 12.1 & [4, 20] & 0.337$_{\textcolor{darkgray}{[.24,.43]}}$ & 14.9 & [8, 20] & 0.41 \\
10 & Risotto AI Funghi & 0.324$_{\textcolor{darkgray}{[.20,.46]}}$ & 10.8 & [3, 20] & 0.468$_{\textcolor{darkgray}{[.33,.60]}}$ & 8.2 & [2, 17] & 0.418$_{\textcolor{darkgray}{[.33,.51]}}$ & 6.8 & [2, 13] & 0.42 \\
11 & Sckwoky & 0.312$_{\textcolor{darkgray}{[.19,.45]}}$ & 11.9 & [4, 22] & 0.449$_{\textcolor{darkgray}{[.31,.59]}}$ & 10.0 & [2, 20] & 0.419$_{\textcolor{darkgray}{[.33,.52]}}$ & 6.7 & [1, 14] & 0.47 \\
12 & HalluVision & 0.307$_{\textcolor{darkgray}{[.19,.44]}}$ & 12.4 & [5, 22] & 0.411$_{\textcolor{darkgray}{[.28,.55]}}$ & 13.3 & [5, 21] & 0.367$_{\textcolor{darkgray}{[.28,.46]}}$ & 12.1 & [5, 17] & 0.47 \\
13 & IrumS & 0.301$_{\textcolor{darkgray}{[.17,.44]}}$ & 13.2 & [4, 25] & 0.395$_{\textcolor{darkgray}{[.25,.54]}}$ & 14.7 & [4, 24] & 0.342$_{\textcolor{darkgray}{[.25,.44]}}$ & 14.5 & [7, 21] & 0.44 \\
14 & SKstars & 0.290$_{\textcolor{darkgray}{[.17,.43]}}$ & 14.2 & [4, 25] & 0.364$_{\textcolor{darkgray}{[.22,.51]}}$ & 17.1 & [7, 25] & 0.315$_{\textcolor{darkgray}{[.22,.41]}}$ & 16.8 & [10, 23] & 0.45 \\
15 & Bubus & 0.282$_{\textcolor{darkgray}{[.17,.41]}}$ & 15.0 & [6, 24] & 0.414$_{\textcolor{darkgray}{[.29,.54]}}$ & 13.1 & [4, 22] & 0.379$_{\textcolor{darkgray}{[.29,.47]}}$ & 10.9 & [4, 17] & 0.42 \\
16 & USP & 0.267$_{\textcolor{darkgray}{[.14,.41]}}$ & 16.4 & [7, 26] & 0.346$_{\textcolor{darkgray}{[.21,.49]}}$ & 18.4 & [9, 25] & 0.280$_{\textcolor{darkgray}{[.19,.38]}}$ & 19.4 & [14, 24] & 0.50 \\
17 & L1cache & 0.267$_{\textcolor{darkgray}{[.18,.36]}}$ & 16.4 & [6, 26] & 0.374$_{\textcolor{darkgray}{[.28,.47]}}$ & 16.2 & [6, 24] & 0.262$_{\textcolor{darkgray}{[.18,.35]}}$ & 20.9 & [15, 26] & 0.45 \\
18 & smurfcat & 0.259$_{\textcolor{darkgray}{[.16,.37]}}$ & 17.4 & [7, 28] & 0.380$_{\textcolor{darkgray}{[.25,.51]}}$ & 16.0 & [7, 24] & 0.346$_{\textcolor{darkgray}{[.27,.43]}}$ & 14.2 & [7, 21] & 0.41 \\
19 & CPS Lab & 0.245$_{\textcolor{darkgray}{[.12,.38]}}$ & 18.7 & [9, 26] & 0.311$_{\textcolor{darkgray}{[.17,.45]}}$ & 21.0 & [13, 26] & 0.255$_{\textcolor{darkgray}{[.17,.35]}}$ & 21.5 & [16, 25] & 0.48 \\
20 & SIT & 0.241$_{\textcolor{darkgray}{[.11,.38]}}$ & 19.1 & [9, 27] & 0.322$_{\textcolor{darkgray}{[.18,.47]}}$ & 20.2 & [12, 25] & 0.255$_{\textcolor{darkgray}{[.16,.35]}}$ & 21.5 & [16, 26] & 0.37 \\
21 & NSU TEAM & 0.217$_{\textcolor{darkgray}{[.10,.35]}}$ & 21.1 & [10, 28] & 0.460$_{\textcolor{darkgray}{[.31,.61]}}$ & 9.1 & [2, 19] & 0.399$_{\textcolor{darkgray}{[.30,.50]}}$ & 9.0 & [2, 16] & 0.48 \\
22 & hccl-jy & 0.214$_{\textcolor{darkgray}{[.10,.34]}}$ & 21.4 & [12, 28] & 0.320$_{\textcolor{darkgray}{[.19,.46]}}$ & 20.4 & [12, 26] & 0.290$_{\textcolor{darkgray}{[.21,.38]}}$ & 18.7 & [13, 24] & 0.51 \\
23 & HalluciNauts & 0.214$_{\textcolor{darkgray}{[.10,.34]}}$ & 21.8 & [13, 28] & 0.294$_{\textcolor{darkgray}{[.17,.43]}}$ & 22.1 & [15, 27] & 0.249$_{\textcolor{darkgray}{[.17,.34]}}$ & 22.0 & [17, 26] & 0.45 \\
24 & Scalar Nitk & 0.208$_{\textcolor{darkgray}{[.08,.35]}}$ & 22.0 & [12, 28] & 0.240$_{\textcolor{darkgray}{[.11,.38]}}$ & 24.6 & [19, 28] & 0.208$_{\textcolor{darkgray}{[.12,.30]}}$ & 24.5 & [20, 28] & 0.45 \\
25 & DSR & 0.197$_{\textcolor{darkgray}{[.10,.31]}}$ & 23.2 & [16, 28] & 0.332$_{\textcolor{darkgray}{[.23,.45]}}$ & 19.8 & [12, 25] & 0.255$_{\textcolor{darkgray}{[.18,.34]}}$ & 21.5 & [16, 26] & 0.48 \\
26 & Testing & 0.196$_{\textcolor{darkgray}{[.07,.34]}}$ & 23.1 & [14, 28] & 0.218$_{\textcolor{darkgray}{[.09,.36]}}$ & 25.3 & [21, 28] & 0.187$_{\textcolor{darkgray}{[.10,.28]}}$ & 25.7 & [22, 28] & 0.20 \\
27 & baseline & 0.155$_{\textcolor{darkgray}{[.03,.30]}}$ & 25.4 & [18, 27] & 0.155$_{\textcolor{darkgray}{[.03,.30]}}$ & 26.6 & [24, 27] & 0.155$_{\textcolor{darkgray}{[.07,.25]}}$ & 26.6 & [24, 27] & \textit{0.00} \\
28 & WINNAR749 & 0.155$_{\textcolor{darkgray}{[.03,.30]}}$ & 25.4 & [18, 27] & 0.155$_{\textcolor{darkgray}{[.03,.30]}}$ & 26.6 & [24, 27] & 0.155$_{\textcolor{darkgray}{[.07,.25]}}$ & 26.6 & [24, 27] & --- \\
\bottomrule
\end{tabular}
\caption{Leaderboard for English sorted by $\text{Corr}_\text{lbl}$ score. MR = mean rank, RI = 95\% rank interval. $P(>\!\text{next})$: bootstrap probability that the next-lower-ranked team outperforms this team on $\text{Corr}_\text{lbl}$; \textbf{bold} $\geq 0.95$ = robust separation, \textit{italic} $\leq 0.20$ = possible inversion.}
\label{tab:leaderboard_en}
\end{table*}
\begin{table*}[t]
\centering
\small
\setlength{\tabcolsep}{3pt}
\begin{tabular}{r l r r r r r r r r r c}

\toprule
\multicolumn{2}{c}{} &
\multicolumn{3}{c}{Corr$_\text{lbl}$} &
\multicolumn{3}{c}{Corr} &
\multicolumn{3}{c}{IoU} &
\\
\cmidrule(lr){3-5} \cmidrule(lr){6-8} \cmidrule(lr){9-11}
\# & Team &
Score$_{\textcolor{darkgray}{[95\% CI]}}$ & MR & RI &
Score$_{\textcolor{darkgray}{[95\% CI]}}$ & MR & RI &
Score$_{\textcolor{darkgray}{[95\% CI]}}$ & MR & RI &
$P(>\!\text{next})$ \\
\midrule
1 & vroom-vroom & 0.504$_{\textcolor{darkgray}{[.40,.61]}}$ & 2.1 & [1, 6] & 0.608$_{\textcolor{darkgray}{[.51,.71]}}$ & 1.7 & [1, 5] & 0.534$_{\textcolor{darkgray}{[.43,.64]}}$ & 2.4 & [1, 7] & 0.35 \\
2 & TÜRKSAT & 0.485$_{\textcolor{darkgray}{[.39,.59]}}$ & 3.4 & [1, 9] & 0.581$_{\textcolor{darkgray}{[.48,.68]}}$ & 3.2 & [1, 9] & 0.524$_{\textcolor{darkgray}{[.42,.62]}}$ & 3.2 & [1, 10] & 0.38 \\
3 & dynamos & 0.469$_{\textcolor{darkgray}{[.36,.57]}}$ & 4.0 & [1, 8] & 0.530$_{\textcolor{darkgray}{[.42,.64]}}$ & 6.0 & [2, 10] & 0.458$_{\textcolor{darkgray}{[.35,.56]}}$ & 7.5 & [2, 12] & 0.48 \\
4 & champ & 0.467$_{\textcolor{darkgray}{[.36,.57]}}$ & 4.2 & [1, 8] & 0.550$_{\textcolor{darkgray}{[.45,.65]}}$ & 4.5 & [1, 9] & 0.489$_{\textcolor{darkgray}{[.39,.59]}}$ & 5.0 & [1, 9] & 0.41 \\
5 & SmurfCat & 0.457$_{\textcolor{darkgray}{[.35,.56]}}$ & 5.0 & [1, 10] & 0.544$_{\textcolor{darkgray}{[.44,.65]}}$ & 5.1 & [1, 10] & 0.501$_{\textcolor{darkgray}{[.40,.61]}}$ & 4.4 & [1, 10] & 0.48 \\
6 & Risotto AI Funghi & 0.454$_{\textcolor{darkgray}{[.35,.56]}}$ & 5.4 & [2, 11] & 0.557$_{\textcolor{darkgray}{[.46,.66]}}$ & 4.4 & [1, 10] & 0.516$_{\textcolor{darkgray}{[.41,.62]}}$ & 3.5 & [1, 9] & 0.27 \\
7 & eye-be-am & 0.426$_{\textcolor{darkgray}{[.32,.53]}}$ & 7.3 & [3, 12] & 0.518$_{\textcolor{darkgray}{[.41,.62]}}$ & 6.9 & [2, 12] & 0.460$_{\textcolor{darkgray}{[.36,.56]}}$ & 7.4 & [2, 13] & 0.46 \\
8 & caml & 0.422$_{\textcolor{darkgray}{[.32,.53]}}$ & 7.5 & [3, 12] & 0.478$_{\textcolor{darkgray}{[.37,.58]}}$ & 9.9 & [5, 15] & 0.408$_{\textcolor{darkgray}{[.30,.52]}}$ & 11.6 & [6, 16] & 0.44 \\
9 & HalluVision & 0.415$_{\textcolor{darkgray}{[.31,.52]}}$ & 8.1 & [3, 13] & 0.512$_{\textcolor{darkgray}{[.41,.61]}}$ & 7.3 & [2, 12] & 0.444$_{\textcolor{darkgray}{[.34,.55]}}$ & 8.6 & [3, 14] & 0.22 \\
10 & Sckwoky & 0.376$_{\textcolor{darkgray}{[.27,.48]}}$ & 10.8 & [6, 17] & 0.488$_{\textcolor{darkgray}{[.38,.59]}}$ & 9.1 & [3, 15] & 0.467$_{\textcolor{darkgray}{[.36,.57]}}$ & 6.8 & [1, 13] & 0.37 \\
11 & USP & 0.360$_{\textcolor{darkgray}{[.25,.47]}}$ & 12.0 & [7, 18] & 0.422$_{\textcolor{darkgray}{[.32,.53]}}$ & 13.9 & [9, 19] & 0.381$_{\textcolor{darkgray}{[.28,.49]}}$ & 13.6 & [8, 18] & 0.40 \\
12 & Bit-by-bit & 0.348$_{\textcolor{darkgray}{[.24,.45]}}$ & 13.3 & [8, 20] & 0.431$_{\textcolor{darkgray}{[.32,.54]}}$ & 13.3 & [8, 19] & 0.431$_{\textcolor{darkgray}{[.33,.54]}}$ & 9.8 & [4, 15] & 0.39 \\
13 & SIT & 0.334$_{\textcolor{darkgray}{[.22,.45]}}$ & 14.3 & [9, 20] & 0.375$_{\textcolor{darkgray}{[.26,.49]}}$ & 17.0 & [12, 20] & 0.343$_{\textcolor{darkgray}{[.23,.46]}}$ & 16.1 & [11, 19] & 0.50 \\
14 & smurfcat & 0.334$_{\textcolor{darkgray}{[.24,.43]}}$ & 14.7 & [8, 22] & 0.455$_{\textcolor{darkgray}{[.35,.56]}}$ & 11.6 & [6, 18] & 0.409$_{\textcolor{darkgray}{[.31,.51]}}$ & 11.6 & [6, 17] & 0.47 \\
15 & CPS Lab & 0.330$_{\textcolor{darkgray}{[.23,.44]}}$ & 14.9 & [10, 22] & 0.417$_{\textcolor{darkgray}{[.31,.53]}}$ & 14.3 & [9, 19] & 0.392$_{\textcolor{darkgray}{[.29,.50]}}$ & 12.9 & [7, 18] & 0.32 \\
16 & DSR & 0.309$_{\textcolor{darkgray}{[.21,.41]}}$ & 16.7 & [11, 22] & 0.434$_{\textcolor{darkgray}{[.33,.54]}}$ & 13.0 & [8, 18] & 0.339$_{\textcolor{darkgray}{[.24,.44]}}$ & 16.4 & [12, 21] & 0.49 \\
17 & Scalar Nitk & 0.308$_{\textcolor{darkgray}{[.20,.42]}}$ & 16.7 & [10, 22] & 0.353$_{\textcolor{darkgray}{[.24,.47]}}$ & 18.0 & [13, 21] & 0.323$_{\textcolor{darkgray}{[.22,.44]}}$ & 17.2 & [13, 21] & 0.49 \\
18 & L1cache & 0.307$_{\textcolor{darkgray}{[.21,.41]}}$ & 16.9 & [11, 22] & 0.411$_{\textcolor{darkgray}{[.30,.52]}}$ & 14.7 & [9, 19] & --- & --- & -- & 0.49 \\
19 & falcons & 0.306$_{\textcolor{darkgray}{[.20,.42]}}$ & 16.9 & [11, 21] & 0.380$_{\textcolor{darkgray}{[.27,.49]}}$ & 16.7 & [12, 20] & 0.375$_{\textcolor{darkgray}{[.27,.48]}}$ & 13.9 & [8, 18] & 0.44 \\
20 & Testing & 0.300$_{\textcolor{darkgray}{[.19,.41]}}$ & 17.4 & [12, 22] & 0.323$_{\textcolor{darkgray}{[.21,.44]}}$ & 19.4 & [16, 21] & 0.298$_{\textcolor{darkgray}{[.19,.41]}}$ & 18.5 & [15, 21] & \textit{0.09} \\
21 & WINNAR749 & 0.255$_{\textcolor{darkgray}{[.15,.37]}}$ & 20.3 & [16, 21] & 0.255$_{\textcolor{darkgray}{[.15,.37]}}$ & 20.9 & [20, 21] & 0.256$_{\textcolor{darkgray}{[.15,.37]}}$ & 19.7 & [18, 20] & \textit{0.00} \\
22 & baseline & 0.255$_{\textcolor{darkgray}{[.15,.37]}}$ & 20.3 & [16, 21] & 0.255$_{\textcolor{darkgray}{[.15,.37]}}$ & 20.9 & [20, 21] & 0.256$_{\textcolor{darkgray}{[.15,.37]}}$ & 19.7 & [18, 20] & --- \\
\bottomrule
\end{tabular}
\caption{Leaderboard for Chinese sorted by $\text{Corr}_\text{lbl}$ score. MR = mean rank, RI = 95\% rank interval. $P(>\!\text{next})$: bootstrap probability that the next-lower-ranked team outperforms this team on $\text{Corr}_\text{lbl}$; \textbf{bold} $\geq 0.95$ = robust separation, \textit{italic} $\leq 0.20$ = possible inversion.}
\label{tab:leaderboard_zh}
\end{table*}
\begin{table*}[t]
\centering
\small
\setlength{\tabcolsep}{3pt}
\begin{tabular}{r l r r r r r r r r r c}

\toprule
\multicolumn{2}{c}{} &
\multicolumn{3}{c}{Corr$_\text{lbl}$} &
\multicolumn{3}{c}{Corr} &
\multicolumn{3}{c}{IoU} &
\\
\cmidrule(lr){3-5} \cmidrule(lr){6-8} \cmidrule(lr){9-11}
\# & Team &
Score$_{\textcolor{darkgray}{[95\% CI]}}$ & MR & RI &
Score$_{\textcolor{darkgray}{[95\% CI]}}$ & MR & RI &
Score$_{\textcolor{darkgray}{[95\% CI]}}$ & MR & RI &
$P(>\!\text{next})$ \\
\midrule
1 & vroom-vroom & 0.477$_{\textcolor{darkgray}{[.34,.61]}}$ & 2.0 & [1, 7] & 0.586$_{\textcolor{darkgray}{[.45,.72]}}$ & 1.9 & [1, 7] & 0.514$_{\textcolor{darkgray}{[.42,.62]}}$ & 2.0 & [1, 6] & 0.35 \\
2 & TÜRKSAT & 0.452$_{\textcolor{darkgray}{[.31,.60]}}$ & 3.3 & [1, 11] & 0.545$_{\textcolor{darkgray}{[.40,.68]}}$ & 3.7 & [1, 11] & 0.499$_{\textcolor{darkgray}{[.40,.60]}}$ & 2.8 & [1, 9] & 0.27 \\
3 & champ & 0.408$_{\textcolor{darkgray}{[.28,.54]}}$ & 5.3 & [1, 12] & 0.516$_{\textcolor{darkgray}{[.39,.65]}}$ & 4.9 & [1, 11] & 0.477$_{\textcolor{darkgray}{[.39,.57]}}$ & 3.7 & [1, 8] & 0.47 \\
4 & dynamos & 0.404$_{\textcolor{darkgray}{[.28,.53]}}$ & 5.5 & [1, 13] & 0.502$_{\textcolor{darkgray}{[.37,.63]}}$ & 5.8 & [1, 12] & 0.440$_{\textcolor{darkgray}{[.35,.53]}}$ & 6.6 & [2, 11] & 0.45 \\
5 & SmurfCat & 0.397$_{\textcolor{darkgray}{[.25,.55]}}$ & 6.2 & [1, 14] & 0.483$_{\textcolor{darkgray}{[.34,.63]}}$ & 7.2 & [1, 14] & 0.436$_{\textcolor{darkgray}{[.34,.54]}}$ & 7.1 & [2, 12] & 0.44 \\
6 & eye-be-am & 0.388$_{\textcolor{darkgray}{[.26,.53]}}$ & 6.6 & [2, 13] & 0.482$_{\textcolor{darkgray}{[.34,.62]}}$ & 7.2 & [2, 14] & 0.434$_{\textcolor{darkgray}{[.34,.53]}}$ & 7.1 & [2, 12] & 0.40 \\
7 & Risotto AI Funghi & 0.377$_{\textcolor{darkgray}{[.25,.51]}}$ & 7.4 & [2, 14] & 0.505$_{\textcolor{darkgray}{[.37,.63]}}$ & 5.7 & [2, 12] & 0.457$_{\textcolor{darkgray}{[.36,.55]}}$ & 5.3 & [2, 11] & 0.50 \\
8 & caml & 0.375$_{\textcolor{darkgray}{[.26,.51]}}$ & 7.5 & [2, 14] & 0.482$_{\textcolor{darkgray}{[.35,.61]}}$ & 7.1 & [2, 13] & 0.399$_{\textcolor{darkgray}{[.32,.48]}}$ & 9.9 & [5, 13] & 0.46 \\
9 & Bit-by-bit & 0.371$_{\textcolor{darkgray}{[.24,.51]}}$ & 7.9 & [2, 15] & 0.471$_{\textcolor{darkgray}{[.33,.62]}}$ & 8.0 & [2, 14] & 0.414$_{\textcolor{darkgray}{[.33,.51]}}$ & 8.7 & [3, 12] & 0.43 \\
10 & HalluVision & 0.358$_{\textcolor{darkgray}{[.23,.49]}}$ & 8.8 & [2, 16] & 0.456$_{\textcolor{darkgray}{[.32,.58]}}$ & 8.9 & [3, 15] & 0.422$_{\textcolor{darkgray}{[.33,.51]}}$ & 8.1 & [3, 12] & 0.24 \\
11 & USP & 0.319$_{\textcolor{darkgray}{[.19,.46]}}$ & 11.5 & [4, 18] & 0.391$_{\textcolor{darkgray}{[.25,.53]}}$ & 12.6 & [6, 17] & 0.336$_{\textcolor{darkgray}{[.24,.43]}}$ & 13.1 & [9, 16] & 0.49 \\
12 & Sckwoky & 0.316$_{\textcolor{darkgray}{[.18,.45]}}$ & 11.7 & [5, 18] & 0.436$_{\textcolor{darkgray}{[.30,.57]}}$ & 10.0 & [2, 16] & 0.446$_{\textcolor{darkgray}{[.35,.54]}}$ & 6.3 & [1, 12] & 0.41 \\
13 & smurfcat & 0.299$_{\textcolor{darkgray}{[.19,.42]}}$ & 13.2 & [6, 21] & 0.397$_{\textcolor{darkgray}{[.28,.52]}}$ & 12.6 & [6, 18] & 0.379$_{\textcolor{darkgray}{[.30,.46]}}$ & 11.2 & [6, 15] & 0.43 \\
14 & L1cache & 0.290$_{\textcolor{darkgray}{[.20,.39]}}$ & 13.4 & [5, 19] & 0.388$_{\textcolor{darkgray}{[.29,.49]}}$ & 12.7 & [5, 17] & 0.291$_{\textcolor{darkgray}{[.20,.39]}}$ & 15.0 & [12, 18] & 0.46 \\
15 & SIT & 0.285$_{\textcolor{darkgray}{[.15,.43]}}$ & 13.7 & [6, 19] & 0.358$_{\textcolor{darkgray}{[.22,.52]}}$ & 14.2 & [8, 18] & 0.298$_{\textcolor{darkgray}{[.20,.41]}}$ & 14.7 & [11, 18] & 0.22 \\
16 & Scalar Nitk & 0.241$_{\textcolor{darkgray}{[.11,.38]}}$ & 16.2 & [10, 21] & 0.269$_{\textcolor{darkgray}{[.13,.43]}}$ & 17.3 & [13, 21] & 0.236$_{\textcolor{darkgray}{[.14,.35]}}$ & 17.5 & [15, 20] & 0.51 \\
17 & CPS Lab & 0.240$_{\textcolor{darkgray}{[.12,.38]}}$ & 16.3 & [11, 21] & 0.310$_{\textcolor{darkgray}{[.18,.46]}}$ & 16.1 & [11, 19] & 0.268$_{\textcolor{darkgray}{[.18,.37]}}$ & 16.1 & [13, 19] & 0.49 \\
18 & DSR & 0.240$_{\textcolor{darkgray}{[.12,.37]}}$ & 16.4 & [10, 21] & 0.324$_{\textcolor{darkgray}{[.20,.47]}}$ & 15.7 & [10, 19] & 0.271$_{\textcolor{darkgray}{[.19,.36]}}$ & 16.1 & [13, 19] & 0.28 \\
19 & Testing & 0.204$_{\textcolor{darkgray}{[.07,.35]}}$ & 18.2 & [13, 21] & 0.216$_{\textcolor{darkgray}{[.08,.38]}}$ & 18.9 & [16, 21] & 0.193$_{\textcolor{darkgray}{[.10,.29]}}$ & 19.2 & [17, 21] & \textit{0.20} \\
20 & WINNAR749 & 0.167$_{\textcolor{darkgray}{[.04,.31]}}$ & 19.3 & [15, 20] & 0.168$_{\textcolor{darkgray}{[.04,.32]}}$ & 19.7 & [18, 20] & 0.168$_{\textcolor{darkgray}{[.09,.27]}}$ & 19.7 & [18, 20] & \textit{0.00} \\
21 & baseline & 0.167$_{\textcolor{darkgray}{[.04,.31]}}$ & 19.3 & [15, 20] & 0.168$_{\textcolor{darkgray}{[.04,.32]}}$ & 19.7 & [18, 20] & 0.168$_{\textcolor{darkgray}{[.09,.27]}}$ & 19.7 & [18, 20] & --- \\
\bottomrule
\end{tabular}
\caption{Leaderboard for French sorted by $\text{Corr}_\text{lbl}$ score. MR = mean rank, RI = 95\% rank interval. $P(>\!\text{next})$: bootstrap probability that the next-lower-ranked team outperforms this team on $\text{Corr}_\text{lbl}$; \textbf{bold} $\geq 0.95$ = robust separation, \textit{italic} $\leq 0.20$ = possible inversion.}
\label{tab:leaderboard_fr}
\end{table*}
\begin{table*}[t]
\centering
\small
\setlength{\tabcolsep}{3pt}
\begin{tabular}{r l r r r r r r r r r c}

\toprule
\multicolumn{2}{c}{} &
\multicolumn{3}{c}{Corr$_\text{lbl}$} &
\multicolumn{3}{c}{Corr} &
\multicolumn{3}{c}{IoU} &
\\
\cmidrule(lr){3-5} \cmidrule(lr){6-8} \cmidrule(lr){9-11}
\# & Team &
Score$_{\textcolor{darkgray}{[95\% CI]}}$ & MR & RI &
Score$_{\textcolor{darkgray}{[95\% CI]}}$ & MR & RI &
Score$_{\textcolor{darkgray}{[95\% CI]}}$ & MR & RI &
$P(>\!\text{next})$ \\
\midrule
1 & vroom-vroom & 0.450$_{\textcolor{darkgray}{[.32,.58]}}$ & 2.9 & [1, 9] & 0.563$_{\textcolor{darkgray}{[.42,.70]}}$ & 3.2 & [1, 9] & 0.480$_{\textcolor{darkgray}{[.38,.57]}}$ & 3.0 & [1, 8] & 0.47 \\
2 & TÜRKSAT & 0.445$_{\textcolor{darkgray}{[.31,.58]}}$ & 3.5 & [1, 11] & 0.552$_{\textcolor{darkgray}{[.41,.69]}}$ & 4.1 & [1, 12] & 0.490$_{\textcolor{darkgray}{[.39,.59]}}$ & 2.7 & [1, 8] & 0.41 \\
3 & Risotto AI Funghi & 0.429$_{\textcolor{darkgray}{[.30,.56]}}$ & 4.3 & [1, 11] & 0.548$_{\textcolor{darkgray}{[.41,.69]}}$ & 4.2 & [1, 11] & 0.471$_{\textcolor{darkgray}{[.37,.56]}}$ & 3.7 & [1, 9] & 0.43 \\
4 & champ & 0.421$_{\textcolor{darkgray}{[.29,.55]}}$ & 4.6 & [1, 11] & 0.540$_{\textcolor{darkgray}{[.41,.67]}}$ & 4.4 & [1, 10] & 0.462$_{\textcolor{darkgray}{[.37,.55]}}$ & 4.0 & [1, 9] & 0.34 \\
5 & dynamos & 0.400$_{\textcolor{darkgray}{[.28,.53]}}$ & 6.0 & [1, 14] & 0.513$_{\textcolor{darkgray}{[.38,.64]}}$ & 6.2 & [1, 13] & 0.425$_{\textcolor{darkgray}{[.34,.51]}}$ & 7.0 & [2, 12] & 0.45 \\
6 & SmurfCat & 0.392$_{\textcolor{darkgray}{[.26,.54]}}$ & 6.7 & [1, 14] & 0.499$_{\textcolor{darkgray}{[.35,.65]}}$ & 7.3 & [1, 15] & 0.444$_{\textcolor{darkgray}{[.35,.54]}}$ & 5.7 & [1, 11] & 0.42 \\
7 & caml & 0.379$_{\textcolor{darkgray}{[.25,.51]}}$ & 7.6 & [2, 14] & 0.483$_{\textcolor{darkgray}{[.35,.62]}}$ & 8.4 & [3, 14] & 0.386$_{\textcolor{darkgray}{[.29,.48]}}$ & 10.4 & [6, 14] & 0.47 \\
8 & eye-be-am & 0.374$_{\textcolor{darkgray}{[.25,.51]}}$ & 7.9 & [2, 14] & 0.490$_{\textcolor{darkgray}{[.36,.62]}}$ & 7.9 & [2, 14] & 0.412$_{\textcolor{darkgray}{[.33,.50]}}$ & 8.2 & [3, 12] & 0.43 \\
9 & Bit-by-bit & 0.366$_{\textcolor{darkgray}{[.24,.50]}}$ & 8.6 & [2, 15] & 0.504$_{\textcolor{darkgray}{[.36,.64]}}$ & 6.8 & [1, 13] & 0.421$_{\textcolor{darkgray}{[.32,.52]}}$ & 7.4 & [2, 12] & 0.42 \\
10 & HalluVision & 0.352$_{\textcolor{darkgray}{[.22,.49]}}$ & 9.6 & [3, 16] & 0.441$_{\textcolor{darkgray}{[.30,.58]}}$ & 11.1 & [4, 16] & 0.389$_{\textcolor{darkgray}{[.29,.49]}}$ & 9.9 & [4, 14] & 0.41 \\
11 & Sckwoky & 0.338$_{\textcolor{darkgray}{[.20,.48]}}$ & 10.6 & [3, 16] & 0.493$_{\textcolor{darkgray}{[.35,.63]}}$ & 7.6 & [1, 14] & 0.430$_{\textcolor{darkgray}{[.34,.53]}}$ & 6.6 & [1, 12] & 0.47 \\
12 & USP & 0.332$_{\textcolor{darkgray}{[.20,.48]}}$ & 10.9 & [3, 17] & 0.421$_{\textcolor{darkgray}{[.28,.57]}}$ & 12.1 & [5, 17] & 0.355$_{\textcolor{darkgray}{[.26,.46]}}$ & 12.3 & [7, 16] & 0.30 \\
13 & L1cache & 0.300$_{\textcolor{darkgray}{[.21,.40]}}$ & 13.2 & [6, 19] & 0.404$_{\textcolor{darkgray}{[.31,.50]}}$ & 13.1 & [6, 18] & 0.282$_{\textcolor{darkgray}{[.20,.37]}}$ & 15.7 & [12, 19] & 0.52 \\
14 & smurfcat & 0.298$_{\textcolor{darkgray}{[.19,.41]}}$ & 13.5 & [6, 21] & 0.419$_{\textcolor{darkgray}{[.28,.56]}}$ & 12.2 & [5, 17] & 0.365$_{\textcolor{darkgray}{[.29,.44]}}$ & 11.6 & [6, 16] & 0.45 \\
15 & SIT & 0.290$_{\textcolor{darkgray}{[.16,.43]}}$ & 13.9 & [7, 19] & 0.383$_{\textcolor{darkgray}{[.24,.53]}}$ & 14.1 & [8, 17] & 0.312$_{\textcolor{darkgray}{[.22,.42]}}$ & 14.5 & [11, 17] & 0.40 \\
16 & CPS Lab & 0.271$_{\textcolor{darkgray}{[.14,.40]}}$ & 14.9 & [8, 19] & 0.379$_{\textcolor{darkgray}{[.24,.52]}}$ & 14.4 & [8, 18] & 0.320$_{\textcolor{darkgray}{[.23,.42]}}$ & 14.2 & [10, 17] & 0.26 \\
17 & Scalar Nitk & 0.229$_{\textcolor{darkgray}{[.10,.37]}}$ & 17.1 & [12, 21] & 0.281$_{\textcolor{darkgray}{[.14,.43]}}$ & 17.7 & [14, 21] & 0.242$_{\textcolor{darkgray}{[.15,.34]}}$ & 17.4 & [15, 20] & 0.47 \\
18 & DSR & 0.225$_{\textcolor{darkgray}{[.12,.35]}}$ & 17.5 & [13, 21] & 0.320$_{\textcolor{darkgray}{[.20,.45]}}$ & 16.8 & [13, 19] & 0.245$_{\textcolor{darkgray}{[.16,.34]}}$ & 17.5 & [14, 19] & 0.44 \\
19 & Testing & 0.216$_{\textcolor{darkgray}{[.09,.35]}}$ & 17.7 & [13, 21] & 0.248$_{\textcolor{darkgray}{[.12,.39]}}$ & 18.6 & [16, 21] & 0.214$_{\textcolor{darkgray}{[.13,.32]}}$ & 18.6 & [16, 21] & \textit{0.13} \\
20 & WINNAR749 & 0.163$_{\textcolor{darkgray}{[.04,.30]}}$ & 19.4 & [16, 20] & 0.165$_{\textcolor{darkgray}{[.04,.31]}}$ & 19.9 & [18, 20] & 0.163$_{\textcolor{darkgray}{[.09,.25]}}$ & 19.8 & [18, 20] & \textit{0.00} \\
21 & baseline & 0.163$_{\textcolor{darkgray}{[.04,.30]}}$ & 19.4 & [16, 20] & 0.165$_{\textcolor{darkgray}{[.04,.31]}}$ & 19.9 & [18, 20] & 0.163$_{\textcolor{darkgray}{[.09,.25]}}$ & 19.8 & [18, 20] & --- \\
\bottomrule
\end{tabular}
\caption{Leaderboard for Italian sorted by $\text{Corr}_\text{lbl}$ score. MR = mean rank, RI = 95\% rank interval. $P(>\!\text{next})$: bootstrap probability that the next-lower-ranked team outperforms this team on $\text{Corr}_\text{lbl}$; \textbf{bold} $\geq 0.95$ = robust separation, \textit{italic} $\leq 0.20$ = possible inversion.}
\label{tab:leaderboard_it}
\end{table*}

\section{Painting the white roses red: Strategy-specific performance scores}
\label{append:strategy-scores}

To complement the ranking-based analysis in \Cref{sec:strategy-analysis}, we report the underlying system scores for each annotation strategy in Tables \ref{tab:strategy_scores_en_cor}--\ref{tab:strategy_scores_fr_iou}. These tables provide the complete results for all teams across the global ranking and for each of the strategy-level partitions of the SHEEP dataset (human-written, randomly sampled, and silver-label or model assisted selection), for each language and evaluation metric. Teams are ordered by their score on the global ranking dataset to facilitate comparison with the main leaderboard. Unlike the official leaderboard, where the score is computed as the mean of a bootstrap strategy, the global score shown here is computed with the best submission of each team without resampling.  

These results make explicit how the choice of annotation strategy affects absolute performance estimates. In particular, they allow the reader to distinguish changes in overall score from changes in relative system ordering: a strategy may shift scores for most systems without substantially altering the leaderboard, while selective changes in individual systems can produce the rank inversions discussed in the main text. Together with the rank correlations and rank-flow plots in \Cref{sec:strategy-analysis}, these tables provide the underlying evidence for our analysis of strategy-dependent evaluation effects.

\begin{table}[t]
\centering
\small
\setlength{\tabcolsep}{5pt}
\resizebox{\columnwidth}{!}{
\begin{tabular}{r l r r r r}
\toprule
\# & Team &
\multicolumn{1}{c}{Global} &
\multicolumn{1}{c}{Human} &
\multicolumn{1}{c}{Rndm} &
\multicolumn{1}{c}{Silver} \\
\midrule
1 & TÜRKSAT & \textbf{0.553} & 0.646 & \textbf{0.492} & \textbf{0.522} \\
2 & vroom-vroom & 0.533 & \textbf{0.684} & 0.440 & 0.474 \\
3 & medusa & 0.527 & 0.626 & 0.462 & 0.491 \\
4 & champ & 0.490 & 0.562 & 0.428 & 0.478 \\
5 & dynamos & 0.478 & 0.583 & 0.420 & 0.429 \\
6 & eye-be-am & 0.469 & 0.562 & 0.419 & 0.424 \\
7 & Risotto\_AI\_F. & 0.468 & 0.554 & 0.414 & 0.434 \\
8 & Bit-by-bit & 0.461 & 0.553 & 0.406 & 0.422 \\
9 & SmurfCat & 0.460 & 0.571 & 0.396 & 0.413 \\
10 & NSU TEAM & 0.459 & 0.528 & 0.413 & 0.435 \\
11 & Sckwoky & 0.449 & 0.484 & 0.440 & 0.423 \\
12 & caml & 0.424 & 0.512 & 0.393 & 0.369 \\
13 & Bubus & 0.413 & 0.458 & 0.408 & 0.375 \\
14 & HalluVision & 0.410 & 0.459 & 0.389 & 0.384 \\
15 & IrumS & 0.395 & 0.506 & 0.340 & 0.339 \\
16 & L1cache & 0.373 & 0.297 & 0.436 & 0.389 \\
17 & SKstars & 0.364 & 0.472 & 0.323 & 0.299 \\
18 & USP & 0.346 & 0.373 & 0.338 & 0.327 \\
19 & DSR & 0.332 & 0.372 & 0.302 & 0.320 \\
20 & SIT & 0.321 & 0.325 & 0.344 & 0.297 \\
21 & hccl-jy & 0.320 & 0.352 & 0.336 & 0.274 \\
22 & CPS Lab & 0.311 & 0.271 & 0.345 & 0.317 \\
23 & HalluciNauts & 0.294 & 0.285 & 0.318 & 0.281 \\
24 & Scalar\_Nitk & 0.240 & 0.252 & 0.227 & 0.242 \\
25 & Testing & 0.217 & 0.140 & 0.299 & 0.216 \\
26 & WINNAR749 & 0.155 & 0.013 & 0.296 & 0.162 \\
27 & baseline & 0.155 & 0.013 & 0.296 & 0.162 \\
\bottomrule
\end{tabular}}
\caption{Strategy-specific Corr scores for the English leaderboard. 
Teams are ordered by their score on the full dataset (Global). 
Human-written, Random, and Silver-label report scores computed on the corresponding 
data partitions. Boldface indicates the highest score within each column.}
\label{tab:strategy_scores_en_cor}
\end{table}

\begin{figure*}[t]
    \centering

    % ================= ENGLISH =================
    \begin{minipage}[t]{0.48\textwidth}
        \centering
        \includegraphics[
            width=\linewidth,
            trim={3cm 0 1cm 0.6cm},
            clip
        ]{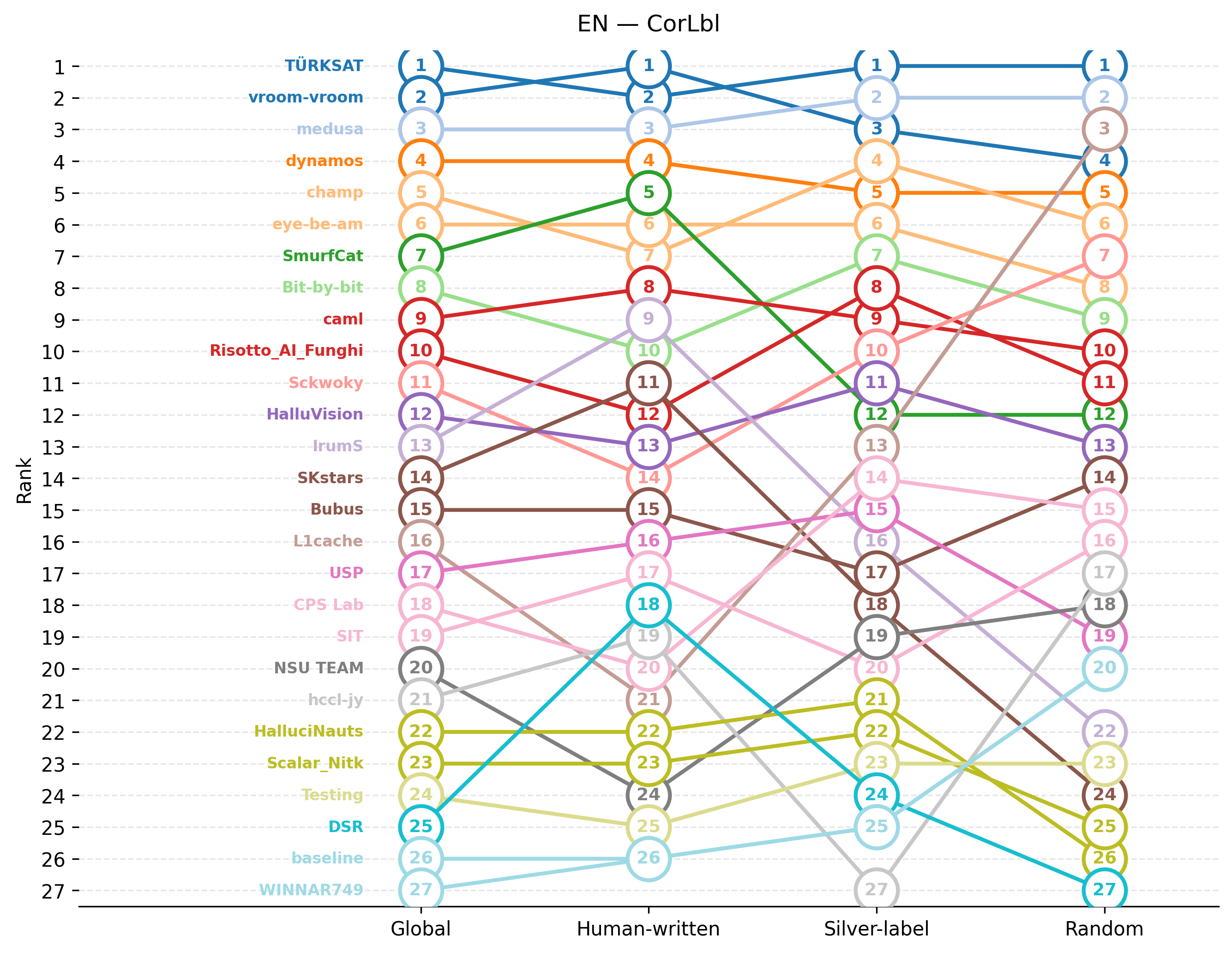}

        \caption{Rank flows across data partitions for the English systems on $\text{Corr}_\text{lbl}$. Each node represents a system's rank within a partition (General vs. Human/Silver/Random subsets).}
        \label{fig:rank-stability-en-corlbl}
    \end{minipage}
    \hfill
    % ================= CHINESE =================
    \begin{minipage}[t]{0.48\textwidth}
        \centering
        \includegraphics[
            width=\linewidth,
            trim={3cm 0 1cm 0.6cm},
            clip
        ]{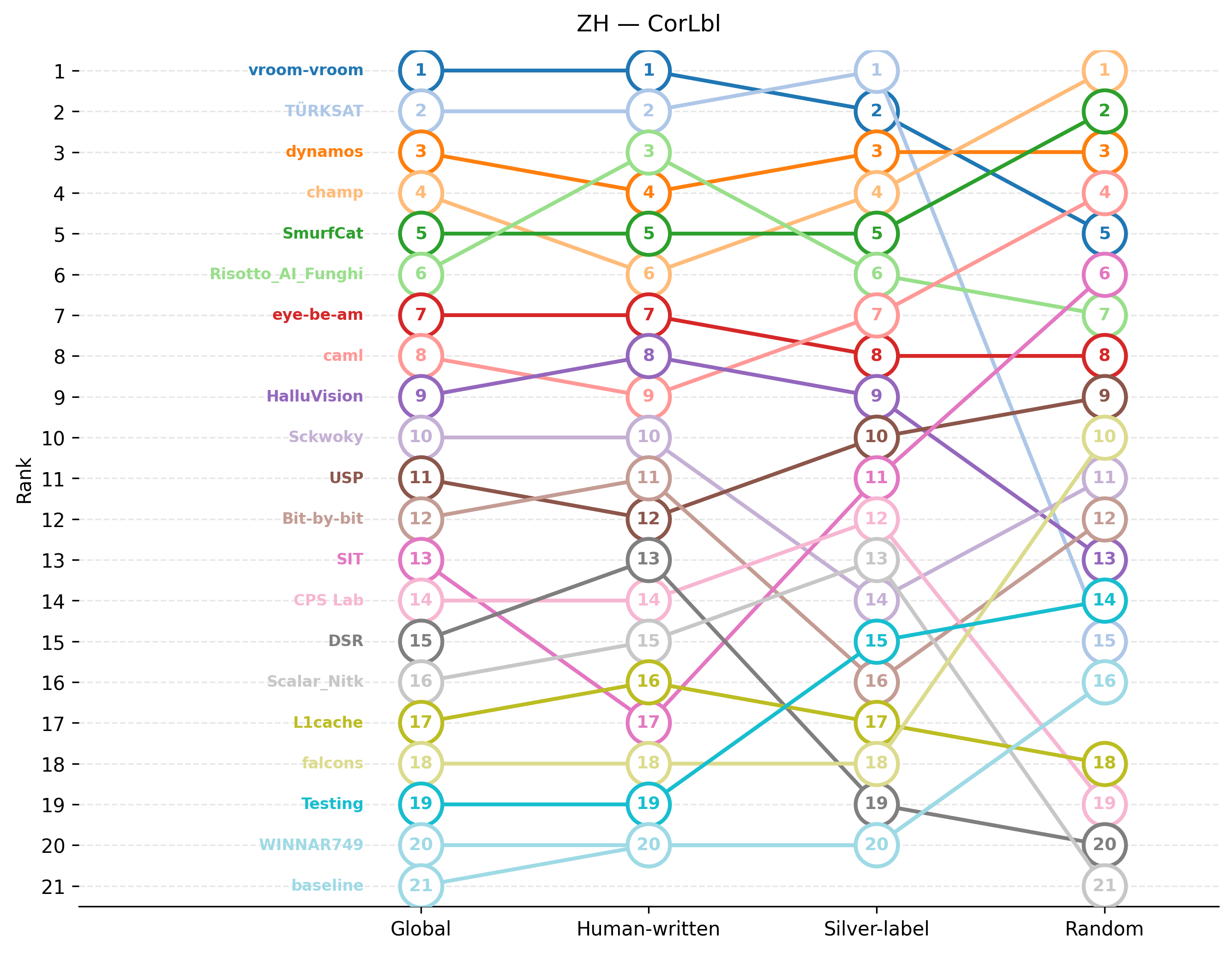}

        \caption{Rank flows across data partitions for the Chinese systems on $\text{Corr}_\text{lbl}$. Each node represents a system's rank within a partition (General vs. Human/Silver/Random subsets).}
        \label{fig:rank-stability-zh-corlbl}
    \end{minipage}
    \\[2cm]
        % ================= Italian =================
    \begin{minipage}[t]{0.48\textwidth}
        \centering
        \includegraphics[
            width=\linewidth,
            trim={3cm 0 1cm 0.6cm},
            clip
        ]{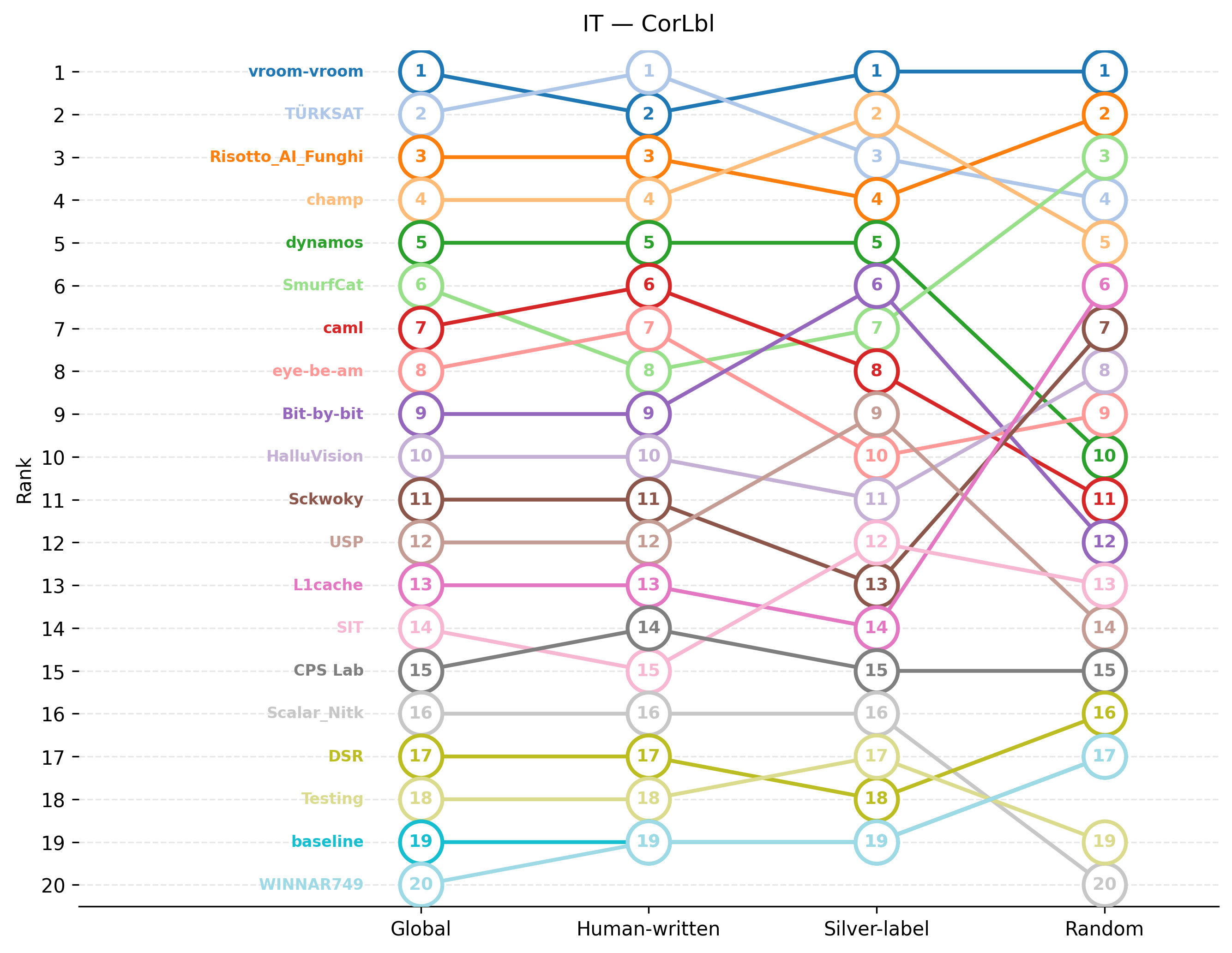}

        \caption{Rank flows across data partitions for the Italian systems on $\text{Corr}_\text{lbl}$. Each node represents a system's rank within a partition (General vs. Human/Silver/Random subsets). }
        \label{fig:rank-stability-it-corlbl}
    \end{minipage}
    \hfill
    % ================= French =================
    \begin{minipage}[t]{0.48\textwidth}
        \centering
        \includegraphics[
            width=\linewidth,
            trim={3cm 0 1cm 0.6cm},
            clip
        ]{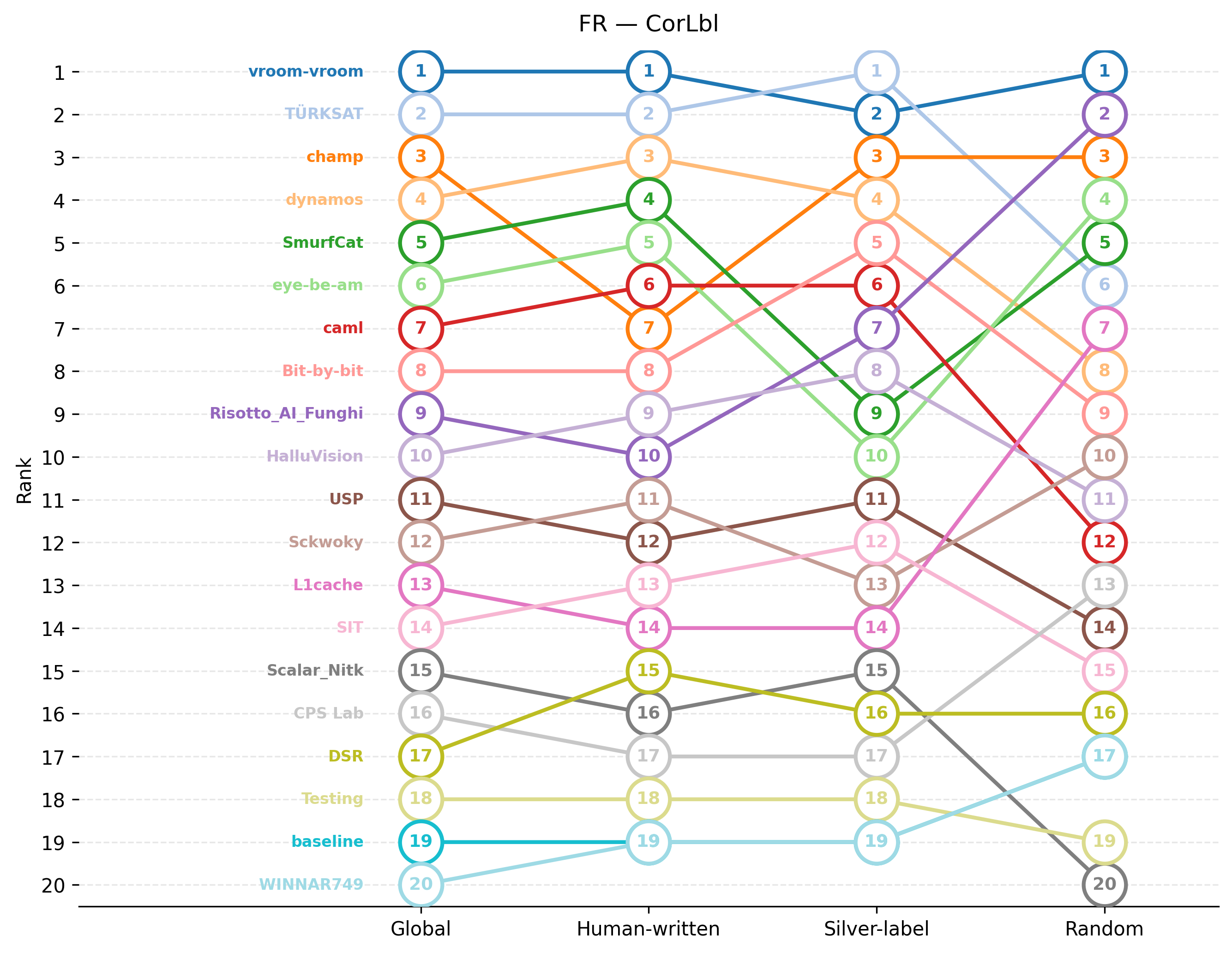}

        \caption{Rank flows across data partitions for the French systems on $\text{Corr}_\text{lbl}$. Each node represents a system's rank within a partition (General vs. Human/Silver/Random subsets).}
        \label{fig:rank-stability-fr-corlbl}
    \end{minipage}
\end{figure*}

\begin{table}[t]
\centering
\small
\setlength{\tabcolsep}{5pt}
\resizebox{\columnwidth}{!}{
\begin{tabular}{r l r r r r}
\toprule
\# & Team &
\multicolumn{1}{c}{Global} &
\multicolumn{1}{c}{Human} &
\multicolumn{1}{c}{Random} &
\multicolumn{1}{c}{Silver} \\
\midrule
1 & TÜRKSAT & \textbf{0.434} & 0.488 & \textbf{0.407} & \textbf{0.406} \\
2 & vroom-vroom & 0.425 & \textbf{0.536} & 0.374 & 0.366 \\
3 & medusa & 0.416 & 0.477 & 0.388 & 0.383 \\
4 & dynamos & 0.388 & 0.453 & 0.368 & 0.344 \\
5 & champ & 0.372 & 0.398 & 0.361 & 0.358 \\
6 & eye-be-am & 0.357 & 0.409 & 0.353 & 0.311 \\
7 & SmurfCat & 0.352 & 0.435 & 0.343 & 0.280 \\
8 & Bit-by-bit & 0.341 & 0.371 & 0.348 & 0.304 \\
9 & caml & 0.339 & 0.379 & 0.346 & 0.294 \\
10 & Risotto\_AI\_Funghi & 0.324 & 0.331 & 0.344 & 0.300 \\
11 & Sckwoky & 0.313 & 0.297 & 0.353 & 0.292 \\
12 & HalluVision & 0.308 & 0.301 & 0.337 & 0.287 \\
13 & IrumS & 0.302 & 0.373 & 0.292 & 0.242 \\
14 & SKstars & 0.290 & 0.350 & 0.288 & 0.236 \\
15 & Bubus & 0.282 & 0.287 & 0.320 & 0.242 \\
16 & USP & 0.267 & 0.260 & 0.300 & 0.245 \\
17 & L1cache & 0.267 & 0.169 & 0.377 & 0.265 \\
18 & CPS Lab & 0.245 & 0.176 & 0.310 & 0.252 \\
19 & SIT & 0.242 & 0.217 & 0.307 & 0.207 \\
20 & NSU TEAM & 0.217 & 0.145 & 0.302 & 0.207 \\
21 & hccl-jy & 0.214 & 0.184 & 0.303 & 0.162 \\
22 & HalluciNauts & 0.214 & 0.165 & 0.278 & 0.202 \\
23 & Scalar\_Nitk & 0.208 & 0.147 & 0.286 & 0.196 \\
24 & DSR & 0.198 & 0.190 & 0.228 & 0.177 \\
25 & Testing & 0.197 & 0.108 & 0.291 & 0.196 \\
26 & WINNAR749 & 0.155 & 0.013 & 0.296 & 0.162 \\
27 & baseline & 0.155 & 0.013 & 0.296 & 0.162 \\
\bottomrule
\end{tabular}}
\caption{Strategy-specific $\text{Corr}_\text{lbl}$ scores for the English leaderboard.
Teams are ordered by their score on the full dataset (Global).
Human-written, Random, and Silver-label report scores computed on the corresponding
data partitions. Boldface indicates the highest score within each column.}
\label{tab:strategy_scores_en_cor_lbl}
\end{table}
\begin{table}[t]
\centering
\small
\setlength{\tabcolsep}{5pt}
\begin{tabular}{r l r r r r}
\toprule
\# & Team &
\multicolumn{1}{c}{Global} &
\multicolumn{1}{c}{Human} &
\multicolumn{1}{c}{Rndm} &
\multicolumn{1}{c}{Silver} \\
\midrule
1 & TÜRKSAT & \textbf{0.487} & 0.551 & \textbf{0.444} & \textbf{0.464} \\
2 & medusa & 0.455 & 0.516 & 0.424 & 0.425 \\
3 & vroom-vroom & 0.449 & \textbf{0.565} & 0.388 & 0.393 \\
4 & champ & 0.432 & 0.482 & 0.390 & 0.422 \\
5 & Sckwoky & 0.419 & 0.453 & 0.408 & 0.396 \\
6 & Risotto\_AI\_F. & 0.418 & 0.483 & 0.383 & 0.389 \\
7 & eye-be-am & 0.410 & 0.458 & 0.400 & 0.373 \\
8 & SmurfCat & 0.405 & 0.479 & 0.375 & 0.361 \\
9 & NSU TEAM & 0.399 & 0.436 & 0.377 & 0.384 \\
10 & dynamos & 0.393 & 0.458 & 0.370 & 0.354 \\
11 & Bit-by-bit & 0.393 & 0.467 & 0.340 & 0.371 \\
12 & Bubus & 0.379 & 0.394 & 0.384 & 0.361 \\
13 & HalluVision & 0.368 & 0.373 & 0.381 & 0.350 \\
14 & IrumS & 0.342 & 0.416 & 0.322 & 0.290 \\
15 & caml & 0.337 & 0.384 & 0.346 & 0.284 \\
16 & SKstars & 0.315 & 0.394 & 0.302 & 0.252 \\
17 & hccl-jy & 0.291 & 0.270 & 0.340 & 0.265 \\
18 & USP & 0.281 & 0.277 & 0.306 & 0.261 \\
19 & L1cache & 0.262 & 0.200 & 0.314 & 0.274 \\
20 & DSR & 0.256 & 0.246 & 0.273 & 0.248 \\
21 & CPS Lab & 0.255 & 0.198 & 0.316 & 0.254 \\
22 & SIT & 0.255 & 0.227 & 0.311 & 0.230 \\
23 & HalluciNauts & 0.249 & 0.228 & 0.295 & 0.229 \\
24 & Scalar\_Nitk & 0.208 & 0.150 & 0.289 & 0.190 \\
25 & Testing & 0.187 & 0.094 & 0.289 & 0.183 \\
26 & WINNAR749 & 0.155 & 0.013 & 0.296 & 0.162 \\
27 & baseline & 0.155 & 0.013 & 0.296 & 0.162 \\
\bottomrule
\end{tabular}
\caption{Strategy-specific IoU scores for the English leaderboard. 
Teams are ordered by their score on the full dataset (Global). 
Human-written, Random, and Silver-label report scores computed on the corresponding 
data partitions. Boldface indicates the highest score within each column.}
\label{tab:strategy_scores_en_iou}
\end{table}

\begin{table}[t]
\centering
\small
\setlength{\tabcolsep}{5pt}
\begin{tabular}{r l r r r r}
\toprule
\# & Team &
\multicolumn{1}{c}{Global} &
\multicolumn{1}{c}{Human} &
\multicolumn{1}{c}{Rndm} &
\multicolumn{1}{c}{Silver} \\
\midrule
1 & vroom-vroom & \textbf{0.608} & \textbf{0.691} & 0.599 & \textbf{0.543} \\
2 & TÜRKSAT & 0.581 & 0.667 & 0.540 & 0.539 \\
3 & Risotto\_AI\_F. & 0.558 & 0.623 & 0.573 & 0.489 \\
4 & champ & 0.551 & 0.563 & \textbf{0.610} & 0.495 \\
5 & SmurfCat & 0.544 & 0.589 & 0.588 & 0.472 \\
6 & dynamos & 0.530 & 0.551 & 0.581 & 0.472 \\
7 & eye-be-am & 0.518 & 0.548 & 0.572 & 0.450 \\
8 & HalluVision & 0.512 & 0.535 & 0.545 & 0.465 \\
9 & Sckwoky & 0.488 & 0.508 & 0.568 & 0.410 \\
10 & caml & 0.478 & 0.460 & 0.566 & 0.426 \\
11 & DSR & 0.434 & 0.406 & 0.512 & 0.399 \\
12 & Bit-by-bit & 0.431 & 0.375 & 0.548 & 0.390 \\
13 & USP & 0.422 & 0.338 & 0.519 & 0.421 \\
14 & CPS Lab & 0.417 & 0.357 & 0.505 & 0.402 \\
15 & L1cache & 0.411 & 0.302 & 0.530 & 0.416 \\
16 & falcons & 0.380 & 0.265 & 0.537 & 0.360 \\
17 & SIT & 0.375 & 0.204 & 0.558 & 0.384 \\
18 & Scalar\_Nitk & 0.353 & 0.238 & 0.470 & 0.365 \\
19 & Testing & 0.323 & 0.149 & 0.501 & 0.339 \\
20 & WINNAR749 & 0.255 & 0.048 & 0.480 & 0.264 \\
21 & baseline & 0.255 & 0.048 & 0.480 & 0.264 \\
\bottomrule
\end{tabular}
\caption{Strategy-specific Corr scores for the Chinese leaderboard. 
Teams are ordered by their score on the full dataset (Global). 
Human-written, Random, and Silver-label report scores computed on the corresponding 
data partitions. Boldface indicates the highest score within each column.}
\label{tab:strategy_scores_zh_cor}
\end{table}
\begin{table}[t]
\centering
\small
\setlength{\tabcolsep}{5pt}
\resizebox{\columnwidth}{!}{
\begin{tabular}{r l r r r r}
\toprule
\# & Team &
\multicolumn{1}{c}{Global} &
\multicolumn{1}{c}{Human} &
\multicolumn{1}{c}{Random} &
\multicolumn{1}{c}{Silver} \\
\midrule
1 & vroom-vroom & \textbf{0.504} & \textbf{0.555} & 0.528 & 0.441 \\
2 & TÜRKSAT & 0.484 & 0.531 & 0.487 & \textbf{0.442} \\
3 & dynamos & 0.469 & 0.462 & 0.538 & 0.422 \\
4 & champ & 0.467 & 0.442 & \textbf{0.560} & 0.417 \\
5 & SmurfCat & 0.456 & 0.445 & 0.544 & 0.399 \\
6 & Risotto\_AI\_Funghi & 0.454 & 0.487 & 0.511 & 0.381 \\
7 & eye-be-am & 0.426 & 0.418 & 0.507 & 0.370 \\
8 & caml & 0.422 & 0.376 & 0.532 & 0.378 \\
9 & HalluVision & 0.415 & 0.403 & 0.490 & 0.367 \\
10 & Sckwoky & 0.376 & 0.332 & 0.497 & 0.321 \\
11 & USP & 0.360 & 0.228 & 0.503 & 0.366 \\
12 & Bit-by-bit & 0.347 & 0.259 & 0.495 & 0.311 \\
13 & SIT & 0.335 & 0.154 & 0.522 & 0.349 \\
14 & CPS Lab & 0.330 & 0.206 & 0.471 & 0.330 \\
15 & DSR & 0.309 & 0.210 & 0.454 & 0.283 \\
16 & Scalar\_Nitk & 0.308 & 0.173 & 0.442 & 0.324 \\
17 & L1cache & 0.307 & 0.162 & 0.472 & 0.306 \\
18 & falcons & 0.307 & 0.153 & 0.499 & 0.293 \\
19 & Testing & 0.301 & 0.121 & 0.487 & 0.315 \\
20 & WINNAR749 & 0.255 & 0.048 & 0.480 & 0.264 \\
21 & baseline & 0.255 & 0.048 & 0.480 & 0.264 \\
\bottomrule
\end{tabular}}
\caption{Strategy-specific $\text{Corr}_\text{lbl}$ scores for the Chinese leaderboard.
Teams are ordered by their score on the full dataset (Global).
Human-written, Random, and Silver-label report scores computed on the corresponding
data partitions. Boldface indicates the highest score within each column.}
\label{tab:strategy_scores_zh_cor_lbl}
\end{table}
\begin{table}[t]
\centering
\small
\setlength{\tabcolsep}{5pt}
\begin{tabular}{r l r r r r}
\toprule
\# & Team &
\multicolumn{1}{c}{Global} &
\multicolumn{1}{c}{Human} &
\multicolumn{1}{c}{Rndm} &
\multicolumn{1}{c}{Silver} \\
\midrule
1 & vroom-vroom & \textbf{0.534} & 0.574 & 0.566 & 0.476 \\
2 & TÜRKSAT & 0.524 & \textbf{0.575} & 0.515 & \textbf{0.487} \\
3 & Risotto\_AI\_F. & 0.516 & 0.557 & 0.553 & 0.452 \\
4 & SmurfCat & 0.501 & 0.507 & 0.570 & 0.442 \\
5 & champ & 0.489 & 0.458 & \textbf{0.579} & 0.448 \\
6 & Sckwoky & 0.466 & 0.468 & 0.551 & 0.401 \\
7 & eye-be-am & 0.460 & 0.449 & 0.532 & 0.416 \\
8 & dynamos & 0.458 & 0.426 & 0.546 & 0.419 \\
9 & HalluVision & 0.444 & 0.422 & 0.512 & 0.412 \\
10 & Bit-by-bit & 0.431 & 0.380 & 0.536 & 0.394 \\
11 & caml & 0.408 & 0.342 & 0.537 & 0.367 \\
12 & CPS Lab & 0.392 & 0.324 & 0.492 & 0.375 \\
13 & USP & 0.381 & 0.269 & 0.496 & 0.389 \\
14 & falcons & 0.375 & 0.253 & 0.530 & 0.362 \\
15 & SIT & 0.343 & 0.156 & 0.540 & 0.353 \\
16 & DSR & 0.339 & 0.237 & 0.460 & 0.334 \\
17 & Scalar\_Nitk & 0.323 & 0.194 & 0.458 & 0.331 \\
18 & Testing & 0.298 & 0.111 & 0.489 & 0.315 \\
19 & WINNAR749 & 0.255 & 0.048 & 0.480 & 0.264 \\
20 & baseline & 0.255 & 0.048 & 0.480 & 0.264 \\
\bottomrule
\end{tabular}
\caption{Strategy-specific IoU scores for the Chinese leaderboard. 
Teams are ordered by their score on the full dataset (Global). 
Human-written, Random, and Silver-label report scores computed on the corresponding 
data partitions. Boldface indicates the highest score within each column.}
\label{tab:strategy_scores_zh_iou}
\end{table}

\begin{table}[t]
\centering
\small
\setlength{\tabcolsep}{5pt}
\begin{tabular}{r l r r r r}
\toprule
\# & Team &
\multicolumn{1}{c}{Global} &
\multicolumn{1}{c}{Human} &
\multicolumn{1}{c}{Rndm} &
\multicolumn{1}{c}{Silver} \\
\midrule
1 & vroom-vroom & \textbf{0.563} & 0.624 & \textbf{0.535} & \textbf{0.534} \\
2 & TÜRKSAT & 0.553 & \textbf{0.665} & 0.483 & 0.514 \\
3 & Risotto\_AI\_F. & 0.548 & 0.610 & 0.520 & 0.519 \\
4 & champ & 0.540 & 0.600 & 0.497 & 0.523 \\
5 & dynamos & 0.513 & 0.572 & 0.463 & 0.501 \\
6 & Bit-by-bit & 0.505 & 0.538 & 0.464 & 0.507 \\
7 & SmurfCat & 0.499 & 0.553 & 0.464 & 0.481 \\
8 & Sckwoky & 0.493 & 0.520 & 0.495 & 0.468 \\
9 & eye-be-am & 0.491 & 0.551 & 0.467 & 0.459 \\
10 & caml & 0.483 & 0.537 & 0.442 & 0.469 \\
11 & HalluVision & 0.441 & 0.447 & 0.462 & 0.419 \\
12 & USP & 0.421 & 0.408 & 0.412 & 0.437 \\
13 & L1cache & 0.404 & 0.329 & 0.472 & 0.417 \\
14 & SIT & 0.383 & 0.318 & 0.418 & 0.411 \\
15 & CPS Lab & 0.378 & 0.344 & 0.407 & 0.386 \\
16 & DSR & 0.319 & 0.246 & 0.393 & 0.323 \\
17 & Scalar\_Nitk & 0.280 & 0.204 & 0.330 & 0.306 \\
18 & Testing & 0.247 & 0.144 & 0.320 & 0.278 \\
19 & baseline & 0.163 & 0.025 & 0.313 & 0.165 \\
20 & WINNAR749 & 0.163 & 0.025 & 0.313 & 0.165 \\
\bottomrule
\end{tabular}
\caption{Strategy-specific Corr scores for the Italian leaderboard. 
Teams are ordered by their score on the full dataset (Global). 
Human-written, Random, and Silver-label report scores computed on the corresponding 
data partitions. Boldface indicates the highest score within each column.}
\label{tab:strategy_scores_it_cor}
\end{table}
\begin{table}[t]
\centering
\small
\setlength{\tabcolsep}{5pt}
\begin{tabular}{r l r r r r}
\toprule
\# & Team &
\multicolumn{1}{c}{Global} &
\multicolumn{1}{c}{Human} &
\multicolumn{1}{c}{Rndm} &
\multicolumn{1}{c}{Silver} \\
\midrule
1 & vroom-vroom & \textbf{0.451} & 0.508 & \textbf{0.456} & \textbf{0.401} \\
2 & TÜRKSAT & 0.444 & \textbf{0.530} & 0.419 & 0.391 \\
3 & Risotto\_AI\_F. & 0.430 & 0.482 & 0.428 & 0.388 \\
4 & champ & 0.421 & 0.460 & 0.418 & 0.393 \\
5 & dynamos & 0.399 & 0.437 & 0.382 & 0.382 \\
6 & SmurfCat & 0.393 & 0.404 & 0.421 & 0.362 \\
7 & caml & 0.380 & 0.414 & 0.376 & 0.355 \\
8 & eye-be-am & 0.376 & 0.410 & 0.386 & 0.341 \\
9 & Bit-by-bit & 0.367 & 0.361 & 0.372 & 0.369 \\
10 & HalluVision & 0.353 & 0.330 & 0.396 & 0.340 \\
11 & Sckwoky & 0.337 & 0.327 & 0.404 & 0.297 \\
12 & USP & 0.334 & 0.298 & 0.361 & 0.343 \\
13 & L1cache & 0.297 & 0.210 & 0.405 & 0.290 \\
14 & SIT & 0.292 & 0.202 & 0.371 & 0.306 \\
15 & CPS Lab & 0.274 & 0.203 & 0.335 & 0.286 \\
16 & Scalar\_Nitk & 0.229 & 0.161 & 0.289 & 0.240 \\
17 & DSR & 0.225 & 0.136 & 0.330 & 0.219 \\
18 & Testing & 0.215 & 0.112 & 0.302 & 0.236 \\
19 & baseline & 0.163 & 0.025 & 0.313 & 0.165 \\
20 & WINNAR749 & 0.163 & 0.025 & 0.313 & 0.165 \\
\bottomrule
\end{tabular}
\caption{Strategy-specific $\text{Corr}_\text{lbl}$ scores for the Italian leaderboard. 
Teams are ordered by their score on the full dataset (Global). 
Human-written, Random, and Silver-label report scores computed on the corresponding 
data partitions. Boldface indicates the highest score within each column.}
\label{tab:strategy_scores_it_cor_lbl}
\end{table}
\begin{table}[t]
\centering
\small
\setlength{\tabcolsep}{5pt}
\begin{tabular}{r l r r r r}
\toprule
\# & Team &
\multicolumn{1}{c}{Global} &
\multicolumn{1}{c}{Human} &
\multicolumn{1}{c}{Rndm} &
\multicolumn{1}{c}{Silver} \\
\midrule
1 & TÜRKSAT & \textbf{0.490} & \textbf{0.571} & 0.445 & 0.458 \\
2 & vroom-vroom & 0.479 & 0.508 & \textbf{0.476} & 0.458 \\
3 & Risotto\_AI\_F. & 0.470 & 0.501 & 0.463 & 0.450 \\
4 & champ & 0.461 & 0.476 & 0.445 & \textbf{0.461} \\
5 & SmurfCat & 0.443 & 0.464 & 0.428 & 0.438 \\
6 & Sckwoky & 0.430 & 0.424 & 0.457 & 0.415 \\
7 & dynamos & 0.424 & 0.450 & 0.399 & 0.421 \\
8 & Bit-by-bit & 0.421 & 0.424 & 0.407 & 0.428 \\
9 & eye-be-am & 0.413 & 0.441 & 0.412 & 0.392 \\
10 & HalluVision & 0.391 & 0.355 & 0.440 & 0.383 \\
11 & caml & 0.385 & 0.399 & 0.373 & 0.382 \\
12 & USP & 0.353 & 0.315 & 0.362 & 0.378 \\
13 & CPS Lab & 0.320 & 0.266 & 0.364 & 0.331 \\
14 & SIT & 0.311 & 0.223 & 0.373 & 0.336 \\
15 & L1cache & 0.284 & 0.226 & 0.338 & 0.292 \\
16 & DSR & 0.246 & 0.164 & 0.335 & 0.246 \\
17 & Scalar\_Nitk & 0.243 & 0.159 & 0.310 & 0.263 \\
18 & Testing & 0.214 & 0.105 & 0.303 & 0.236 \\
19 & baseline & 0.163 & 0.025 & 0.313 & 0.165 \\
20 & WINNAR749 & 0.163 & 0.025 & 0.313 & 0.165 \\
\bottomrule
\end{tabular}
\caption{Strategy-specific IoU scores for the Italian leaderboard. 
Teams are ordered by their score on the full dataset (Global). 
Human-written, Random, and Silver-label report scores computed on the corresponding 
data partitions. Boldface indicates the highest score within each column.}
\label{tab:strategy_scores_it_iou}
\end{table}

\begin{table}[t]
\centering
\small
\setlength{\tabcolsep}{5pt}
\begin{tabular}{r l r r r r}
\toprule
\# & Team &
\multicolumn{1}{c}{Global} &
\multicolumn{1}{c}{Human} &
\multicolumn{1}{c}{Rndm} &
\multicolumn{1}{c}{Silver} \\
\midrule
1 & vroom-vroom & \textbf{0.587} & \textbf{0.678} & \textbf{0.537} & \textbf{0.550} \\
2 & TÜRKSAT & 0.547 & 0.621 & 0.484 & 0.536 \\
3 & champ & 0.517 & 0.554 & 0.496 & 0.501 \\
4 & Risotto\_AI\_F. & 0.506 & 0.548 & 0.504 & 0.472 \\
5 & dynamos & 0.504 & 0.583 & 0.462 & 0.468 \\
6 & eye-be-am & 0.484 & 0.560 & 0.487 & 0.416 \\
7 & SmurfCat & 0.484 & 0.569 & 0.463 & 0.426 \\
8 & caml & 0.481 & 0.552 & 0.441 & 0.451 \\
9 & Bit-by-bit & 0.471 & 0.517 & 0.439 & 0.458 \\
10 & HalluVision & 0.458 & 0.514 & 0.428 & 0.434 \\
11 & Sckwoky & 0.438 & 0.428 & 0.455 & 0.433 \\
12 & USP & 0.394 & 0.376 & 0.394 & 0.408 \\
13 & L1cache & 0.387 & 0.297 & 0.475 & 0.396 \\
14 & SIT & 0.360 & 0.302 & 0.395 & 0.382 \\
15 & DSR & 0.326 & 0.269 & 0.398 & 0.317 \\
16 & CPS Lab & 0.308 & 0.267 & 0.395 & 0.273 \\
17 & Scalar\_Nitk & 0.269 & 0.206 & 0.328 & 0.277 \\
18 & Testing & 0.215 & 0.107 & 0.323 & 0.222 \\
19 & baseline & 0.168 & 0.030 & 0.328 & 0.157 \\
20 & WINNAR749 & 0.168 & 0.030 & 0.328 & 0.157 \\
\bottomrule
\end{tabular}
\caption{Strategy-specific Corr scores for the French leaderboard. 
Teams are ordered by their score on the full dataset (Global). 
Human-written, Random, and Silver-label report scores computed on the corresponding 
data partitions. Boldface indicates the highest score within each column.}
\label{tab:strategy_scores_fr_cor}
\end{table}
\begin{table}[t]
\centering
\small
\setlength{\tabcolsep}{5pt}
\begin{tabular}{r l r r r r}
\toprule
\# & Team &
\multicolumn{1}{c}{Global} &
\multicolumn{1}{c}{Human} &
\multicolumn{1}{c}{Rndm} &
\multicolumn{1}{c}{Silver} \\
\midrule
1 & vroom-vroom & \textbf{0.474} & \textbf{0.540} & \textbf{0.451} & 0.435 \\
2 & TÜRKSAT & 0.451 & 0.493 & 0.421 & \textbf{0.440} \\
3 & champ & 0.410 & 0.403 & 0.426 & 0.402 \\
4 & dynamos & 0.407 & 0.442 & 0.391 & 0.389 \\
5 & SmurfCat & 0.397 & 0.437 & 0.421 & 0.342 \\
6 & eye-be-am & 0.388 & 0.407 & 0.426 & 0.341 \\
7 & caml & 0.376 & 0.403 & 0.373 & 0.356 \\
8 & Risotto\_AI\_F. & 0.375 & 0.345 & 0.431 & 0.355 \\
9 & Bit-by-bit & 0.375 & 0.374 & 0.389 & 0.363 \\
10 & HalluVision & 0.362 & 0.364 & 0.375 & 0.349 \\
11 & USP & 0.320 & 0.272 & 0.359 & 0.331 \\
12 & Sckwoky & 0.318 & 0.275 & 0.379 & 0.307 \\
13 & L1cache & 0.290 & 0.178 & 0.417 & 0.287 \\
14 & SIT & 0.288 & 0.202 & 0.357 & 0.308 \\
15 & CPS Lab & 0.243 & 0.153 & 0.369 & 0.218 \\
16 & Scalar\_Nitk & 0.242 & 0.159 & 0.304 & 0.264 \\
17 & DSR & 0.240 & 0.161 & 0.340 & 0.226 \\
18 & Testing & 0.205 & 0.091 & 0.320 & 0.210 \\
19 & baseline & 0.168 & 0.030 & 0.328 & 0.157 \\
20 & WINNAR749 & 0.168 & 0.030 & 0.328 & 0.157 \\
\bottomrule
\end{tabular}
\caption{Strategy-specific $\text{Corr}_\text{lbl}$ scores for the French leaderboard. 
Teams are ordered by their score on the full dataset (Global). 
Human-written, Random, and Silver-label report scores computed on the corresponding 
data partitions. Boldface indicates the highest score within each column.}
\label{tab:strategy_scores_fr_cor_lbl}
\end{table}
\begin{table}[t]
\centering
\small
\setlength{\tabcolsep}{5pt}
\begin{tabular}{r l r r r r}
\toprule
\# & Team &
\multicolumn{1}{c}{Global} &
\multicolumn{1}{c}{Human} &
\multicolumn{1}{c}{Rndm} &
\multicolumn{1}{c}{Silver} \\
\midrule
1 & vroom-vroom & \textbf{0.515} & \textbf{0.572} & \textbf{0.487} & 0.488 \\
2 & TÜRKSAT & 0.500 & 0.541 & 0.460 & \textbf{0.497} \\
3 & champ & 0.478 & 0.485 & 0.466 & 0.481 \\
4 & Risotto\_AI\_F. & 0.457 & 0.463 & 0.477 & 0.434 \\
5 & Sckwoky & 0.447 & 0.417 & 0.474 & 0.451 \\
6 & dynamos & 0.440 & 0.466 & 0.414 & 0.439 \\
7 & SmurfCat & 0.436 & 0.476 & 0.447 & 0.392 \\
8 & eye-be-am & 0.433 & 0.473 & 0.446 & 0.388 \\
9 & HalluVision & 0.423 & 0.403 & 0.429 & 0.434 \\
10 & Bit-by-bit & 0.414 & 0.436 & 0.377 & 0.425 \\
11 & caml & 0.399 & 0.435 & 0.368 & 0.394 \\
12 & USP & 0.337 & 0.293 & 0.357 & 0.360 \\
13 & SIT & 0.299 & 0.223 & 0.357 & 0.319 \\
14 & L1cache & 0.293 & 0.185 & 0.394 & 0.305 \\
15 & DSR & 0.272 & 0.181 & 0.369 & 0.272 \\
16 & CPS Lab & 0.267 & 0.206 & 0.367 & 0.239 \\
17 & Scalar\_Nitk & 0.236 & 0.162 & 0.310 & 0.241 \\
18 & Testing & 0.193 & 0.078 & 0.313 & 0.196 \\
19 & baseline & 0.168 & 0.030 & 0.328 & 0.157 \\
20 & WINNAR749 & 0.168 & 0.030 & 0.328 & 0.157 \\
\bottomrule
\end{tabular}
\caption{Strategy-specific IoU scores for the French leaderboard. 
Teams are ordered by their score on the full dataset (Global). 
Human-written, Random, and Silver-label report scores computed on the corresponding 
data partitions. Boldface indicates the highest score within each column.}
\label{tab:strategy_scores_fr_iou}
\end{table}

\section{Curiouser and curiouser! Additional metrics}

To further characterize the span extraction performance, we report additional span-level metrics in \Cref{tab:f1-analysis}. Precision measures the reliability of extracted spans by quantifying the proportion of predicted spans that correspond to valid reference annotations, while recall captures the extent to which annotated spans are recovered by the model. We combine these two aspects through the F1 score, which provides a balanced measure of extraction quality under the trade-off between false positive and false negative predictions. In addition, we report span accuracy, computed as $\frac{TP}{TP+FP+FN}$, to quantify the agreement between predicted and reference span sets while accounting for both spurious detections and missed annotations. This formulation is preferred over conventional accuracy because the space of possible non-annotated spans is not explicitly defined in span extraction tasks, making true negatives unsuitable as an evaluation signal. All metrics are calculated at the class level to provide a detailed view of performance across different error categories.

\begin{table}[]
    \centering
    \begin{tabular}{cccc}
     \toprule
        \textbf{Lang.} & \textbf{Annot. abs. \%} & \textbf{Total} & \textbf{System abs.\%}\\
        \midrule
        %EN &  186& 1201\\
        %FR & 207& 1233\\
        %IT & 205& 1254\\
        %ZH & 309&1210\\
        EN &  15.48& 1201 & 31.78\\
        FR & 16.78& 1233 & 30.65\\
        IT & 16.34& 1254 & 32.91\\
        ZH & 25.53& 1210 & 45.98\\
        \midrule
        \textbf{Total} & 18.51& 4898 & 34.87\\
         \bottomrule
    \end{tabular}
    \caption{Comparison of empty (abstinence) annotation versus participation system submissions on test set.}
    \label{tab:stats_emptyannot}
\end{table}

\begin{figure*}[t!]
    \centering
    \hspace{-0.2cm}
    \begin{subfigure}{0.4\linewidth}
        \centering
        \includegraphics[width=0.8\linewidth, trim={.35cm 0 0 0cm}, clip]{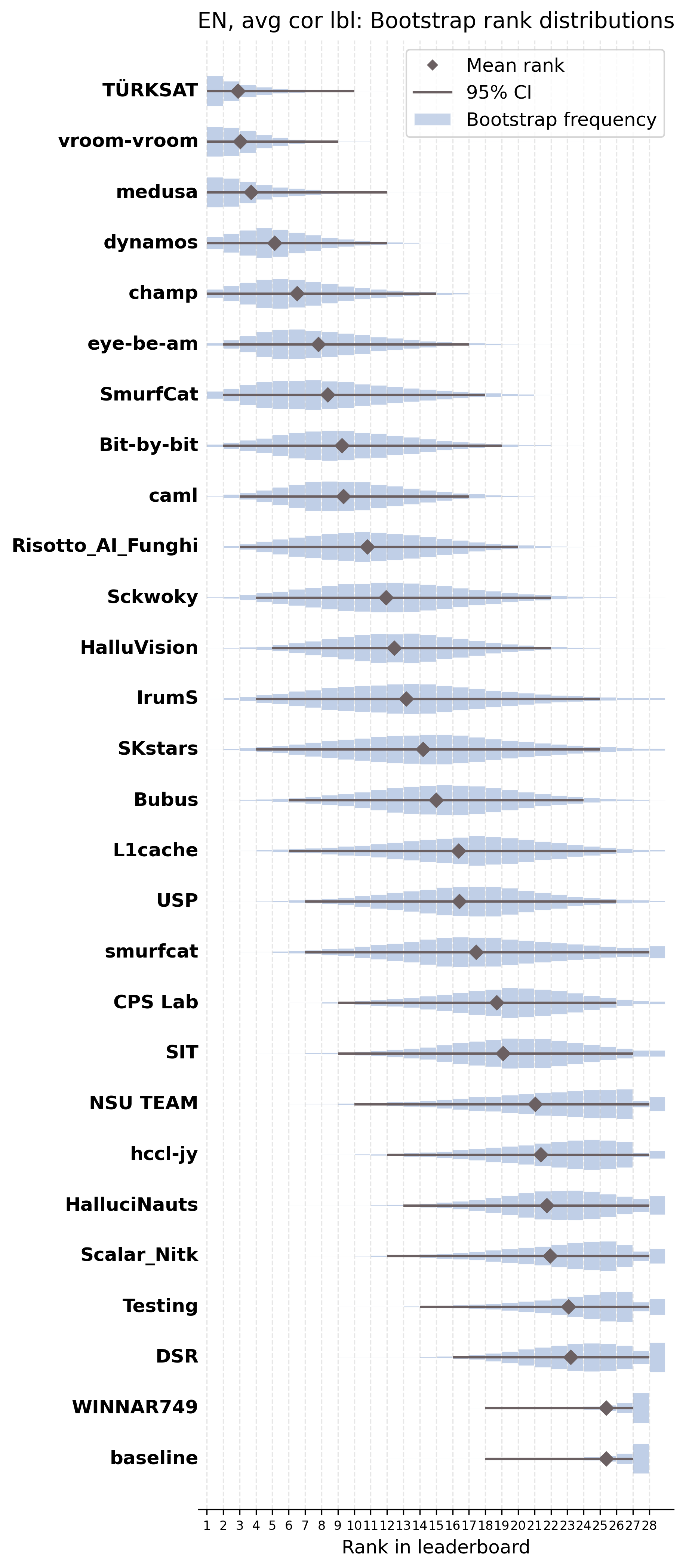}
        % \caption{FR}
    \end{subfigure}
    \begin{subfigure}{0.4\linewidth}
        \centering
        \includegraphics[width=1\linewidth, trim={.35cm 0 0 0cm}, clip]{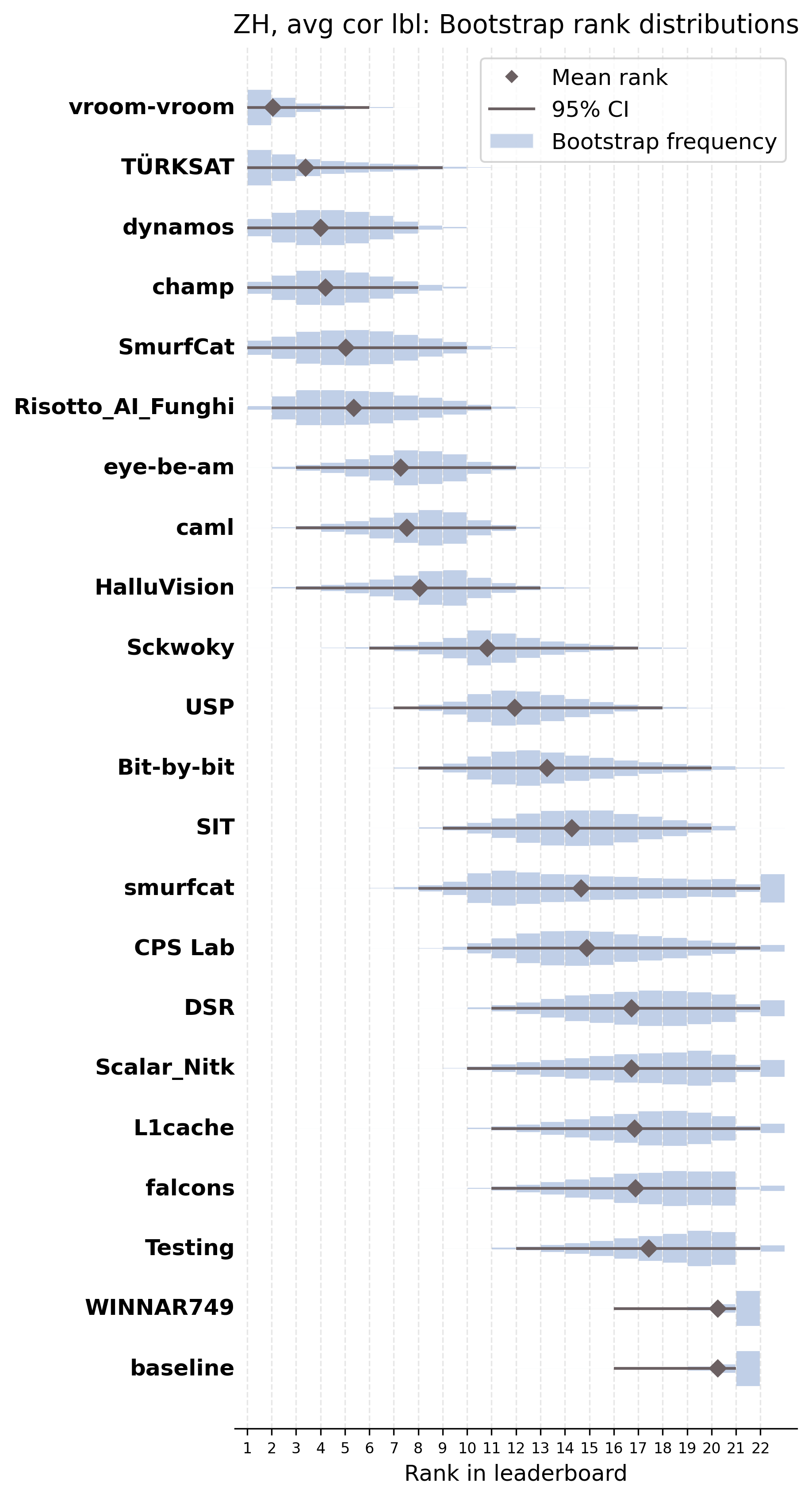}
        % \caption{FR}
    \end{subfigure}\\
    \begin{subfigure}{0.4\linewidth}
        \centering
        \includegraphics[width=1\linewidth, trim={.35cm 0 0 0cm}, clip]{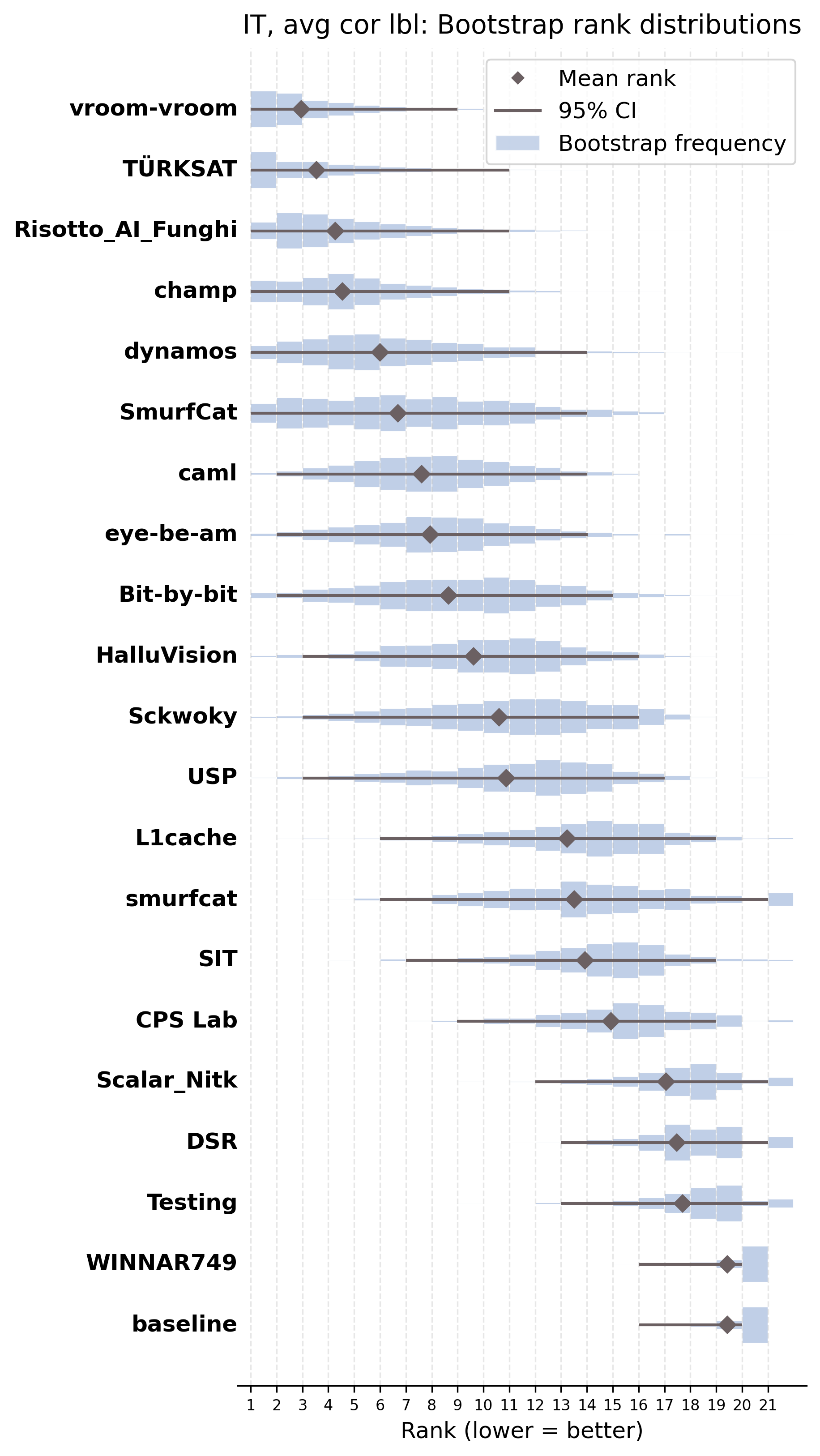}
        % \caption{IT}
    \end{subfigure}
    % \vspace{-2mm}
    \begin{subfigure}{0.4\linewidth}
        \centering
        \includegraphics[width=1\linewidth, trim={.35cm 0 0.2cm 0cm}, clip]{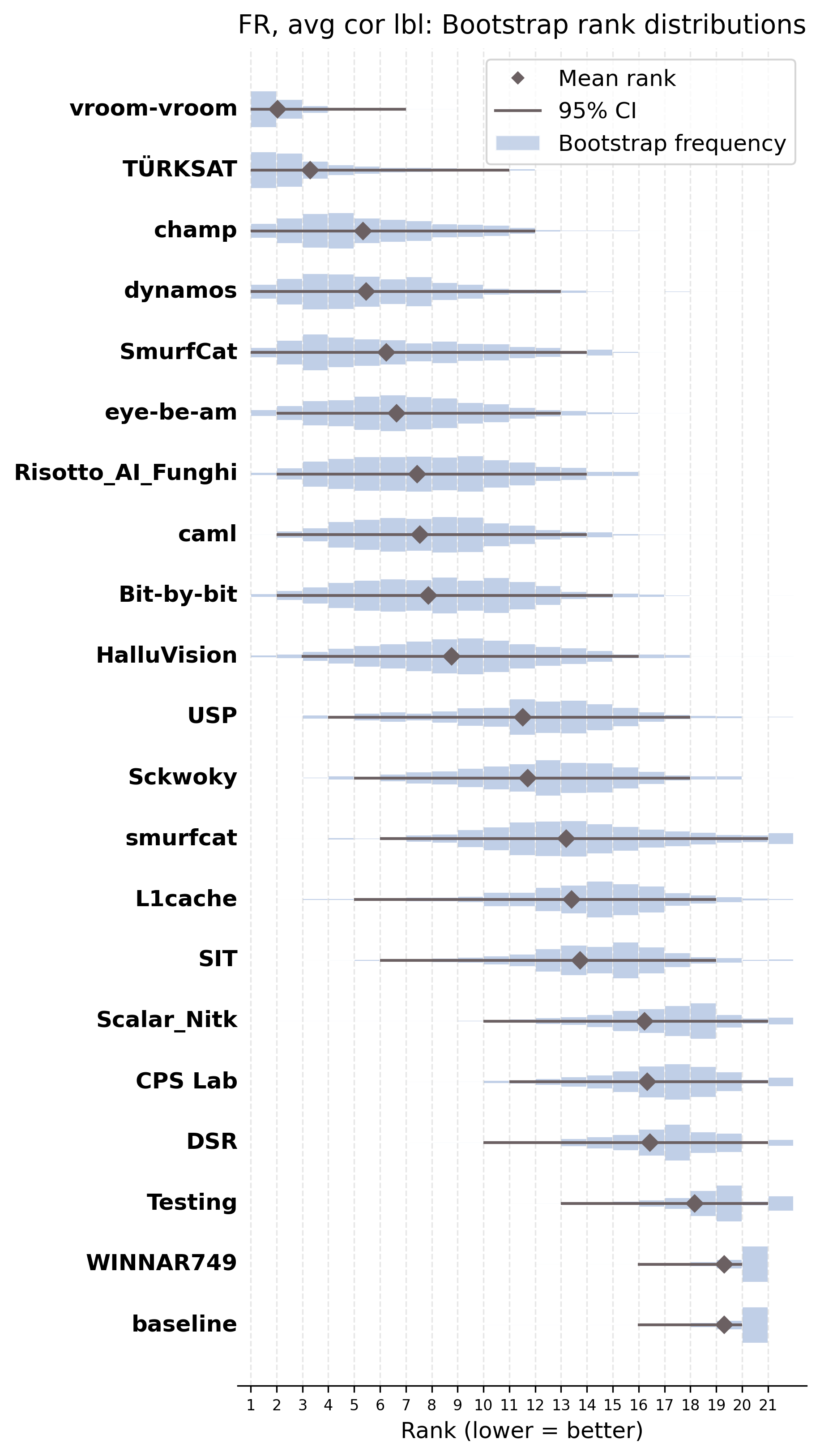}
        % \caption{ZH}
    \end{subfigure}
    % \vspace{-5mm}

    \caption{Bootstrap rank distributions with means and confidence intervals at 95\% for ZH, IT, FR submissions under $\text{Corr}_\text{lbl}$ metric. Distributions show rank frequencies across 25,000 bootstrap samples.}
    \label{fig:all_rank-dist}
\end{figure*}

\begin{figure*}
    \centering
    \includegraphics[width=0.8\textwidth]{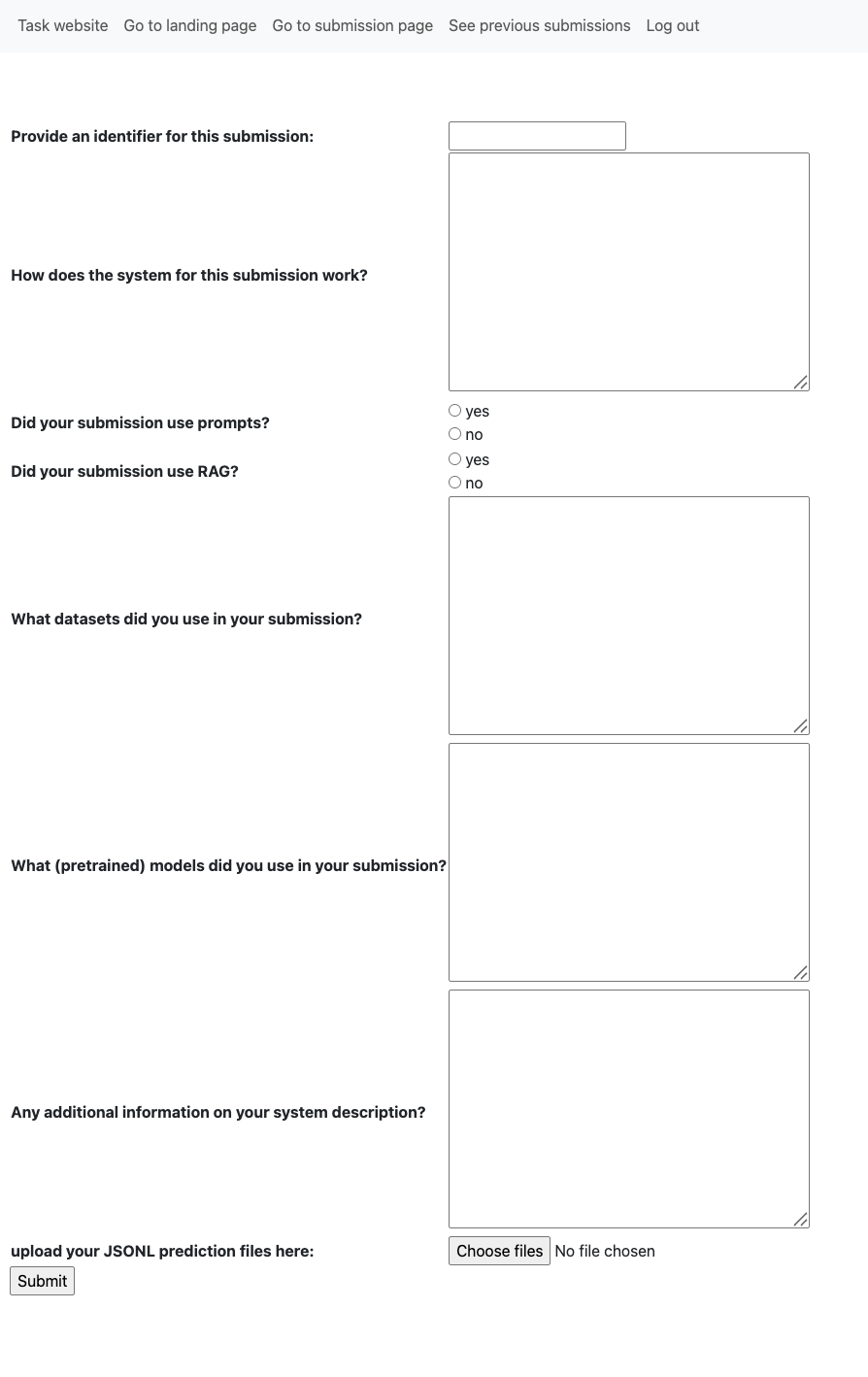}
    \caption{SHROOM-Visions platform participant submission interface.}
    \label{fig:submission-platform}
\end{figure*}

\begin{table*}[]
    \centering

\resizebox{\textwidth}{!}{
\begin{tabular}{l l cccc}
\toprule
\textbf{Class} & \textbf{Lang} &\textbf{Precision} & \textbf{Recall} & \textbf{F1-Score} & \textbf{Span-Acc} \\
\midrule
\texttt{OCR} & EN &0.274$_{\pm0.197}$/0.672& 0.233$_{\pm0.151}$/0.533& 0.209$_{\pm0.136}$/0.516& 0.123$_{\pm0.087}$/0.347\\
 &FR& 0.225$_{\pm0.177}$/0.583& 0.260$_{\pm0.172}$/0.601& 0.195$_{\pm0.130}$/0.512& 0.114$_{\pm0.083}$/0.344\\
 &IT& 0.241$_{\pm0.193}$/0.685& 0.274$_{\pm0.175}$/0.611& 0.195$_{\pm0.138}$/0.536& 0.115$_{\pm0.089}$/0.366\\
 & ZH&0.251$_{\pm0.181}$/0.652& 0.240$_{\pm0.164}$/0.663& 0.188$_{\pm0.115}$/0.510& 0.108$_{\pm0.072}$/0.342\\

\midrule
\texttt{invention} &EN& 0.136$_{\pm0.104}$/0.453& 0.148$_{\pm0.103}$/0.466& 0.110$_{\pm0.077}$/0.319& 0.060$_{\pm0.044}$/0.190\\
& FR& 0.113$_{\pm0.103}$/0.448& 0.172$_{\pm0.094}$/0.431& 0.111$_{\pm0.083}$/0.288& 0.061$_{\pm0.047}$/0.168\\
 &IT& 0.159$_{\pm0.122}$/0.494& 0.189$_{\pm0.111}$/0.480& 0.142$_{\pm0.094}$/0.330& 0.079$_{\pm0.055}$/0.198\\
 &ZH& 0.150$_{\pm0.109}$/0.506& 0.167$_{\pm0.089}$/0.357& 0.139$_{\pm0.088}$/0.344& 0.077$_{\pm0.052}$/0.208\\

\midrule
\texttt{mischar}.& EN& 0.189$_{\pm0.143}$/0.608& 0.143$_{\pm0.113}$/0.471& 0.122$_{\pm0.083}$/0.338& 0.067$_{\pm0.048}$/0.203\\
 &FR& 0.158$_{\pm0.132}$/0.500& 0.160$_{\pm0.106}$/0.456& 0.123$_{\pm0.088}$/0.322& 0.068$_{\pm0.050}$/0.192\\
 &IT& 0.201$_{\pm0.155}$/0.526& 0.176$_{\pm0.114}$/0.478& 0.141$_{\pm0.098}$/0.335& 0.079$_{\pm0.057}$/0.201\\
&ZH & 0.222$_{\pm0.181}$/0.542& 0.141$_{\pm0.121}$/0.674& 0.112$_{\pm0.090}$/0.345& 0.062$_{\pm0.052}$/0.208\\

\midrule
\texttt{miscoun.} & EN&0.316$_{\pm0.210}$/0.726& 0.335$_{\pm0.199}$/0.672& 0.288$_{\pm0.182}$/0.601& 0.181$_{\pm0.122}$/0.430\\
 &FR& 0.369$_{\pm0.256}$/0.824& 0.439$_{\pm0.252}$/0.774& 0.349$_{\pm0.233}$/0.647& 0.234$_{\pm0.168}$/0.479\\
 &IT& 0.363$_{\pm0.241}$/0.760& 0.376$_{\pm0.199}$/0.621& 0.319$_{\pm0.199}$/0.621& 0.206$_{\pm0.140}$/0.451\\
 & ZH&0.348$_{\pm0.246}$/0.780& 0.321$_{\pm0.206}$/0.845& 0.272$_{\pm0.176}$/0.588& 0.169$_{\pm0.118}$/0.416\\

\midrule

\texttt{Other} & EN&0.069$_{\pm0.105}$/0.667& 0.075$_{\pm0.068}$/0.386& 0.049$_{\pm0.057}$/0.239& 0.026$_{\pm0.031}$/0.136\\
 &FR& 0.023$_{\pm0.047}$/0.333& 0.038$_{\pm0.055}$/0.378& 0.020$_{\pm0.031}$/0.163& 0.010$_{\pm0.017}$/0.089\\
 &IT& 0.041$_{\pm0.068}$/0.400& 0.045$_{\pm0.071}$/0.526& 0.028$_{\pm0.036}$/0.213& 0.015$_{\pm0.019}$/0.119\\
 & ZH&0.017$_{\pm0.038}$/0.250& 0.025$_{\pm0.046}$/0.353& 0.013$_{\pm0.022}$/0.118& 0.007$_{\pm0.012}$/0.063\\
\bottomrule
\end{tabular}}
    \caption{Hallucination class-level summary of the participants system across all the languages. Each entry denotes mean$_{\pm\text{std}}$/max score format (min score = 0.0 for all, so redacted for better reability).}
    \label{tab:f1-analysis}
\end{table*}

\end{document}